\RequirePackage{fix-cm}
\documentclass[twocolumn,twoside]{svjour3}

\usepackage[switch]{lineno}
\usepackage{graphicx}
\usepackage{tabularx}
\usepackage{comment}
\usepackage{amsmath,amssymb,amsfonts}
\usepackage[x11names, table]{xcolor}
\usepackage{mathtools}
\usepackage{color}
\usepackage{multirow}
\usepackage{makecell}
\usepackage{tikz}
\usepackage{fbox}
\usepackage{booktabs}
\usepackage{wrapfig}
\usepackage{enumerate}
\usepackage{enumitem}
\usepackage{rotating}
\usepackage{diagbox}
\usepackage{cite}
\usepackage{marginnote}
\usepackage{microtype}

\usepackage[linesnumbered,ruled,vlined]{algorithm2e}
\usepackage[misc]{ifsym}
\usepackage{circledsteps}
\graphicspath{{fig/}}

\usepackage[pagebackref=true,breaklinks=true,letterpaper=true,colorlinks,bookmarks=false]{hyperref}

\begin{document}

\title{\textbf{A Decade of Action Quality Assessment: Largest Systematic Survey of Trends, Challenges, and Future Directions}}

\titlerunning{A Decade of Action Quality Assessment}
\authorrunning{Hao Yin and Paritosh Parmar et al.}

\author{\rm Hao Yin\textsuperscript{\rm 1, \rm 2*}\thanks{*These authors contributed equally to this work.}\!\and
        \!Paritosh Parmar\textsuperscript{\rm 3*}\!\and
        \!Daoliang Xu\textsuperscript{\rm 1, \rm 2}\!\and
        \!Yang Zhang\textsuperscript{\rm 2}\!\and \\
        \!Tianyou Zheng\textsuperscript{\rm 2 \textrm{\Letter}}\!\and
        \!Weiwei Fu\textsuperscript{\rm 1, \rm 2 \textrm{\Letter}}\!
}

\institute{\textrm{\Letter} Weiwei Fu \at \email{fuww@sibet.ac.cn}\\
            \textrm{\Letter} Tianyou Zheng \at \email{zhengty@sibet.ac.cn}\\
            \textsuperscript{\rm 1} School of Biomedical Engineering (Suzhou), Division of Life Sciences and Medicine, University of Science and Technology of China, Hefei, Anhui, China\\
            \textsuperscript{\rm 2} Suzhou Institute of Biomedical Engineering and Technology, Chinese Academy of Science, Suzhou, Jiangsu, China\\
            \textsuperscript{\rm 3} Institute of High Performance Computing, Agency for Science, Technology and Research, Singapore\\
}

\maketitle

\begin{abstract}
    Action Quality Assessment (AQA)---the ability to quantify the quality of human motion, actions, or skill levels and provide feedback---has far-reaching implications in areas such as low-cost physiotherapy, sports training, and workforce development. As such, it has become a critical field in computer vision \& video understanding over the past decade. Significant progress has been made in AQA methodologies, datasets, \& applications, yet a pressing need remains for a comprehensive synthesis of this rapidly evolving field. In this paper, we present a thorough survey of the AQA landscape, systematically reviewing over 200 research papers using the preferred reporting items for systematic reviews \& meta-analyses (PRISMA) framework. We begin by covering foundational concepts \& definitions, then move to general frameworks \& performance metrics, \& finally discuss the latest advances in methodologies \& datasets. This survey provides a detailed analysis of research trends, performance comparisons, challenges, \& future directions. Through this work, we aim to offer a valuable resource for both newcomers \& experienced researchers, promoting further exploration \& progress in AQA. Data are available at \url{https://haoyin116.github.io/Survey_of_AQA/}.
    
\end{abstract}
\maketitle

\section{Introduction}\label{sec1}

\begin{figure}[h]
    \centering
    \includegraphics[width=\columnwidth]{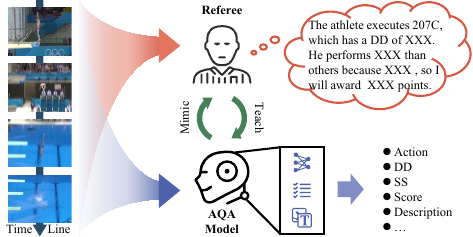}
    \caption{\textbf{AQA model plays the role of referee to evaluate how well actions are performed and score the skill level of performers.}}
    \label{fig_AQA}
\end{figure}

Skills and Action Quality Assessment (\textbf{AQA}) is an \textit{emerging} and \textit{critical} field in video understanding, moving beyond action recognition and action prediction\cite{kong2022human} to \textbf{evaluate how well} actions are performed and score the skill level of performers (see \autoref{fig_AQA}). These techniques are essential in a range of \textbf{domains}, including \textit{sports}\cite{pirsiavash2014assessing, parmar2017learning}, \textit{healthcare}\cite{gao2014jhu, vakanski2018data, capecci2019kimore, li2019automated}, \textit{fitness}\cite{parmar2022domain, li2024egoexo, gu2024exechecker},  \textit{industrial training}\cite{sener2022assembly101, seminara2024differentiable}, \& \textit{AI video content generation}\cite{chen2024gaia} where accurate assessment of human actions/performance is crucial.

For example, in \textit{sports}\cite{pirsiavash2014assessing, parmar2017learning}, it could be used to assess how well an Olympics diver performed, reporting what they did well/correctly, what they did wrong and what was the severity of these errors; take all (ideally) such factors into consideration and quantify the how well was the performance of the diver. Similarly, it could be used for other sports or \textit{physical rehabilitation}\cite{vakanski2018data, capecci2019kimore} or \textit{managing diseases} like Cerebral Palsy \cite{parmar2016measuring}. AQA can be used to train future surgeons by assessing their \textit{surgical performances}\cite{nespolo2024assessing}, assessing them, and giving feedback on what can be improved. In \textit{manufacturing sector}\cite{sener2022assembly101}, AQA can be used to assess the skill levels of workers, train them, and ensure their actions follow safety standards. Over the last few years, \textit{generative AI}\cite{chen2024gaia} has been on the rise, where AQA could serve as an objective metric for evaluating AI-generated video content. 

Automated AQA systems have \textbf{far-reaching implications} across diverse sectors, from \textit{promoting social equity} to \textit{enhancing industrial efficiency}. On a societal level, AQA can democratize access to training and evaluation tools, particularly benefiting under-resourced communities. Low-cost, AI-driven assessments provide underserved athletes with tailored feedback, support fair sports judging\cite{pirsiavash2014assessing, parmar2017learning}, and facilitate remote patient evaluations in healthcare\cite{gao2014jhu, vakanski2018data, capecci2019kimore}, improving access for rural or low-income populations. In skilled actions/vocational training, AQA offers objective skill assessments, helping workers certify and enhance their technical skills\cite{doughty2018s,doughty2019pros}, which is essential for career growth. In industries like manufacturing and automation, AQA can reduce errors, ensure product consistency, and support a skilled workforce, leading to lower costs and higher productivity\cite{sener2022assembly101}. By assessing both human and robotic systems, AQA enables effective human-robot collaboration, boosting workplace safety and efficiency. Looking ahead, the development of ethical and unbiased AQA systems will ensure that these technologies can be equitably applied across various demographics, preventing algorithmic biases from perpetuating inequalities. As such, the continued growth and refinement of AQA technology promises to not only enhance performance and quality control but also play a pivotal role in creating more inclusive, accessible, and efficient systems that benefit society as a whole. As generative AI becomes more prevalent, AQA can also serve as an objective metric for assessing the quality of AI-generated content\cite{chen2024gaia}, ensuring transparency and trust in AI systems.

\begin{figure}[!h]
    \centering
    \includegraphics[width=\columnwidth]{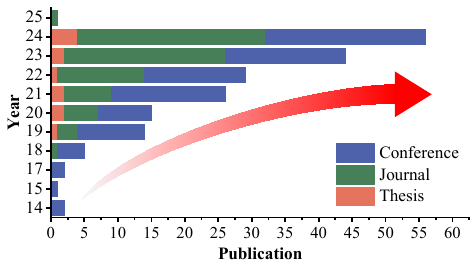}
    \caption{\textbf{The number of papers related to AQA over the past decade.}}
    \label{fig_Num}
\end{figure}

AQA has evolved significantly over the past decade (see \autoref{fig_Num}), marking major advancements across sports\cite{pirsiavash2014assessing, parmar2017learning}, healthcare\cite{gao2014jhu, vakanski2018data, capecci2019kimore}, fitness\cite{parmar2022domain, li2024egoexo}, industrial training\cite{sener2022assembly101}, and generative AI evaluation\cite{chen2024gaia}. With the last comprehensive survey now outdated\cite{lei2019survey, wang2021survey}, there's a critical need for a current synthesis of this fast-growing field. New computer vision and machine learning methodologies have expanded the capabilities of AQA, introducing refined evaluation techniques, metrics, and cross-disciplinary applications that address diverse sector-specific needs—from precise feedback in sports performance to accessibility in remote patient healthcare.

This survey paper addresses these developments by offering a structured overview of the latest AQA methodologies, benchmarks, and evaluation metrics. We also examine the field's unique ethical and societal challenges, emphasizing the importance of unbiased, inclusive assessment systems that promote equitable access to training and evaluation resources across demographics. By consolidating recent progress and illuminating future research directions, this survey aims to serve as a foundational guide for both researchers and practitioners, providing insights that drive innovation in creating accessible, robust, and fair AQA systems.

In this survey, we employed the Preferred Reporting Items for Systematic Reviews and Meta-Analyses (\textbf{PRISMA})\cite{page2021prisma} framework to systematically select relevant papers and continuously update until December 2024. This process unfolded over four key stages: \textbf{1)} \textit{Identification}: A search for the keyword' action quality assessment' in Web of Science and Google Scholar identified 505 potentially relevant papers. \textbf{2)} \textit{Screening}: After initial screening based on titles and abstracts, we removed duplicates and irrelevant entries, leaving 276 papers. \textbf{3)} \textit{Eligibility}: We reviewed the full text of these papers to assess quality, documenting each study's approach, results, and novelty. \textbf{4)} \textit{Inclusion}: Ultimately, 195 papers, including 26 datasets, met the inclusion criteria for this survey.

\begin{figure}[!h]
    \centering
    \includegraphics[width=\columnwidth]{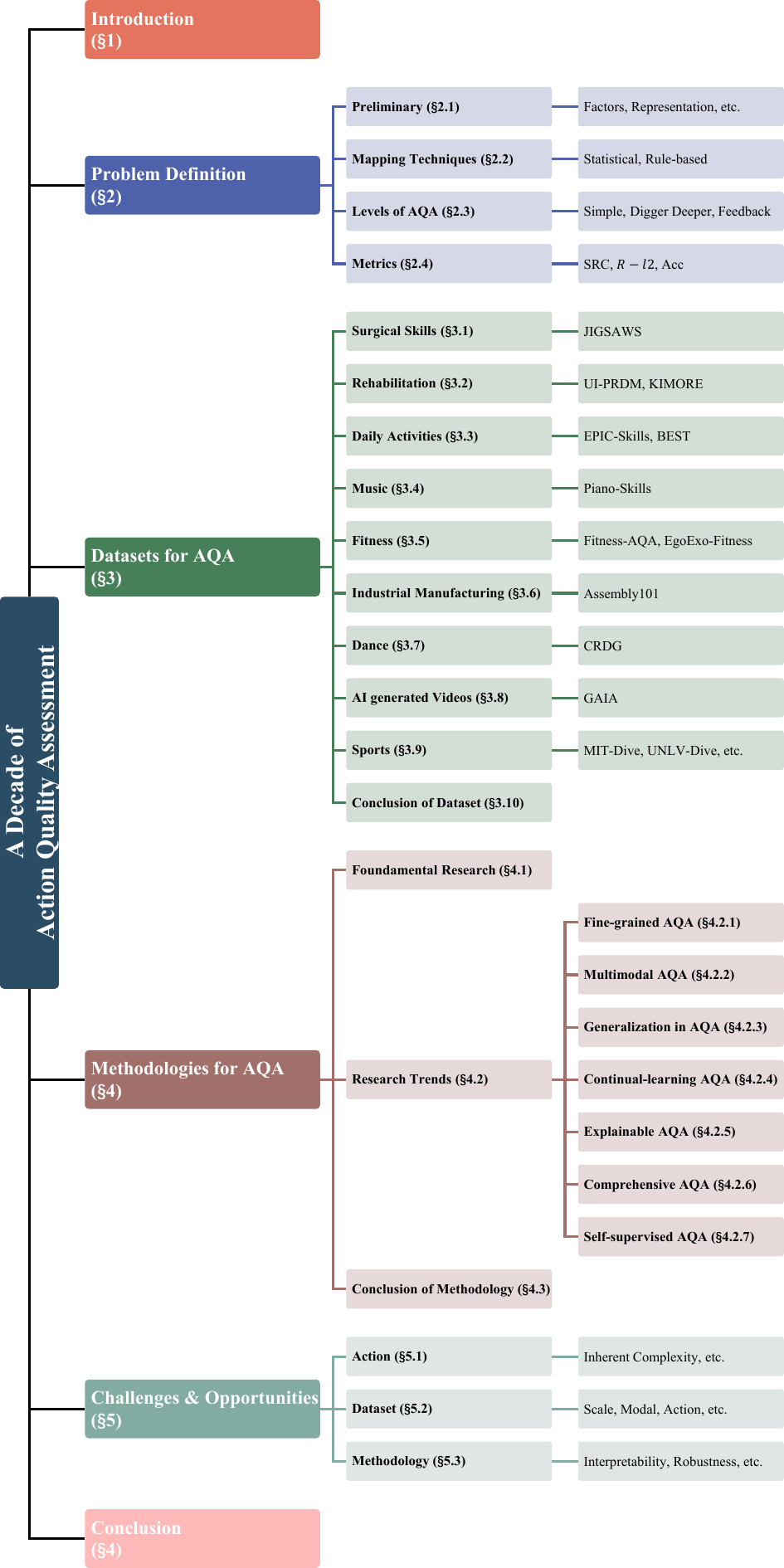}
    \caption{\textbf{Structure of the overall survey.} Zoom in for the best view.}
    \label{fig_structure}
\end{figure}

The rest of the survey paper is organized as follows. To facilitate navigation of the survey's structure, \autoref{fig_structure} provides an overview of the sections and their titles. Section \ref{sec2} defines and classifies the AQA problem, introducing common model evaluation metrics. Section \ref{sec3} offers a compilation of popular AQA datasets, categorizing them by action scenarios and providing detailed descriptions. Section \ref{sec4} reviews fundamental AQA research, summarizing papers from the past decade and dividing them into 7 principal trends, with a detailed discussion of research methodologies and performance comparisons within each trend. Section \ref{sec5} analyzes ongoing challenges in current research and proposes directions for future research. In conclusion, Section \ref{sec6} summarizes the findings of this survey.

\section{Problem Definition}\label{sec2}

\subsection{Preliminary}\label{subsec2.1}
\begin{figure}[]
    \centering
    \includegraphics[width=\columnwidth]{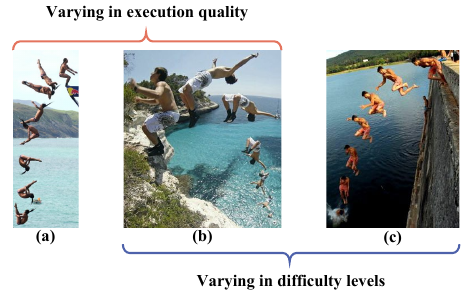} 
    \caption{\textbf{Execution quality vs difficulty level.} A professional cliff diver is shown doing a backflip in (a). Notice how straight his legs are, how crisp his overall form is, and how graceful his execution is, compared to an amateur diving in (b). Amateur's legs are bent at the knees, as rotating with bent legs would result in faster rotation (lesser effort on the diver's part). It generally requires a lot of training and practice to be able to attain and maintain a great form like (a). Now, we compare two amateurs (b) and (c). Even though both have bad forms, doing a backflip, as in (b), is more difficult (requires more effort and skills) than simply falling into the lake like (c).}
    \label{fig_exe}
\end{figure}

\noindent\textbf{Action and its Quality.} First, let's develop some intuition about action quality. Action can be thought of as having two components: \textit{difficulty level} (what action was done), and \textit{execution quality} (how well that action was done) \cite{parmar2019and}. An example of difficulty level is that somersaulting in the Pike position (in sports like Diving or Gymnastic Vault) is more difficult than somersaulting in the Tuck position. So, if executed perfectly, somersaulting in the Pike position is of more worth than somersaulting in the Tuck position. An example of intuiting about execution quality: In the Pike position, having feet together is worthy of more points than having feet apart or crossed over. The action quality score is directly proportional to both difficulty level and execution quality. To give a better intuition regarding execution quality and difficulty level, we have contrasted them in \autoref{fig_exe}. 

\begin{figure*}[]
    \centering
    \includegraphics[width=\textwidth]{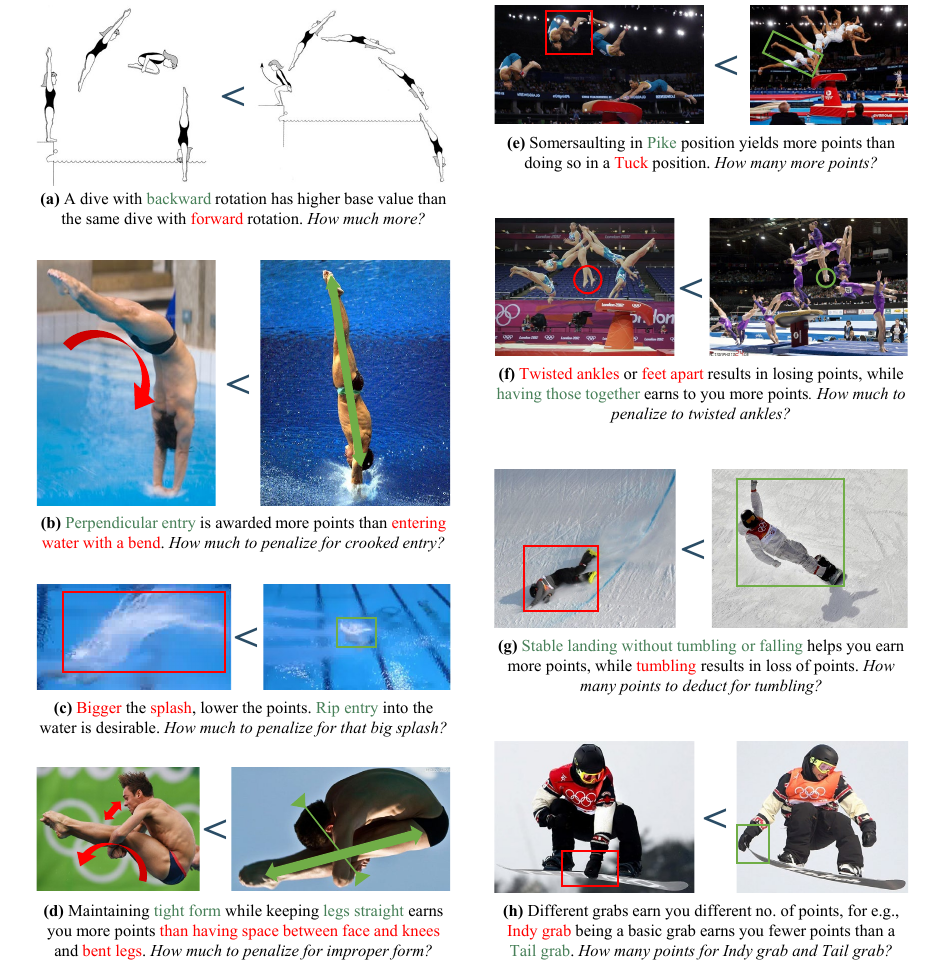} 
    \caption{\textbf{Examples of elements that matter in AQA.} Zoom in for the best view.}
    \label{fig_elements}
\end{figure*}

\noindent\textbf{Factors to be captured by representations.} In \autoref{fig_elements}, we illustrate some instances of action elements that matter in AQA. We mentioned what elements are desirable or hold a higher value. The task of AQA involves not only identifying those elements in a given action sequence but also determining/quantifying their value. E.g., let's consider \autoref{fig_elements}(f); we see that the feet of the gymnast on the left are not together. A judge judging that gymnast's performance would identify that her feet were not together; secondly, they would determine how much to penalize (in terms of the points) for that error--- the larger the gap, the more would be the penalty. Following this example, we can see that a CNN will have to learn to capture elements like these, determine the severity of the error, and penalize accordingly. Moreover, not all the elements may be equally important. Note that this was for just an element, usually, an action instance is composed of multiple such elements; each of those elements needs to be taken into consideration, based on which the final action quality score is calculated. 

\noindent\textbf{Representation formats.} Several kinds of representations are used to capture actors' movements and other relevant factors, such as raw RGB video \cite{pirsiavash2014assessing, parmar2017learning}, optical flow\cite{wang2024cpr, zeng2024multimodal}, skeleton pose sequence \cite{ogata2019temporal, li2021skeleton, okamoto2024hierarchical}, sound \cite{parmar2021piano, zeng2024multimodal, hipiny2023danced}, human-object distance \cite{okamoto2024hierarchical}, etc.

\begin{figure}[h]
    \centering
    \includegraphics[width=\columnwidth]{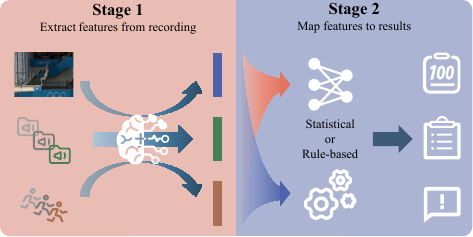} 
    \caption{\textbf{The generic AQA pipeline.}}
    \label{fig_pipeline}
\end{figure}

\noindent\textbf{Generic AQA pipeline.} Typically, AQA methodologies follow a two-stage approach \cite{parmar2017learning} (see \autoref{fig_pipeline}): Stage 1: feature extractors are used to extract relevant features from performance recording (video, audio, etc.); Stage 2: these features are then mapped to generate AQA scores, reports, feedback using statistical or rules-based mapping modules (discussed in the following). AQA research includes improving both these stages.

\subsection{Mapping Techniques}\label{subsec2.2}
\begin{figure*}[]
    \centering
    \includegraphics[width=\textwidth]{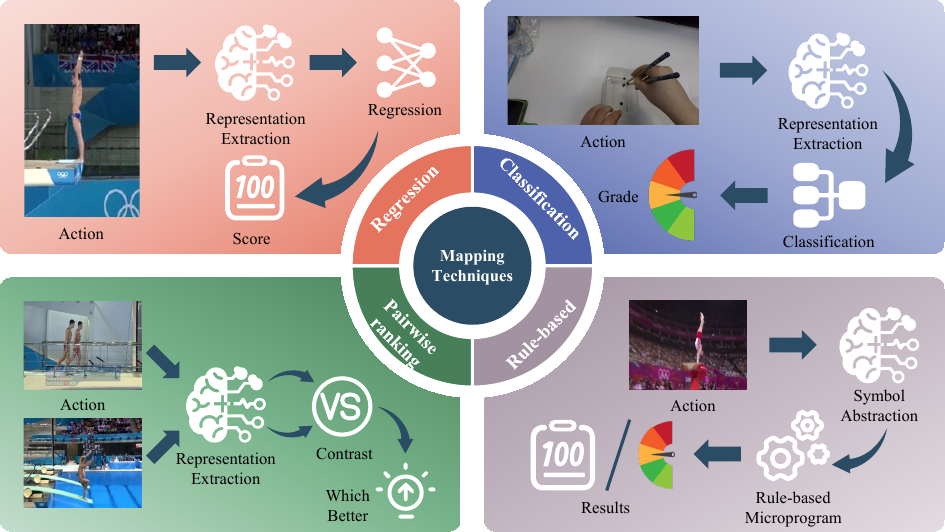} 
    \caption{\centering \textbf{Different formats of mapping techniques used in AQA.}}
    \label{fig_mapping}
\end{figure*}

Mapping from (visual/multimodal) action representation space to output score space can be Statistical or Rules-based in nature (see \autoref{fig_mapping}), discussed in the following.

\noindent\textbf{Statistical methodology.} In this methodology, models learn to estimate the relationship between the visual action representations/features and the action quality score from datasets in a supervised manner. Mapping model parameters are adjusted during training to fit the dataset. Note that while the weights can be adjusted in a data-driven manner, control over what patterns the mapping model fits on is not entirely in our control. Hence, the model might fit shortcut correlations, which might not be reflective of the true (intended) relation between action features and action scores. There are following three formats of statistical methodologies used: 
\begin{enumerate}[wide, labelwidth=!, labelindent=0pt,topsep=0pt]
    \item \textbf{Regression-based scoring.} Here, the statistical model predicts a numerical score on a continuous scale. In regression, the output score is a real number, and the model tries to find a relationship between the input features and this continuous output. For example, the score can be any real quantity between, let's say, 0 to 100. Models include linear regressors, shallow/deep neural nets, support vector regressors (SVR), etc. Regression allows or is used for fine-grained or detailed analysis of action quality. Examples of regression-based methodologies include \cite{pirsiavash2014assessing, parmar2017learning}.
    \item \textbf{Classification-based scoring.} In this case, the model predicts a discrete label or category. In classification, the output action quality score/label is a category or class (not a continuous value), and the model assigns input features to one of these classes. Classification is generally used for coarsely quantifying the action quality. Sometimes this coarse nature can allow in recruiting non-experts for data labeling. Examples of classification-based methodologies include \cite{parmar2016measuring, parmar2021piano, parmar2022domain, bruce2024egcn++}.
    \item \textbf{Pairwise ranking.} In this case, models compare two or more samples to decide which sample is of the highest action quality. Note that this is not a standalone analysis like the regression and classification. Examples of pairwise ranking methodologies include \cite{bertasius2017baller, doughty2018s, doughty2019pros}.
\end{enumerate}

\noindent\textbf{Rules-based scoring.} Unlike statistical methodologies, in rules-based methodologies, features/symbols are processed using a set of rules to compute the action quality score. A set of rules can be hand-crafted incorporating human expert knowledge, or they could be learned from data. In rules-based methodologies, how each individual feature of the performance affected the score can be determined/traced back. As such, the causal nature of the rules makes rules-based interpretable and explainable. Examples of rules-based methodologies include \cite{okamoto2024hierarchical}.

\subsection{Levels of AQA}\label{subsec2.3}
\noindent\textbf{Simple AQA.} In its simplest terms, AQA involves using a model for mapping an action video to a performance/action quality score. Ideally, the idea is that the model will extract all the relevant (spatiotemporal) features that contribute to action quality (score). Values are associated with these extracted features. Models learn to extract relevant action quality features and their values and, based on all the values, output a final score. 

\noindent\textbf{Digging Deeper.} A deeper level of assessment is that the model lists out all the factors and their analysis for what the diver did well, where the diver made a mistake, assessing the severities of these errors and then computing the final score based on it.

\noindent\textbf{Feedback Generation.} Here, the model is able to assess the performance and its good and bad points. Based on this, the model may be able to suggest what parts the actor can work on to improve. Generally, if the model has a level of transparency--has an internal record of the pros and cons of the performance/action, it should be able to suggest a preliminary level of feedback. For example, if the model identifies that a diver had bent their legs in the pike position, it may indicate that the diver may work on keeping their legs during the pike position to improve their score.

\subsection{Metrics}\label{subsec2.4}
The following metrics are used to quantify/measure the performance of AQA models. The metric is chosen depending on the type of mapping model used. For example, correlation between ground truth and predicted scores is used when scores are on a continuous scale, while accuracy is used when action quality labels are discrete.

\noindent\textbf{Spearman's Rank Correlation.} Pirsiavash et al.\cite{pirsiavash2014assessing} first pioneered the assessment of action quality in 2014 and also first introduced the use of Spearman's Rank Correlation (SRC, denoted as $\rho$) to evaluate the performance of regression models. The SRC formula is as follows:
\begin{equation}
    \rho=1-\frac{6\sum_{i=1}^n(R_i-\widehat{R_i})^2}{n(n^2-1)}
\end{equation}
where $R_i$ and $\widehat{R}_i$ represent the ground truth and predicted rankings of the $i$-th sample, respectively. $n$ is the total number of samples. The $\rho$ range is $[-1,1]$, with values closer to 1 indicating better performance. SRC is currently the most widely used performance metric in AQA. However, SRC can only measure the strength and direction of the monotonic relationship between ground truth and predicted rankings without measuring the differences between ground truth and predicted scores. 

\noindent\textbf{Relative $l2$ Distance.} To address this limitation, Yu at al.\cite{yu2021group} proposed Relative $l2$ Distance ($R-l2$) as a new metric for evaluating model performance. In contrast to SRC, which focuses on the ranking of predicted scores, $R-l2$ places more emphasis on the numerical values of the predicted score. Moreover, compared to the traditional $l2$ distance, $R-l2$ takes into account the score intervals between different categories of actions, facilitating cross-category model training. The $R-l2$ formula is as follows:
\begin{equation}
    R-l2=\frac{1}{n}\sum_{i=1}^n(\frac{\vert s_i-\widehat{s_i} \vert}{s_{max}-s_{min}})^2
\end{equation}
where $s_i$ and $\widehat{s_i}$ represent the ground truth and predicted scores of the $i$-th sample. $s_{max}$ and $s_{min}$ represent the maximum and minimum scores of this action category. $n$ is the total number of samples. The $R-l2$ range is $[0,1]$, closer to 0 indicating better performance.

\noindent\textbf{Accuracy.} For other mapping technique formats, pairwise rank and classification, accuracy (Acc) are commonly used as model performance metrics. The accuracy formula is as follows:
\begin{equation}
    Acc=\frac{TP+TN}{TP+TN+FP+FN}
\end{equation}
where TP, TN, FP, and FN represent true positives, true negatives, false positives, and false negatives, respectively.

\section{Datasets for AQA}\label{sec3}

\begin{figure*}[h]
    \centering
    \includegraphics[width=\textwidth]{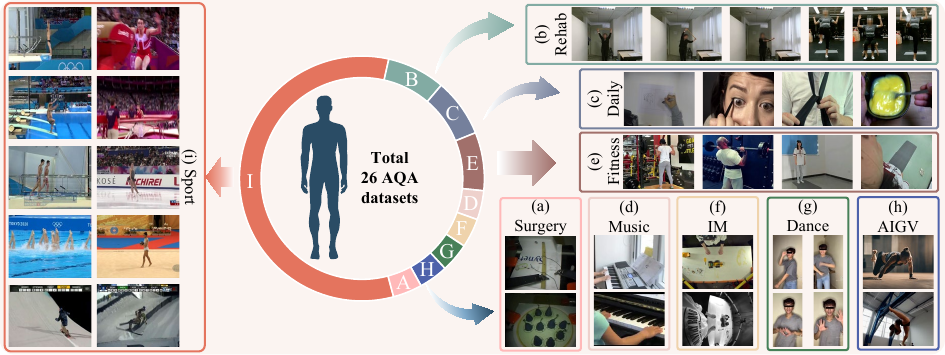}
    \caption{\textbf{Action samples of different dataset domains.} Zoom in for the best view.}
    \label{fig_dataset}
\end{figure*}

\begin{figure*}[!h]
    \centering
    \includegraphics[width=\textwidth]{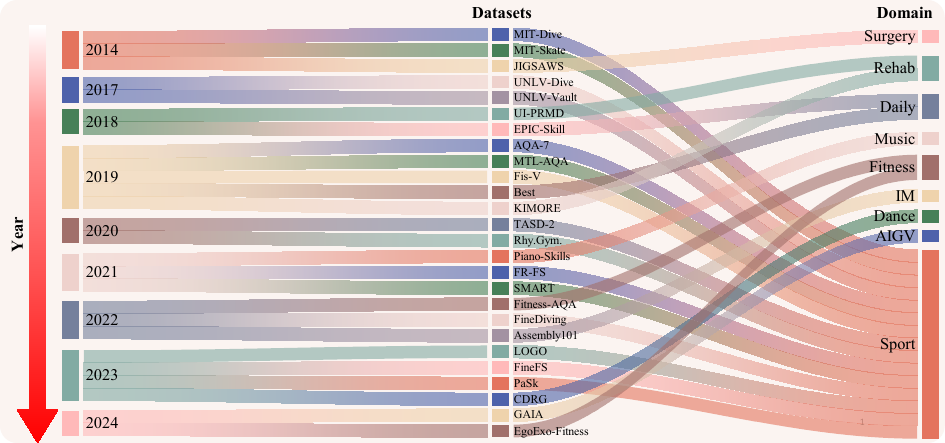}
    \caption{\textbf{An intuitive visualization of the annual distribution of publicly available datasets, including their titles and corresponding domains.}}
    \label{fig_sankey}
\end{figure*}

Datasets are a crucial component of machine learning projects, and they play a key role in AQA by enabling model training, performance evaluation, and real-world deployment. The success and advancements in AQA can be attributed to the development of novel and high-quality datasets. In this section, we systematically collect and summarize all existing datasets since 2014. Following the PRISMA guidelines, we identify 26 publicly available datasets related to skills and action quality assessment across various domains. We classify datasets into 9 domains \{Surgery, Daily Activities, Rehabilitation, Music, Fitness, Industrial Manufacturing, Dance, AIGV, Sports\} as illustrated in \autoref{fig_dataset}. These datasets are summarized in \autoref{table1} and \autoref{fig_sankey}, and we provide a more detailed description of each dataset in the following. We have organized the datasets by domain. Finally, we conclude the section by providing a summary of the datasets.

\begin{table*}[!ht]
\small
    \centering
    \caption{List of publicly available datasets used in skills and action quality assessment. AD: Average Duration, RV: Real Videos, AG: AI-generated, Desc: Description, Rehab: Rehabilitation, IM: Industrial manufacturing. *-indicates only the no. of coarse action types---no. of finegrained action types contained is much higher.} 
    \rowcolors{2}{gray!15}{white}
    \begin{tabularx}{\textwidth}{lcccrXcc}
        \toprule [1.5pt]
        \rowcolor{white} 
        \textbf{Datasets} & \textbf{Year} & \textbf{Sample} & \textbf{Type} & \textbf{AD} & \textbf{Annotation Type} & \textbf{Source} & \textbf{Domain}   \\
        \midrule [1pt]
        MIT-Dive\cite{pirsiavash2014assessing}       & 2014 & 159  & 1  & 2.5s   & Score                    & RV & \textcolor[RGB]{228, 116, 94}{Sport}    \\
        MIT-Skate\cite{pirsiavash2014assessing}      & 2014 & 150  & 1  & 175s   & Score                    & RV & \textcolor[RGB]{228, 116, 94}{Sport}    \\
        JIGSAWS\cite{gao2014jhu}                     & 2014 & 103  & 3  & 92s    & Score, Action            & RV & \textcolor[RGB]{255, 185, 185}{Surgery} \\
        UNLV-Dive\cite{parmar2017learning}           & 2017 & 370  & 1  & 3.8s   & Score                    & RV & \textcolor[RGB]{228, 116, 94}{Sport}    \\
        UNLV-Vault\cite{parmar2017learning}          & 2017 & 176  & 1  & 2.8s   & Score                    & RV & \textcolor[RGB]{228, 116, 94}{Sport}    \\
        UI-PRMD\cite{vakanski2018data}               & 2018 & 100  & 10 & ---    & Grade, Action            & RV & \textcolor[RGB]{117, 128, 156}{Rehab}   \\
        EPIC-Skill\cite{doughty2018s}                & 2018 & 216  & 4  & 86.6s  & Rankings, Action         & RV & \textcolor[RGB]{130, 171, 163}{Daily}   \\
        AQA-7\cite{parmar2019action}                 & 2019 & 1189 & 7*  & 6.7s   & Score, Action            & RV & \textcolor[RGB]{228, 116, 94}{Sport}    \\
        MTL-AQA\cite{parmar2019and}                  & 2019 & 1412 & 52  & 4.1s   & Score, Action, Desc      & RV & \textcolor[RGB]{228, 116, 94}{Sport}    \\
        Fis-V\cite{xu2019learning}                   & 2019 & 500  & 1  & 170s   & Score                    & RV & \textcolor[RGB]{228, 116, 94}{Sport}    \\
        Best\cite{doughty2019pros}                   & 2019 & 500  & 5  & 187.6s & Grade, Rankings, Action  & RV & \textcolor[RGB]{130, 171, 163}{Daily}   \\
        KIMORE\cite{capecci2019kimore}               & 2019 & 353  & 5  & 29.9s  & Score, Action            & RV & \textcolor[RGB]{117, 128, 156}{Rehab}   \\
        TASD-2\cite{gao2020asymmetric}               & 2020 & 606  & 2  & 4.1s   & Score, Action            & RV & \textcolor[RGB]{228, 116, 94}{Sport}    \\
        Rhy.Gym.\cite{zeng2020hybrid}                & 2020 & 1000 & 4  & 95s    & Score, Action            & RV & \textcolor[RGB]{228, 116, 94}{Sport}    \\
        Piano-Skills\cite{parmar2021piano}           & 2021 & 992  & 1  & 160fr  & Grade, Song-difficulty   & RV & \textcolor[RGB]{238, 209, 204}{Music}   \\
        FR-FS\cite{wang2021tsa}                      & 2021 & 417  & 1  & 103fr  & Grade, Action            & RV & \textcolor[RGB]{228, 116, 94}{Sport}    \\
        SMART\cite{chen2021sportscap}                & 2021 & 5000 & 10 & 420fr  & Score, Action            & RV & \textcolor[RGB]{228, 116, 94}{Sport}    \\
        Fitness-AQA\cite{parmar2022domain}           & 2022 & 21284& 3  & 4.1s   & Grade, Action            & RV & \textcolor[RGB]{161, 112, 107}{Fitness} \\
        FineDiving\cite{xu2022finediving}            & 2022 & 3000 & 52 & 4.2s   & Score, Action            & RV & \textcolor[RGB]{228, 116, 94}{Sport}    \\
        Assembly101\cite{sener2022assembly101}       & 2022 & 4321 & 101& 426s   & Score, Action            & RV & \textcolor[RGB]{239, 211, 172}{IM}      \\
        LOGO\cite{zhang2023logo}                     & 2023 & 200  & 12 & 204.2s & Score, Action, Formation & RV & \textcolor[RGB]{228, 116, 94}{Sport}    \\
        FineFS\cite{ji2023localization}              & 2023 & 1167 & 1  & 215s   & Score, Action            & RV & \textcolor[RGB]{228, 116, 94}{Sport}    \\ 
        PaSk\cite{gao2023automatic}                  & 2023 & 1018 & 1  & 10.7s  & Score                    & RV & \textcolor[RGB]{228, 116, 94}{Sport}    \\
        CDRG\cite{hipiny2023danced}                  & 2023 & 240  & 12 & 14.7s  & Rankings, Action         & RV & \textcolor[RGB]{71, 128, 88}{Dance}     \\
        GAIA\cite{chen2024gaia}                      & 2024 & 9180 & 510& 2.8s   & Score, Action            & AG & \textcolor[RGB]{78, 98, 171}{AIGV}       \\
        EgoExo-Fitness\cite{li2024egoexo}            & 2024 & 6131 & 12 & 18.8s  & Score, Action, Desc      & RV & \textcolor[RGB]{161, 112, 107}{Fitness} \\
        \bottomrule [1.5pt]
    \end{tabularx}
    \label{table1}
\end{table*}

\subsection{Surgical Skills}\label{subsubsec3.1}
\textbf{JIGSAWS}, proposed by Gao et al.\cite{gao2014jhu} at MICCAI Workshop 2014, is the first surgical skill dataset for human action modeling. The dataset contains 103 samples with an average sample duration of 92 seconds. Three basic surgical operations were captured by the daVinci Surgical System from eight surgeons with different skill levels. Each operation was repeated five times, and the specific actions are illustrated in \autoref{fig_dataset}(a). There are 39 suturing samples, 36 knot-tying samples, and 28 needle-passing samples. Each sample consists of 76 dimensions of kinematic data (including cartesian positions, orientations, velocities, angular velocities, and gripper angle describing the motion of the manipulator) and video data. The dataset annotations contain action labels and action scores, with a total of 15 action labels corresponding to video frames, and action scores are the total scores of the six sub-elements, with a score interval of 1-5 for each element.

\subsection{Rehabilitation}\label{subsubsec3.2}
The datasets of rehabilitation contain UI-PRMD and KIMORE, which were proposed by Vakanski et al.\cite{vakanski2018data} at Data 2018 and Capecci et al.\cite{capecci2019kimore} at TNSRE 2019, respectively. Detailed actions are illustrated in \autoref{fig_dataset}(b).

\noindent\textbf{UI-PRMD}, fills the gap of a comprehensive dataset of physiotherapy actions. The dataset contains 100 samples of 10 different actions, each repeated 10 times, from 10 healthy subjects simultaneously captured by two systems, Vicon and Kinect. Each sample consists of skeletal sequences, including position and angle. The dataset annotations are correct or incorrect binary-grade labels.

\noindent\textbf{KIMORE}, is the only RGB single-view rehabilitation action dataset. The dataset contains 353 samples of 5 different actions from 44 healthy subjects and 34 patients, with an average sample duration of 29.9 seconds. The dataset was annotated with scores, including $PO_S$ and $CF_S$, with values in the range 0 to 50 for each action as defined by clinicians.

\subsection{Daily Activities}\label{subsubsec3.3}
The datasets of daily activities scenarios contain EPIC-Skill\cite{doughty2018s} and BEST\cite{doughty2019pros}, which were proposed by Doughty et al. at CVPR 2018 and CVPR 2019, respectively. Detailed actions are illustrated in \autoref{fig_dataset}(c).

\noindent\textbf{EPIC-Skill} contains 216 samples divided into 4 sub-datasets: surgery, dough-rolling, drawing, and chopstick-using, with an average sample duration of 86.6 seconds. The surgery sub-dataset is JIGSAWS\cite{gao2014jhu}, the Dough-Rolling sub-dataset was 33 samples selected from Kitchen-based CMU-MMAC\cite{frade2008mmac}, the Drawing sub-dataset contains videos of 4 volunteers repeating the drawing of SONIC and HAND 5 times respectively, and the Chopstick-Using sub-dataset contains videos of 8 volunteers repeating the chucking of beans in a box 5 times. Since JIGSAWS\cite{gao2014jhu} was annotated, the other three sub-datasets and rankings were obtained by pairwise comparison of samples within classes.

\noindent\textbf{BEST} contains 500 samples consisting of 5 daily skill tasks: scrambling eggs, braiding hair, tying a tie, making an origami crane, and applying eyeliner, with an average sample duration of 187.6 seconds. Videos are related action videos downloaded from YouTube. Dataset annotations contain the start and end times of actions in the videos, action grades (Beginner, Intermediate, or Expert), and rankings.

\noindent\textbf{EgoExo4D} \cite{grauman2024ego} is a dataset for skills assessment for daily everyday and sports activities. It contains 1224 samples with a total of 16 hours of video.

\subsection{Music}\label{subsubsec3.4}
\textbf{Piano-Skills}, proposed by Parmar et al.\cite{parmar2021piano} at MMSP 2021, is the first AQA dataset in music and introduced audio into consideration. The dataset contains 992 samples from 61 piano performances, with an average sample duration of 160 frames, as illustrated in \autoref{fig_dataset}(d). The dataset was annotated by a trained pianist with skill levels and song difficulty.

\subsection{Fitness}\label{subsubsec3.5}
\noindent\textbf{Fitness-AQA}, proposed by Parmar et al.\cite{parmar2022domain} at ECCV 2022, is the dataset with the most samples in AQA and the first in fitness. The dataset contains 21284 samples collected from video-sharing sites, covering BackSquat, BarbellRow, and Overhead Press, with an average sample duration of 4.1 seconds, as illustrated in \autoref{fig_dataset}(e). The dataset annotations contain binary-grade labels by two professional gym trainers.

\noindent\textbf{EgoExo-Fitness}, proposed by Li et al.\cite{li2024egoexo} at ECCV 2024. The dataset contains 6131 samples from 86 action sequences by combining 3 to 6 different actions of 12 types of fitness actions, constructed by the synchronized recording of 3 egocentric-view and 3 fixed exocentric-view (third-person) videos. The average duration is about 18.8 seconds. The dataset annotations contain technical key point verification of fitness action, natural language comment on subjects' action, and action quality score (range 1-5 points), which provide interpretable action judgment annotations.

\subsection{Industrial Manufacturing}\label{subsubsec3.6}
\textbf{Assembly101}, proposed by Sener et al.\cite{sener2022assembly101} at CVPR 2022, is currently the data set with most camera views in AQA. The dataset contains 4321 samples, constructed by capturing 362 different volunteers assembling and disassembling 101 different toy cars with 8 fixed and 4 egocentric cameras, as illustrated in \autoref{fig_dataset}(c). The average sample duration of 426 seconds. The dataset annotations contain over 1 million action segments, 1380 fine-grained and 202 coarse-grained action classes, and action scores.

\subsection{Dance}\label{subsubsec3.7}
\textbf{CDRG}, proposed by Hipiny et al.\cite{hipiny2023danced} in IJAIN 2023. \autoref{fig_dataset}(g). The dataset contains 240 samples from 20 subjects performing 12 TikTok dance challenges respectively. Subjects used their own devices to record videos, resulting in different levels of image quality. The average duration of samples is 14.7 seconds. The dataset was annotated by 100 annotators to mark the winning one of each dance pair.

\subsection{AI Generated Videos}\label{subsubsec3.8}
\textbf{GAIA}, proposed by Chen et al.\cite{chen2024gaia} at NeurlIPS 2024, is the first AQA dataset constructed by collecting videos generated by text-to-video (T2V) models to assess their video generation quality. The dataset contains 9180 samples generated by T2V models from 18 different laboratories and commercial platforms, covering a total of 510 actions of the whole body, hand, and face, as illustrated in \autoref{fig_dataset}(h), with an average sample duration of 2.8 seconds. The dataset was annotated as scores of the generated videos from three perspectives, including subject quality, action completeness, and action-scene interaction, with a score interval of 0-100 for each perspective.

\subsection{Sports}\label{subsubsec3.9}
Derived from their clear rules and easy access, sports account for the highest percentage of AQA datasets, including 15 datasets covering 10 different types, as illustrated in \autoref{fig_dataset}(i). In recent years, the advancement of sports AQA datasets has been remarkable, with the publication of numerous high-quality datasets providing important benchmarks for AQA. The rest of this section provides a detailed description of sports datasets.

\noindent\textbf{MIT-Dive \& MIT-Skate}, proposed by Pirsiavash et al.\cite{pirsiavash2014assessing} at ECCV 2014. MIT-Dive dataset contains 159 samples of Olympic diving, all in slow motion from television broadcasts at a frame rate of 60 fps. The average duration of dive samples is 2.5 seconds. MIT-Dive dataset was annotated with scores awarded by referees, ranging from 20 to 100 points. Similarly, MIT-Skate dataset consists of 150 skating samples, with a frame rate of 24 fps. The average duration of skate samples is 175 seconds. The dataset was also annotated with awarded scores, ranging from 0 to 100 points.

\noindent\textbf{UNLV-Dive \& UNLV-Vault}, proposed by Parmar and Morris et al.\cite{parmar2017learning} at CVPR Workshop 2017. UNLV-Dive contains 370 samples with an average sample length of 3.8 seconds and is an extension of the original MIT-Dive dataset. UNLV-Dive was annotated with scores calculated by multiplying the execution score (range 0-30 points) by the difficulty score (no upper limit specified in the rules, range 2.7-4.1 points in the dataset), resulting in a final dataset score. Similarly, UNLV-Vault contains 176 samples, with a sample average length of 2.8 s. UNLV-Vault was annotated with scores, which were calculated by adding the execution score (range 0-10 points) and the difficulty score (range 0-10 points), resulting in a final score.

\noindent\textbf{AQA-7}, proposed by Parmar and Morris et al.\cite{parmar2019action} at WACV 2017. The dataset contains 1189 samples from both the Winter and Summer Olympics, covering 7 different action types (including singles diving-10m platform, gymnastic vault, big air skiing, big air snowboarding, synchronous diving-3m springboard, synchronous diving-10m platform, and trampoline). The average duration of samples is 6.7 seconds. The dataset was annotated with scores specific to each type, reflecting the unique scoring system. For each type, the scoring intervals are as follows: singles diving-10m platform (21.6-102.6), gymnastics vault (12.3-16.87), big air skiing (8-50), big air snowboarding (8-50), synchronized diving 3m springboard (46.2-104.88), synchronized diving 10m platform (49.8-99.36) and trampoline (6.72-62.99).

\noindent\textbf{MTL-AQA}, proposed by Parmar and Morris et al.\cite{parmar2019and} at CVPR 2019. The dataset contains 1412 samples from 16 different competitions, including 10m platform and 3m springboard, male and female athletes, individual or pairs of synchronized divers, and different views. It also pioneered the: 1) natural language detailed qualitative description of the performance; and 2) disassembling (diving) actions into finegrained subcategories, divided into five parts: position, armstand, rotation type, somersaults, and twists. The average duration of samples is 4.1 seconds. The dataset annotations include the difficulty and action scores awarded by seven referees, along with the finegrained action breakdown and detailed qualitative descriptions of the corresponding actions.

\noindent\textbf{Fis-V}, proposed by Xu et al.\cite{xu2019learning} at TCSVT 2019. The dataset contains 500 samples of selected women's singles figure skating short program videos. There are 149 athletes from 20 countries, and the average duration of samples is 170 seconds, with an average frame rate of 25 FPS. The dataset annotations include the total element score (TES) and total program component score (PCS), each awarded by 9 different professional figure skating referees.

\noindent\textbf{TASD-2}, proposed by Gao et al.\cite{gao2020asymmetric} at ECCV 2020, different from previous datasets containing synchronized diving with side view, uses front view of video shot. The dataset contains 606 samples with an average duration of 4.1 seconds. The dataset was annotated with the beginning and end frames of the action as well as the final score, which is the product of the execution score and the difficulty score multiplied by the execution score.

\noindent\textbf{Rhythmic Gymnastics}, proposed by Zeng et al.\cite{zeng2020hybrid} at ACM MM 2020. The dataset contains 1000 samples from the 36th and 37th International Artistic Gymnastics Competitions, encompassing the four types of ball, clubs, hoop, and ribbon. Each type has 250 sample videos, with an average duration of 95 seconds and an average frame rate of 25 FPS. The dataset was annotated with scores, including a difficulty score, an execution score, and a total score, which is the sum of the execution score and difficulty score.

\noindent\textbf{FR-FS}, proposed by Wang et al.\cite{wang2021tsa} at ACM MM 2020, unlike previous figure skating AQA dataset with duration (close to 3 minutes), this contains frame-level information and plans to gradually construct a more fine-grained figure skating AQA system, starting with the identification of the most basic errors. The dataset contains 417 samples from the Fis-V dataset and the PyeongChang Winter Olympics, including multiple athletes' key actions (take-off, rotation, and landing). Of these, 276 are smooth landing videos and 141 are fall videos. The dataset was annotated with key action and binary-grade labels.

\noindent\textbf{SMART}, proposed by Chen et al.\cite{chen2021sportscap} at IJCV 2021. The dataset contains 5000 samples of 10 different action types, including balance beam, diving, uneven bars, vault, hurdling, pole vault, high jump, boxing, keep-fit, and badminton, with an average sample duration of 420 frames. The dataset annotations contain sub-action labels and action assessment scores.

\noindent\textbf{FineDiving}, proposed by Xu et al.\cite{xu2022finediving} at CVPR 2022. The dataset contains 3000 samples from 30 different diving competitions, including the Olympics, World Cup, World Championships, and European Aquatics Championships, and encompasses 52 action types, 29 sub-action types, and 23 difficulty levels. The average duration is 4.2 seconds. The dataset annotations contain semantic and temporal structures, both of which are two-level annotations. In the semantic structure, the action-level label describes the action category, and the step-level label describes the sub-action types of the successive actions. In the temporal structure, the action-level label describes the beginning and end time of the athlete's execution of the complete action, and the step-level label describes the beginning and end frames of the sub-action. The competition referees awarded the score. This dataset was extended in \cite{xu2024procedure}.

\noindent\textbf{LOGO}, proposed by Zhang et al.\cite{zhang2023logo} at CVPR 2023, was constructed as multi-person long video AQA dataset. The dataset contains 200 samples, with 8 athletes in each frame, from 26 artistic swimming events during 2018-2022, with an average sample duration of 204.2 seconds. The dataset annotations are structured in a temporal structure, including fine-grained action and formation annotations. There are 12 types of action annotation and 17 types of formation annotation. The annotations of both action and formation were annotated through frame-by-frame analysis. Scores were awarded by referees with 3 sub-scores and a total score.

\noindent\textbf{FineFS}, proposed by Ji et al.\cite{ji2023localization} at ACM MM 2023. The dataset contains 1167 samples containing RGB videos and skeleton sequences from 72 A-level international events under the new ISU rules from the 2018-2019 season to the 2021-2022 season, with an average sample duration of 215 seconds. The dataset annotations contain 4 levels of scores, 2 levels of sub-action categories, and time segmentation.

\noindent\textbf{PaSk}, proposed by Gao et al.\cite{gao2023automatic} at IJCV 2023, can be used to complement the AQA in a strong primary-secondary relation since the male performer always assists the female performer to execute various actions in pairs figure skating. The dataset contains 1018 samples from figure skating competitions organized by ISU, with an average sample duration of 10.7 seconds. The dataset annotations contain the beginning and end frames of the independent actions and the awarded scores.

\subsection{Conclusion of Dataset}\label{subsec3.10}
Since the fundamental research of the AQA dataset in 2014, 26 popular datasets have emerged within the AQA domain, encompassing a wide range of domains from surgery to sport. Of these, sport accounts for the largest proportion and is the most important application scenario of AQA. A number of researchers have made significant contributions to the construction of datasets. To illustrate, Parmar has proposed a total of 6 widely used datasets before and after\cite{parmar2017learning, parmar2019action, parmar2019and, parmar2021piano, parmar2022domain}, while Xu has proposed the most fine-grained diving action dataset\cite{xu2022finediving}. It is evident that the size of the dataset is continuously expanding, the action types and modality data are constantly being enriched, and the data annotation is moving towards fine-grained annotation, all of which are driving research progress. However, compared to action recognition datasets\cite{ionescu2013human3, kay2017kinetics} and other related datasets\cite{liu2020fsd, sardari2020vi, tang2020comprehensive, moodley2022casa, parmar2022win, xing2022functional, liu2023fine, tang2023flag3d, fang2024better, gan2024skatingverse, li2024finerehab, nagai2024mmw}
, AQA datasets still face challenges in terms of scale, action diversity, data modality, and annotations.

In particular, GAIA\cite{chen2024gaia}, as the first dataset constructed using AI-generated videos, marks a significant breakthrough. It not only broadens the application scenarios of AQA but also introduces AQA into the T2V model to assess the quality of video generation. GAIA also inspires researchers to explore the possibility of using the T2V model to generate massive videos with multiple action types and skill levels based on predefined prompts, thereby easily constructing large-scale AQA datasets.

\section{Methodologies for AQA}\label{sec4}
\begin{figure*}[]
    \centering
    \includegraphics[width=\textwidth]{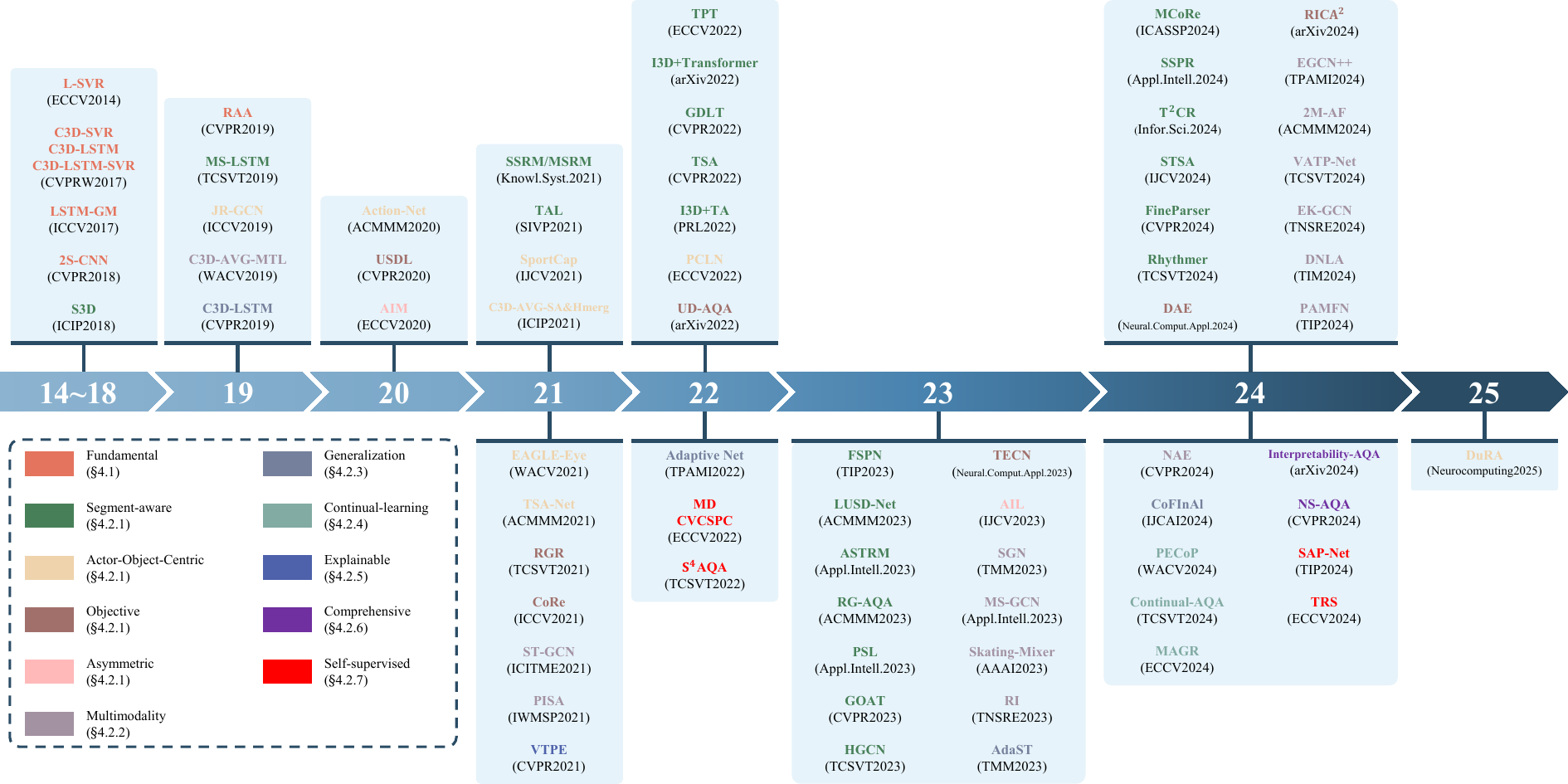}
    \caption{\textbf{Timeline of representative AQA methodologies from 2014 to 2024.} Zoom in for the best view.}
    \label{fig_timeline}
\end{figure*}

Following the comprehensive overview of publicly available datasets in AQA, it is evident that further investigation is required. What methodologies are being employed by researchers to utilize these rich resources of datasets to advance AQA fully? In order to answer this question, this section will concentrate on the methodologies employed in the field of AQA. By categorizing the pertinent and standardized research methodologies related to AQA in the last decade, it is possible to gain insight not only into the research paradigm in this field but also into the prevailing trends in research (see \autoref{fig_timeline}). The preceding decade of research has been classified into 7 principal trends, with comprehensive descriptions of the employed research methodologies and performance.

\subsection{Fundamental Research}\label{subsec4.1}

\begin{table*}[!htbp]
\small
    \caption{Performance comparison of fundamental research methodologies. FD: Frequency domain, TD: Time domain.}
    \centering
    \begin{tabular}{@{}lccccl@{}}
        \toprule [1.5pt]
        \textbf{Model} & \textbf{Year} & \textbf{Format} & \textbf{Metric} & \textbf{Performance} & \textbf{Datasets}\\ \midrule[1pt]
        \multirow{2}{*}{L-SVR\cite{pirsiavash2014assessing}} & \multirow{2}{*}{2014} & \multirow{2}{*}{Regression} & \multirow{2}{*}{SRC} & \multirow{1}{*}{0.4100} & \multirow{1}{*}{MIT-Dive} \\
        & & & & 0.4500 & MIT-Skate \\  \arrayrulecolor{Grey0!30} \midrule [0.5pt]

        PCA-SVM\cite{venkataraman2015dynamical} & 2015 & Regression & SRC & 0.4500 & MIT-Dive \\ \midrule [0.5pt]
        
        SVM (FD)\cite{parmar2016measuring} & 2016 & Classification & Acc & 0.8926 & LAM \\ \midrule [0.5pt]
        AdaBoosted-Tree (TD)\cite{parmar2016measuring} & 2016 & Classification & Acc & 0.9249 & LAM \\ \midrule [0.5pt]
        
        \multirow{2}{*}{C3D-SVR\cite{parmar2017learning}} & \multirow{2}{*}{2017} & \multirow{2}{*}{Regression} & \multirow{2}{*}{SRC} & \multirow{1}{*}{0.7800} & \multirow{1}{*}{UNLV-Dive} \\
        & & & & 0.6600 & UNLV-Vault \\  \arrayrulecolor{Grey0!30} \midrule [0.5pt]
        \multirow{2}{*}{C3D-LSTM\cite{parmar2017learning}} & \multirow{2}{*}{2017} & \multirow{2}{*}{Regression} & \multirow{2}{*}{SRC} & \multirow{1}{*}{0.3600} & \multirow{1}{*}{UNLV-Dive} \\
        & & & & 0.0500 & UNLV-Vault \\  \arrayrulecolor{Grey0!30} \midrule [0.5pt]
        \multirow{2}{*}{C3D-LSTM-SVR\cite{parmar2017learning}} & \multirow{2}{*}{2017} & \multirow{2}{*}{Regression} & \multirow{2}{*}{SRC} & \multirow{1}{*}{0.6600} & \multirow{1}{*}{UNLV-Dive} \\
        & & & & 0.3700 & UNLV-Vault \\  \arrayrulecolor{Grey0!30} \midrule [0.5pt]
        
        LSTM-GM\cite{bertasius2017baller} & 2017 & Pairwise Rank & Acc & 0.7650 & FP-Basketball \\  \arrayrulecolor{Grey0!30} \midrule [0.5pt]

        2S-CNN\cite{doughty2018s}         & 2018 & Pairwise Rank & Acc & 0.7607 & EPIC-Skill    \\  \arrayrulecolor{Grey0!30} \midrule [0.5pt]
        
        \multirow{2}{*}{RAA\cite{doughty2019pros}} & \multirow{2}{*}{2019} & \multirow{2}{*}{Pairwise Rank} & \multirow{2}{*}{Acc} & \multirow{1}{*}{0.8030} & \multirow{1}{*}{EPIC-Skill} \\
        & & & & 0.8120 & BEST \\
        
        \arrayrulecolor{black} \bottomrule [1.5pt]
    \end{tabular}
    \label{table2}
\end{table*}

To the best of our knowledge, AQA research dates back to 1995 when Gordon \cite{gordon1995automated} proposed to apply computer vision technologies such as tracking to AQA. There has been some research on AQA between 1995 and 2014, such as \cite{jug2003trajectory, wnuk2010analyzing, chang2011kinect}. However, the AQA field still lacked datasets to conduct reliable studies and evaluation metrics of AQA methodologies. Pirsiavash et al.\cite{pirsiavash2014assessing} and Gao et al.\cite{gao2014jhu} introduced the first AQA datasets. These include MIT-Dive, MIT-Skate, and JIGSAWS datasets. Pirsiavash et al.\cite{pirsiavash2014assessing} additionally proposed one of the initial machine learning frameworks for AQA by treating AQA as a supervised regression problem. Specifically, in their methodology, spatiotemporal features (pixel gradients and athlete pose features) were fed into a linear support vector regression (\textbf{L-SVR}) model. This methodology yielded an encouraging performance (see \autoref{table2}), while Venkataraman et al.\cite{venkataraman2015dynamical} achieved better performance one year later. Gao et al.\cite{gao2014jhu} mainly collected the surgical skills assessment dataset and employed a methodology proposed by Tao et al.\cite{tao2012hmm}---sparse dictionary learning (SDL) combined with hidden Markov model (HMM) to realize surgical gesture recognition. AQA methodologies thus far were based on traditional/shallow machine learning. 

Parmar et al. \cite{parmar2016measuring} introduced the classification paradigm to AQA by coarsely assessing the quality of user exercises and classifying them into erroneous vs. non-erroneous. They further tried several machine learning models on time-series and frequency-domain representations of human poses. Importantly, this work also concretely highlighted the problem of generalization of AQA models across human subjects.

Parmar and Morris et al.\cite{parmar2017learning} were the first to propose utilizing deep spatiotemporal convolutional features (C3D network \cite{tran2015c3d}) for AQA and proposed the UNLV-Dive and UNLV-Vault datasets based on the work of Pirsiavash et al.\cite{pirsiavash2014assessing}. We believe this work marks the beginning of new age of AQA and the increased interest in AQA from the computer vision community. This work proposed three different frameworks for automated AQA: \textbf{C3D+SVR}, \textbf{C3D+LSTM}, and \textbf{C3D+LSTM+SVR}, and achieved promising performance. These frameworks differed in the way they aggregated clip-level features to obtain global video-level information. Particularly, averaging and LSTM-based\cite{lstm} aggregation were explored. 

Their work significantly advanced AQA in terms of both performance and datasets, with most subsequent research referencing and following this feature extraction methodology. Especially the use of pre-trained networks on large-scale, labeled action recognition datasets\cite{joao2017i3d, carreira2018short}, such as C3D\cite{tran2015c3d}, Resnet\cite{he2016resnet}, I3D\cite{joao2017i3d}, P3D\cite{qiu2017p3d}, and VST\cite{liu2022vst}, as the backbone for feature extraction is very popular in AQA.

In the same year, Bertasius et al.\cite{bertasius2017baller} proposed the first-person basketball dataset and contrastive learning model, which represented an expansion of the mapping techniques from regression to pairwise ranking. This work employs convolutional LSTM and Gaussian mixture models (\textbf{LSTM-GM}) to generate highly nonlinear spatiotemporal features from atomic basketball events, then compute action quality by multiplying features with linear weights learned from the data. LSTM-GM could learn the evaluation criteria and pairwise ranking from pairs of weakly labeled first-person basketball videos.

Doughty et al.\cite{doughty2018s} proposed the first skills assessment dataset for daily life activities, EPIC-Skills, in 2018. Additionally, Doughty\cite{doughty2018s} proposed a contrastive ranking model that utilizes temporal and spatial segment networks (\textbf{2S-CNN}) in combination with video segment ranking loss and similarity loss function to achieve high accuracy pairwise ranking. In 2019, Doughty et al.\cite{doughty2019pros, doughty2021skill} also proposed a rank-aware attention model (\textbf{RAA}), in which a dual attention mechanism is used to focus on the pros and cons of action performance in long video clips and assign different weights to the clips to fuse the data. The model is trained using the combination of ranking loss, disparity loss, rank-aware loss, and diversity loss. This methodology enables accurate skill level assessment in video segments with different skill performances.

\subsection{Research Trends}\label{subsec4.2}
The fundamental research\cite{pirsiavash2014assessing, venkataraman2015dynamical, parmar2016measuring, parmar2017learning, bertasius2017baller, doughty2018s, doughty2019pros} established the formats of task and application scenarios of AQA, offering datasets and references for subsequent research. As computing capability and deep learning technology advanced, AQA has attracted more attention, giving rise to new methodologies. These novel methodologies consistently enhance model performance and exhibit disparate research trends, which can be categorized into multimodal methodologies, generalization, continual learning, explainable, and comprehensive AQA.

\subsubsection{Fine-grained AQA}\label{subsubsec4.2.1}
The rise of fine-grained research as the principal trend in AQA can be attributed to its capacity to address pivotal issues such as the neglect of details, the influence of noise, and low interpretability, which were prevalent in early studies. The advancements in computing capability and deep learning technology have made fine-grained research not only feasible but also mainstream. However, researchers have explored various methodologies to achieve fine-grain. To provide a more comprehensive explanation of fine-grained research, we have divided it into four sub-trends, which are elaborated respectively in the following sections.

\noindent\textbf{(1) Segment-aware feature extraction}

\begin{figure}[h]
    \centering
    \includegraphics[width=\columnwidth]{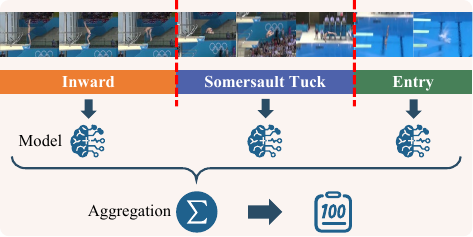}
    \caption{\textbf{Brief illustration of segment-aware feature extraction.}}
    \label{fig_trend_segment}
\end{figure}

Action maybe broken down into phases/segments. For example, a diving action can be broken down into take-off, flight, entry into the water, etc. Some researchers bear the view that computing segment-level features may constitute a better methodology. Fundamental research\cite{pirsiavash2014assessing, venkataraman2015dynamical, parmar2016measuring, parmar2017learning, bertasius2017baller, doughty2018s, doughty2019pros} that we discussed previously did not explicitly/deliberately compute segment aware features, but were rather segment agnostic methodologies. In this section, we discuss research that computes and uses segment-aware features (see \autoref{fig_trend_segment}). The brief illustration and performance of representative research are summarized in \autoref{table3} and \autoref{fig_trend_segment}.

\begin{table*}[!htbp]
\small
    \caption{Performance comparison of representative methodologies for segment-aware feature extraction.}
    \centering
    \begin{tabular}{@{}lccccl@{}}
        \toprule [1.5pt]
        \textbf{Model} & \textbf{Year} & \textbf{Format} & \textbf{Metric} & \textbf{Performance} & \textbf{Datasets}\\ \midrule[1pt]
        
        S3D\cite{xiang2018s3d} & 2018 & Regression & SRC & 0.8600 & MIT-Dive\\ \arrayrulecolor{Grey0!30} \midrule[0.5pt]
        
        \multirow{2}{*}{MS-LSTM\cite{xu2019learning}} & \multirow{2}{*}{2019} & \multirow{2}{*}{Regression} & \multirow{2}{*}{SRC} & \multirow{1}{*}{0.5900} & \multirow{1}{*}{MIT-Skate} \\
        & & & & 0.7150 & Fis-V \\ \midrule [0.5pt]

        MSRM\cite{dong2021learning} & 2021 & Regression & SRC & 0.8752 & UNLV-Dive  \\ \midrule [0.5pt]

        \multirow{2}{*}{TAL\cite{lei2021temporal}} & \multirow{2}{*}{2021} & \multirow{2}{*}{Regression} & \multirow{2}{*}{SRC} & \multirow{1}{*}{0.8649} & \multirow{1}{*}{UNLV-Dive} \\
        & & & & 0.7858 & UNLV-Vault \\ \midrule [0.5pt]

        \multirow{2}{*}{GDLT\cite{xu2022likert}} & \multirow{2}{*}{2022} & \multirow{2}{*}{Regression} & \multirow{2}{*}{SRC} & \multirow{1}{*}{0.7610} & \multirow{1}{*}{Fis-V} \\
        & & & & 0.7650 & Rhy.Gym. \\ \midrule [0.5pt]

        I3D-Transformer\cite{iyer2022action} & 2022 & Regression & SRC & 0.9317 & MTL-AQA  \\ \midrule [0.5pt]

	\multirow{3}{*}{TPT\cite{bai2022action}} & \multirow{3}{*}{2022} & \multirow{3}{*}{Regression} & \multirow{3}{*}{SRC} & \multirow{1}{*}{0.8900} & \multirow{1}{*}{JIGSAWS} \\
        & & & & 0.8715 & AQA-7 \\ 
        & & & & 0.9607 & MTL-AQA \\	\midrule [0.5pt]

        TSA\cite{xu2022finediving} & 2022 & Regression & SRC & 0.9203 & FineDiving  \\ \midrule [0.5pt]
        
        I3D+TA\cite{zhang2022learning} & 2022 & Regression & SRC & 0.9279 & MTL-AQA  \\ \midrule [0.5pt]

        \multirow{3}{*}{FSPN\cite{gedamu2023fine}} & \multirow{3}{*}{2023} & \multirow{3}{*}{Regression} & \multirow{3}{*}{SRC} & \multirow{1}{*}{0.8724} & \multirow{1}{*}{AQA-7} \\
        & & & & 0.9601 & MTL-AQA \\ 
        & & & & 0.9420 & FineDiving \\ \midrule [0.5pt]

        \multirow{2}{*}{LUSD-Net\cite{ji2023localization}} & \multirow{2}{*}{2023} & \multirow{2}{*}{Regression} & \multirow{2}{*}{SRC} & \multirow{1}{*}{0.7600} & \multirow{1}{*}{Fis-V} \\
        & & & & 0.7920 & FineFS \\ \midrule [0.5pt]
        
        ASTRM\cite{lian2023improving} & 2023 & Regression & SRC & 0.9222 & FineDiving  \\ \midrule [0.5pt]
        
        RG-AQA\cite{liu2023figure} & 2023 & Regression & SRC & 0.9346 & RFJS  \\ \midrule [0.5pt]
        
        PSL\cite{zhang2023labe} & 2023 & Regression & SRC & 0.8700 & UNLV-Dive  \\ \midrule [0.5pt]
        
        GOAT\cite{zhang2023logo} & 2023 & Regression & SRC & 0.5599 & LOGO  \\ \midrule [0.5pt]

        \multirow{3}{*}{HGCN\cite{zhou2023hierarchical}} & \multirow{3}{*}{2023} & \multirow{3}{*}{Regression} & \multirow{3}{*}{SRC} & 
        \multirow{1}{*}{0.9000} & \multirow{1}{*}{JIGSAWS} \\
        & & & & 0.8501 & AQA-7 \\ 
        & & & & 0.9390 & MTL-AQA \\	\midrule [0.5pt]
        
        MCoRe\cite{an2024multi} & 2024 & Regression & SRC & 0.9232 & FineDiving  \\ \midrule [0.5pt]

        \multirow{2}{*}{SSPR\cite{huang2024assessing}} & \multirow{2}{*}{2024} & \multirow{2}{*}{Regression} & \multirow{2}{*}{SRC} & \multirow{1}{*}{0.9257} & \multirow{1}{*}{UNLV-Dive} \\
        & & & & 0.8538 & AQA-7 \\ \midrule [0.5pt]

        \multirow{4}{*}{T$^2$CR\cite{ke2024two}} & \multirow{4}{*}{2024} & \multirow{4}{*}{Regression} & \multirow{4}{*}{SRC} & \multirow{1}{*}{0.9100} & \multirow{1}{*}{JIGSAWS} \\
        & & & & 0.8726 & AQA-7 \\
        & & & & 0.9638 & MTL-AQA \\
        & & & & 0.9275 & FineDiving \\ \midrule [0.5pt]

        STSA\cite{xu2024procedure} & 2024 & Regression & SRC & 0.9397 & FineDiving  \\ \midrule [0.5pt]

        \multirow{2}{*}{FineParser\cite{xu2024fineparser}} & \multirow{2}{*}{2024} & \multirow{2}{*}{Regression} & \multirow{2}{*}{SRC} & \multirow{1}{*}{0.9580} & \multirow{1}{*}{MTL-AQA} \\
        & & & & 0.9435 & FineDiving \\ \midrule [0.5pt]

        Rhythmer\cite{luo2024rhythmer} & 2024 & Pairwise Rank & Acc & 0.8845 & EPIC-Skill \\
        \arrayrulecolor{black} \bottomrule [1.5pt]
    \end{tabular}
    \label{table3}
\end{table*}

The first segment-aware methodology was proposed by Xiang et al.\cite{xiang2018s3d}, utilizing multiple P3D\cite{qiu2017p3d} to construct the stacked 3D regressor (\textbf{S3D}). By utilizing ED-TCN\cite{lea2017edtcn}, the entire video is segmented into video clips of different stages of dives, which are then fed into the corresponding P3D to extract features. Subsequently, the features of the separate stages are fused to predict the score. Similarly, Dong et al.\cite{dong2021learning} and Zhang et al.\cite{zhang2023labe} proposed methodologies also based on ED-TCN and P3D. \textbf{SSRM/MSRM}\cite{dong2021learning} (multi-stage regression module) first segments videos into 5 segmentation by ED-TCN and inputs them into the corresponding stage regression module. Moreover, three different training strategies are proposed according to different action scenarios (the overall-score-guided scenario, execution-score-guided scenario, and difficulty-level-based overall-score-guided scenario) to achieve pertinent and rational assessment. \textbf{PSL}\cite{zhang2023labe} (label-reconstruction-based pseudo-subscore learning) uses the overall score as training labels and generates pseudo-score labels for each sub-stage, thereby addressing the issue of a lack of quality labels for sub-stages.

In addition to models based on CNNs like P3D\cite{qiu2017p3d}, Transformer\cite{vaswani2017attention} have also been applied to AQA\cite{iyer2022action, xu2022likert, bai2022action, fang2023end, gedamu2023fine}. Alali et al.\cite{iyer2022action} simply explored different Transformer-based architectures, with \textbf{I3D-Transformer} decoder performing best. While Xu et al.\cite{xu2022likert}, Bai et al.\cite{bai2022action}, and Gedamu et al.\cite{gedamu2023fine} implemented modifications to Transformer, mainly on the loss function. For example, \textbf{GDLT}\cite{xu2022likert} (grade-decoupling Likert Transformer) regards the score of long-term action as the combination of quantified grades and corresponding responses estimated from the video. So GDLT extracts clip features by VST\cite{liu2022vst} and feeds into the temporal context encoder to obtain the K and V, combined with a set of grades prototypes (Q) to extract grade-aware features by a grade-aware decoder (Transformer). Lastly, scores are predicted by combining the quantitative values and response intensities from grade-aware features. Similarly, Bai et al.\cite{bai2022action, bai2023towards}proposed a temporal parsing Transformer (\textbf{TPT}), which decouples the holistic video features into local temporal features. TPT utilizes a set of learnable queries to represent the atomic temporal patterns for a specific action, combined with holistic features to decode video into a fixed number of part representations. Meanwhile, the ranking and sparsity loss are proposed on cross-attention responses to guide the queries to attend to temporally ordered clips. Finally, CoRe\cite{yu2021group} is adopted to estimate the relative score between input and exemplar videos. \textbf{FSPN}\cite{gedamu2023fine}  (fine-grained spatiotemporal parsing network) extract spatiotemporal action features from pairwise inputs and feed into sub-action parsing module to extract fine-grained sub-action features. Using the spatio-temporal multi-scale Transformer module to capture the long-range spatio-temporal dependencies between sub-actions. Finally, multiple scale features are concatenated to calculate the final score.

Segment-aware research also focuses on \textit{how to segment and utilize segment information}\cite{xu2019learning, lei2021temporal, ji2023localization, zhang2023logo, lian2023improving, liu2023figure, an2024multi, ke2024two, han2023mla, lin2024automatic, hao2023establishment, mourchid2023d, chen2024long, wang2024action}. Xu et al.\cite{xu2019learning} proposed \textbf{MS-LSTM} (self-attentive LSTM and multi-scale convolutional skip LSTM), combining S-LSTM for clip selection and M-LSTM for key action extraction, which together predict TES and PCS scores. Lei et al.\cite{lei2021temporal} introduced temporal attention learning (\textbf{TAL}), simulating the difference in human attention to different action stages when assessing. The video features are extracted by I3D\cite{joao2017i3d} before being fed into the attention learning module, which computes the prediction loss for each clip to dynamically adapt the clip weights and form weighted temporal fusion features. Similarly, \textbf{I3D-TA}\cite{zhang2022learning} utilizes the time-aware attention (TA) mechanism to learn the clip relationship from adversarial loss, which compares the average aggregation result with TA. This enables I3D-TA to capture important clips.

Moreover, Lian and Shao\cite{lian2023improving} proposed an improved methodology based on cross-stage temporal inference, which feeds video features into the across-staged temporal reasoning module (\textbf{ASTRM}) to learn the temporal relationships between stages. Additionally, kernel density estimation (KDE) is employed to recalculate label density, addressing the issue of data imbalance. Liu et al.\cite{liu2023figure} proposed \textbf{RG-AQA} by introducing live, playback, and example videos with a three-stream comparative learning methodology, which is performed between live and sample videos and between live and playback videos, respectively. This ensures that the model focuses on the action itself and learns action features under varying viewpoints and scales. Similarly, An et al.\cite{an2024multi} proposed multi-stage contrastive regression (\textbf{MCoRe}), which uses 2D CNN and gated-shift module (GSM) to extract the spatiotemporal features of the video and bidirectional GRU to divide the video into multiple consecutive stages. With the contrastive learning strategy, positive and negative sample pairs are established between different videos within the same stage. This enhances the accuracy of stage segmentation and the capacity to perceptually differentiate between features\cite{millan2020fine, li2021improving, ke2024two, xu2024fineparser}. Ke et al.\cite{ke2024two} proposed two-path target-aware contrastive regression (\textbf{T$^2$CR}), combining direct regression and contrastive regression to model action-scoring relationships using global spatio-temporal features and local action differences. Xu et al.\cite{xu2024fineparser} proposed a fine-grained spatio-temporal action parser (\textbf{FineParser}) consisting of four modules: the spatial action parser (SAP) extracts the key action regions of the athlete in the video, the temporal action parser (TAP) divides actions into a series of continuous steps, the static visual encoder (SVE) enhances the representation of actions, and the fine-grained contrastive regression (FineReg) compares the action steps in the query video with those in the sample video, assesses the differences in their quality, and generates a final action score. 

Most of the above methodologies tend to segment video into equally sized clips\cite{li2018end, li2018scoringnet, liu2019low, wang2020assessing, wang2020towards, epstein2021learning, parmar2021hallucinet, parsa2021multi, parsa2020deep, abedi2021improving, li2022precise, wang2022skeleton, liu2023multi, gao2024resfnn}, which could lead to inter-clip confusion and inter-clip incoherence. Therefore, many researchers proposed \textit{unequal-length segmentation}\cite{farabi2022improving, li2022hand, ding2023sedskill, zhou2023hierarchical, murthy2023divenet, huang2024assessing, li2024segmentation, luo2024rhythmer}. \textbf{HGCN}\cite{zhou2023hierarchical} (hierarchical graph convolutional network) eliminates semantic confusion within clips through the clip refinement module, moving superfluous information to the clip where its missing part is located. Then, the scene construction module combines multiple consecutive shots into meaningful scenes to capture the dynamics of local actions. Next, the action aggregation module generates video-level features based on the dependencies between scenes and finally predicts the score through the score distribution regression module. \textbf{SSPR}\cite{huang2024assessing} (semantic-sequence performance regression) extracts video features via I3D\cite{joao2017i3d} and feeds into MS-TCN\cite{farha2019mstcn} to segment them into unequal-length clips. A feature fusion module consisting of multiple 1D convolutions to fuse semantic features of the clips, and the clip scores are predicted by regression. \textbf{Rhythmer}\cite{luo2024rhythmer} (rhythm-aware Transformer) adaptively mines rhythm patterns related to task durations, allowing the model to dynamically adjust its focus based on the timing and sequence features of the video. Additionally, Rhythmer incorporates a co-attention module, which highlights duration-related information when comparing video segments with similar execution times. These significantly improve evaluation performance across varying task durations.

Clip-based AQA research is not only limited to methodologies but also to \textit{datasets}\cite{xu2022finediving, ji2023localization, zhang2023logo, xu2024procedure}. Xu et al.\cite{xu2022finediving} proposed the first fine-grained sports video dataset FineDiving and temporal segmentation attention module (\textbf{TSA}), which parses the semantics and temporal structure of actions to enhance the transparency and accuracy of AQA. The quality is assessed by successively performing procedure segmentation, procedure-aware cross-attention learning, and fine-grained contrastive regression. Meanwhile, TSA is supervised by step transition labels and action score labels, which guide the model in focusing on sample regions that are consistent with the query step and quantifying their differences to predict reliable action scores. On this basis, Xu et al.\cite{xu2024procedure} further proposed FineDiving+ and spatial-temporal segmentation attention module (\textbf{STAT}). In comparison to TSA, STSA incorporates a spatial motion attention module (SMA) to facilitate implicit supervision, thereby enabling the model to discern foreground action regions and filter out background noise. Different from Xu et al.\cite{xu2022finediving, xu2024procedure}, Ji et al.\cite{ji2023localization} and Zhang et al.\cite{zhang2023logo} proposed datasets focusing on long video processing. Ji et al.\cite{ji2023localization} proposed FineFS and localization-assisted uncertainty score disentanglement network (\textbf{LUSD-Net}). LUSD-Net extracts video clip features through VST\cite{zhang2023logo}, feeds them into uncertainty score disentanglement module to decouple features into representations for PCS and TES, technical subaction localization module to locate key technical actions, and temporal interaction encoder to integrate clip features to capture contextual relationships. Finally, PCS and TES are predicted separately, and the reliability of the score is improved by uncertainty regression. Zhang et al.\cite{zhang2023logo} proposed LOGO and group-aware attention (\textbf{GOAT}), using GCN to extract spatial group features and combining temporal features to focus on key action clips while ignoring irrelevant information, improving processing for long, complex videos.

\noindent\textbf{(2) Actor-Object-Centric Representations}

\begin{figure}[h]
    \centering
    \includegraphics[width=\columnwidth]{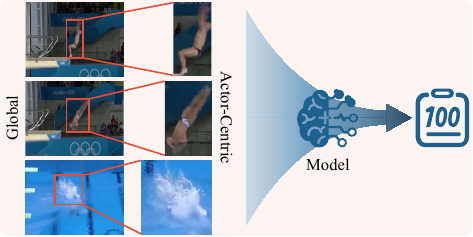}
    \caption{\textbf{Brief illustration of actor-object-centric representations.}}
    \label{fig_trend_actor}
\end{figure}

Pirsiavash et al.\cite{pirsiavash2014assessing} proposed to use skeletons as features. However, just using skeletons as features would not allow to take into consideration critical non-human factors such as splash. Taking a simple and more generalized feature extraction route, which would allow to focus on human and non-human factors, Parmar and Morris et al.\cite{parmar2017learning} proposed to use generalized 3DCNN features (C3D features). Following that, many of the fundamental research\cite{pirsiavash2014assessing, venkataraman2015dynamical, parmar2016measuring, parmar2017learning, bertasius2017baller, doughty2018s, doughty2019pros} typically extracts global image features as the input for the model. However, the global image inevitably contains irrelevant background or noise, particularly in complex scenes with diverse backgrounds. This interference could affect the model's judgment and lead to inaccurate assessments. Therefore, by focusing on the detailed analysis of the key parts and action subjects, from global image features to actor-object-centric features, it is possible to avoid interference from irrelevant factors and improve sensitivity to the details of the action. The brief illustration and performance of representative research are summarized in \autoref{fig_trend_actor} and \autoref{table4}.

\begin{table*}[!htbp]
\small
    \caption{Performance comparison of representative methodologies for actor-object-centric representations.}
    \centering
    \begin{tabular}{@{}lccccl@{}}
        \toprule [1.5pt]
        \textbf{Model} & \textbf{Year} & \textbf{Format} & \textbf{Metric} & \textbf{Performance} & \textbf{Datasets}\\ \midrule[1pt]
        
        \multirow{2}{*}{JR-GCN\cite{pan2019action}} & \multirow{2}{*}{2019} & \multirow{2}{*}{Regression} & \multirow{2}{*}{SRC} & \multirow{1}{*}{0.5700} & \multirow{1}{*}{JIGSAWS} \\
        & & & & 0.7849 & AQA-7 \\ \arrayrulecolor{Grey0!30} \midrule [0.5pt]

        \multirow{2}{*}{ACTION-Net\cite{zeng2020hybrid}} & \multirow{2}{*}{2020} & \multirow{2}{*}{Regression} & \multirow{2}{*}{SRC} & \multirow{1}{*}{0.6150} & \multirow{1}{*}{MIT-Skate} \\
        & & & & 0.6175 & Rhy.Gym. \\ \midrule [0.5pt]

        \multirow{2}{*}{SportCap\cite{chen2021sportscap}} & \multirow{2}{*}{2021} & \multirow{2}{*}{Regression} & \multirow{2}{*}{SRC} & \multirow{1}{*}{0.8620} & \multirow{1}{*}{MTL-AQA} \\
        & & & & 0.6170 & SMART \\ \midrule [0.5pt]
        
        C3D-AVG-SA\&Hmreg\cite{nagai2021action} & 2021 & Regression & SRC & 0.8970 & MTL-AQA \\ \midrule [0.5pt]

        \multirow{2}{*}{EAGLE-Eye\cite{nekoui2021eagle}} & \multirow{2}{*}{2021} & \multirow{2}{*}{Regression} & \multirow{2}{*}{SRC} & \multirow{1}{*}{0.6010} & \multirow{1}{*}{MIT-SKate} \\
        & & & & 0.9158 & AQA-7 \\ \midrule [0.5pt]

        \multirow{2}{*}{TSA-Net\cite{wang2021tsa}} & \multirow{2}{*}{2021} & \multirow{2}{*}{Regression} & \multirow{2}{*}{SRC} & \multirow{1}{*}{0.8476} & \multirow{1}{*}{AQA-7} \\
        & & & & 0.9393 & MTL-AQA \\ \midrule [0.5pt]

        \multirow{2}{*}{PCLN\cite{li2022pairwise}} & \multirow{2}{*}{2022} & \multirow{2}{*}{Regression} & \multirow{2}{*}{SRC} & \multirow{1}{*}{0.8795} & \multirow{1}{*}{AQA-7} \\
        & & & & 0.9014 & MTL-AQA \\ \midrule [0.5pt]

        NS-AQA \cite{okamoto2024hierarchical} & 2024 & Regression & Expert Preference & 0.9610 & FineDiving \\ \midrule [0.5pt]

        \multirow{2}{*}{DuRA\cite{huang2025dual}} & \multirow{2}{*}{2025} & \multirow{2}{*}{Regression} & \multirow{2}{*}{SRC} & \multirow{1}{*}{0.8727} & \multirow{1}{*}{AQA-7} \\
        & & & & 0.9533 & MTL-AQA \\
        
        \arrayrulecolor{black} \bottomrule [1.5pt]
    \end{tabular}
    \label{table4}
\end{table*}

 A different strategy focuses on local images of key parts, extracts skeletal information, and combines the two\cite{sardari2019view, wang2020towards, qiu2022pose, zhu2022robust, hirosawa2023action, hirosawa2023expert, hirosawa2024computer, lei2020learning, nekoui2020falcons, fan2022hightlight, zhang2023toward, zhou2023video, chen2024unlabeled, gallardo2024gymetricpose, wen2024learning, wu2024research, wang2024cpr, zhou2024attention, pan2019action, chen2021sportscap, nekoui2021eagle, lemos2021human}. Pan et al.\cite{pan2019action} firstly proposed joint relation GCN (\textbf{JR-GCN}), focusing on joint interactions and action. Building on the insights gained from GCN, the joint commonality module and joint difference module were employed to calculate joint and neighborhood motion features and temporal and spatial differences of the input local patches around joints, respectively. Integrating the whole-scene feature facilitates precise, fine-grained assessment with an acceptable degree of interpretability. Similarly, Chen et al.\cite{chen2021sportscap} proposed \textbf{SportCap}. The motion embedding module extracts implicit motion features and explicit 3D motion details, embedding them into high-dimensional space. Then, using spatial-temporal GCN (ST-GCN) to model the spatial-temporal relationships between body joints. Additionally, fine-grained action understanding is achieved through a sub-motion classifier and semantic attribute mapping module, integrating multiple action attributes into high-level action labels. Nekoui et al.\cite{nekoui2021eagle, nekoui2022intelligent} proposed (\textbf{EAGLE-Eye}). The joints coordination assessor (JCA) captures local skeletal point dynamics and interactions through multi-scale temporal convolutions, while the appearance dynamics assessor (ADA) extracts global appearance dynamics. With explicit spatiotemporal attention mechanisms, the network could focus on critical time points and spatial regions.

The other strategy focuses on the action subject\cite{ganesh2019novel, li2019manipulation, zeng2020hybrid, nagai2021action, wang2021tsa, li2021and, freire2022towards, wang2023three, zhao2023knowledge, anastasiou2023keep, freire2023x3d, zhou2023prior, kondo2024zeal, huang2024full, li2022pairwise, huang2025dual}. For example, Zeng et al.\cite{zeng2020hybrid} focused on the analysis of both motion information and static pose information,  proposed \textbf{ACTION-Net}. The network uses I3D\cite{joao2017i3d} and ResNet\cite{he2016resnet} to extract motion and background information from video segments and the spatial information of the athlete's pose and appearance in specific frames. These features are then aggregated to produce dynamic and static features. The context-aware attention module aggregates all video segments/frames to produce dynamic/static features in each stream, then fed into a regression network to predict scores. Different from ACTION-NET\cite{zeng2020hybrid}, Nagai et al.\cite{nagai2021action}, Wang et al.\cite{wang2021tsa}, Li et al.\cite{li2022pairwise}, and Haung et al.\cite{huang2025dual} proposed methodologies that ignore irrelevant backgrounds by highlighting the primary subject of actions. But Nagai et al.\cite{nagai2021action} and Wang et al.\cite{wang2021tsa} achieved by direct regression,  Li et al.\cite{li2022pairwise}, and Haung et al.\cite{huang2025dual} combined contrastive learning and regression network. 

\textbf{C3D-AVG-SA\&HMreg}\cite{nagai2021action} encompass the scene adversarial loss (SAL) and human-masked regression loss (HMRL). SAL restrains the model's dependency on scene background information through adversarial training. In contrast, HMRL ensures the model's concentration on actions by regressing to a fixed score when the target action is not visible. While, \textbf{TSA-Net}\cite{wang2021tsa} combines target tracking and self-attention mechanisms, using VOT trackers and I3D\cite{joao2017i3d} to extract tracking results and features, which are then aggregated using the TSA module and fed into regression network to predict scores. \textbf{PCLN}\cite{li2022pairwise} uses Resnet\cite{he2016resnet} to extract features of paired videos, then calculate the expected score and predicted difference score, respectively. \textbf{DuRA}\cite{huang2025dual} leverages both semantic-level grade prototypes and individual-level reference samples to enhance the focus on action details while filtering out irrelevant information. At the semantic level, the rating-guided attention (RGA) module is used to emphasize local features critical to action quality. At the individual level, consistency preserving (CP) constraints are used to refine the representation of local features and suppress distractions.

Okamoto and Parmar\cite{okamoto2024hierarchical} proposed to use \textbf{dedicated human pose estimators} and \textbf{object detectors} like platform detectors and splash detectors to extract Actor and Object-centric information. And then processes this information using rules to generate detailed AQA reports, including scores.

Another way to guide model's attention to athlete and other important features like splash is using \textbf{detailed supervision} as done by Parmar and Morris \cite{parmar2019and}. By optimizing the network end-to-end for detailed action classification, detailed action description and AQA scoring, the network would be able to better characterize the action by better focusing on the actor, action, and important factors like splash. \\

\noindent\textbf{(3) From Subjective to Objective Evaluation}

"\textit{If you want fairness, come to sports; if you want absolute fairness, don't come to sports!}" — From film "\textbf{Pegasus}"

\begin{figure}[h]
    \centering
    \includegraphics[width=\columnwidth]{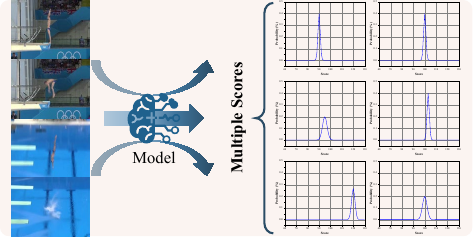}
    \caption{\textbf{Brief illustration of objective evaluation.}}
    \label{fig_trend_object}
\end{figure}

Fairness has always been a critical concern in competitive sports. Various measures, such as multi-judge scoring systems, have been implemented to make competitions as fair as possible. However, biases still exist, including nationalistic bias\cite{emerson2009bais1, ansorge1988bais2}, where judges may favor athletes from their own country or give preference to athletes performing more difficult maneuvers\cite{morgan2014harder}. Fundamental research\cite{pirsiavash2014assessing, venkataraman2015dynamical, parmar2016measuring, parmar2017learning, bertasius2017baller, doughty2018s, doughty2019pros} used referee scores as ground truth without accounting for the uncertainty and potential ambiguity arising from subjective assessments. This reliance on subjective scores affected the accuracy and fairness of the models and hindered their practical application. Therefore, extracting effective information from existing scoring labels and improving the accuracy of AQA has become a research focus. Researchers have introduced methodologies such as uncertainty modeling\cite{tang2020uncertainty, zhou2022uncertainty, karunaratne2021objectively, li2023gaussian, majeedi2024rica, zhang2024auto}, contrastive regression\cite{jain2021action, yu2021group} to reduce the subjective bias in human scoring. The brief illustration and performance of representative research are summarized in \autoref{fig_trend_object} and \autoref{table6}.

\begin{table*}[!htbp]
\small
    \caption{Performance comparison of representative methodologies for objective evaluation.}
    \centering
    \begin{tabular}{@{}lccccl@{}}
        \toprule [2pt]
        \textbf{Model} & \textbf{Year} & \textbf{Format} & \textbf{Metric} & \textbf{Performance} & \textbf{Datasets}\\ \midrule [1pt]
        \multirow{3}{*}{USDL\cite{tang2020uncertainty}} & \multirow{3}{*}{2020} & \multirow{3}{*}{Regression} & \multirow{3}{*}{SRC} & 
        \multirow{1}{*}{0.7000} & \multirow{1}{*}{JIGSAWS} \\
        & & & & 0.8102 & AQA-7 \\ 
        & & & & 0.9273 & MTL-AQA \\	\arrayrulecolor{Grey0!30} \midrule [0.5pt]

        \multirow{3}{*}{RGR\cite{jain2021action}} & \multirow{3}{*}{2021} & \multirow{3}{*}{Regression} & \multirow{3}{*}{SRC} & \multirow{1}{*}{0.6900} & \multirow{1}{*}{UNLV-Dive} \\
        & & & & 0.5360 & UNLV-Vault \\	
        & & & & 0.7600 & MTL-AQA \\ \midrule [0.5pt]

        \multirow{3}{*}{CoRe\cite{yu2021group}} & \multirow{3}{*}{2021} & \multirow{3}{*}{Regression} & \multirow{3}{*}{SRC} & 
        \multirow{1}{*}{0.8400} & \multirow{1}{*}{JIGSAWS} \\
        & & & & 0.8410 & AQA-7 \\ 
        & & & & 0.9512 & MTL-AQA \\ \midrule [0.5pt]

        \multirow{3}{*}{UD-AQA\cite{zhou2022uncertainty}} & \multirow{3}{*}{2022} & \multirow{3}{*}{Regression} & \multirow{3}{*}{SRC} & \multirow{1}{*}{0.8900} & \multirow{1}{*}{JIGSAWS} \\
        & & & & 0.9545 & MTL-AQA \\	
        & & & & 0.9341 & FineDiving \\ \midrule [0.5pt]

        \multirow{2}{*}{TECN\cite{li2023gaussian}} & \multirow{2}{*}{2023} & \multirow{2}{*}{Regression} & \multirow{2}{*}{SRC} & \multirow{1}{*}{0.8504} & \multirow{1}{*}{AQA-7} \\
        & & & & 0.9095 & MTL-AQA \\ \midrule [0.5pt]

        \multirow{3}{*}{DAE\cite{zhang2024auto}} & \multirow{3}{*}{2024} & \multirow{3}{*}{Regression} & \multirow{3}{*}{SRC} & 
        \multirow{1}{*}{0.7600} & \multirow{1}{*}{JIGSAWS} \\
        & & & & 0.8258 & AQA-7 \\ 
        & & & & 0.9232 & MTL-AQA \\ \midrule [0.5pt]

        NS-AQA \cite{okamoto2024hierarchical} & 2024 & Regression & Expert Preference & 0.9610 & FineDiving \\ \midrule [0.5pt]

        \multirow{3}{*}{RICA$^2$\cite{majeedi2024rica}} & \multirow{3}{*}{2024} & \multirow{3}{*}{Regression} & \multirow{3}{*}{SRC} & \multirow{1}{*}{0.9200} & \multirow{1}{*}{JIGSAWS} \\
        & & & & 0.9512 & MTL-AQA \\ 
        & & & & 0.9421 & FineDiving \\ 


        \arrayrulecolor{black} \bottomrule [2pt]
    \end{tabular}
    \label{table5}
\end{table*}

Tang et al.\cite{tang2020uncertainty} was the first to introduce uncertainty modeling into AQA to reduce the intrinsic ambiguity in score labels caused by multiple judges and their subjective assessments, proposed uncertainty-aware score distribution learning (\textbf{USDL}), which generates a gaussian distribution of the labeled scores and calculates predicted score distribution of a given video and score label. By minimizing the KL divergence between two distributions, the network is optimized to fit the actual scoring process better. Drawing inspiration from the Olympic multi-judge system, USDL could extend to K pathways, where scores from different paths are weighted and fused to predict the final score. Similarly, several researchers have further explored uncertainty-based methodologies. Zhou et al.\cite{zhou2022uncertainty} proposed uncertainty-driven AQA (\textbf{UD-AQA}), consisting of a deterministic branch and a latent branch. The deterministic branch extracts video features by I3D\cite{joao2017i3d}, while the latent branch models the inherent ambiguity of the video using conditional variational auto-encoder (CVAE), combining both branches to predict scores. Li et al.\cite{li2023gaussian} proposed a gaussian-guided frame sequence encoder network, which uses ResNet\cite{he2016resnet} to extract features from videos, then fed into sequence-based temporal encoder convolutional network (\textbf{TECN}) to extract the spatio-temporal semantic information of actions. Gaussian loss function is used to maximize the probability of the predicted scores fitting the distribution. Zhang et al.\cite{zhang2024auto} proposed distribution auto-encoder (\textbf{DAE}), a regression model that encodes video features into a score distribution and uses a reparameterization trick to sample predictions, simultaneously learning to predict scores and uncertainty. Majeedi et al.\cite{majeedi2024rica} proposed a rubric-informed, calibrated assessment of actions (\textbf{RICA$^2$}), which represents action steps and scoring criteria as a directed acyclic graph (DAG). GNN is used to capture uncertainty in the scoring process by generating probabilistic embeddings for each action step, which are then propagated upward to calculate the final score. 

In contrast to uncertainty modeling, contrastive regression\cite{jain2021action, yu2021group, sun2023novel, joung2023contrastive, liu2024hierarchical, huang2024full} is used to reduce subjective scoring bias by comparing sample and reference videos. Jain et al.\cite{jain2021action, jain2020assessment} proposed reference-guided regression (\textbf{RGR}), transforming an action scoring task into an action video similarity scoring task. Videos are processed in pairs using the Siamese-LSTM network, where C3D\cite{tran2015c3d} extracts clip features and feeds them into LSTM to generate global action features. Subsequently, extracted features are concatenated and fed into a regression model, which outputs a similarity judgment. Besides, RGR could calculate clip similarity between sample and reference videos, allowing for more fine-grained and interpretable scoring. Yu et al.\cite{yu2021group} proposed the contrastive regression (\textbf{CoRe}) framework, which selects a reference video for each sample video based on action category and difficulty. Paired spatiotemporal features are extracted using I3D and combined with the reference score, then fed into a group-aware regression tree (GART). GART classifies features from coarse to fine, regressing the score difference between the two videos.

\noindent\textbf{(4) Modeling Asymmetric Relationships}

\begin{figure}[h]
    \centering
    \includegraphics[width=\columnwidth]{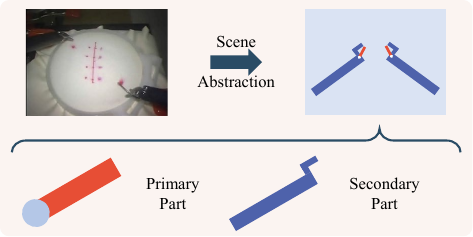}
    \caption{\textbf{Brief illustration of modeling asymmetric relationships.}}
    \label{fig_trend_asymmteric}
\end{figure}

Fundamental research\cite{pirsiavash2014assessing, venkataraman2015dynamical, parmar2016measuring, parmar2017learning, bertasius2017baller, doughty2018s, doughty2019pros} concentrate on single-agent actions. For multiple agents in the scene, a general assumption is that the impact of different agents on the quality is equivalent. However, in daily life and competitive sports, it is evident that many behaviors comprise interactions with asymmetric relationships between different agents. Therefore, previous research fails to consider priority in real-world scenarios and lacks the modeling of asymmetric relationships between different agents. Asymmetric research facilitates more detailed feature extraction, so we consider this to be a fine-grained category. The brief illustration and performance of representative research are summarized in \autoref{fig_trend_asymmteric} and \autoref{table5}.

The asymmetric relationship was first proposed in AQA and made major contributions by Gao et al.\cite{gao2020asymmetric, gao2023automatic}. To date, this research include two models, \textbf{AIM}\cite{gao2020asymmetric} and \textbf{AIL}\cite{gao2023automatic}, and two dedicated datasets, TASD-2\cite{gao2020asymmetric} and PaSK\cite{gao2023automatic}. AIM\cite{gao2020asymmetric}, as the first asymmetric relationship AQA model, is used to explicitly model asymmetric interactions between agents. Initially, AIM manually categorizes participating agents as primary or secondary, calculates the difference features between them and fuses the primary information to obtain the primary-secondary information. These are fed into LSTM to learn the temporal interaction, thereby obtaining complete AIM features. Additionally, global scene features are extracted using I3D\cite{joao2017i3d} and fused with AIM features via an attention mechanism to predict action scores. Building on AIM, AIL\cite{gao2023automatic} introduced two key modules: the automatic assigner and the operation search module, which can automatically classify the primary and secondary agents in an action and dynamically adjust their structure for different interactive actions. Furthermore, AIL can adaptively model different action scenarios, including strong asymmetry and weak asymmetry, enhancing the flexibility and generalization ability of the model.

\begin{table}[!htbp]
\small
    \caption{Performance comparison of representative methodologies for modeling asymmetric relationships. Yr.: Year, Fmt.: Format, Met.: Metric, Perf.: Performance.}
    \centering
    \begin{tabular}{@{}lccccl@{}}
        \toprule [1.5pt]
        \textbf{Model} & \textbf{Yr.} & \textbf{Fmt.} & \textbf{Met.} & \textbf{Perf.} & \textbf{Datasets}\\ \midrule [1pt]
        \multirow{3}{*}{AIM\cite{gao2020asymmetric}} & \multirow{3}{*}{'20} & \multirow{3}{*}{Reg} & \multirow{3}{*}{SRC} & \multirow{1}{*}{0.7100} & \multirow{1}{*}{JIGSAWS} \\
        & & & & 0.7789 & AQA-7 \\	
        & & & & 0.8831 & TASD-2 \\ \arrayrulecolor{Grey0!30} \midrule [0.5pt]

        \multirow{5}{*}{AIL\cite{gao2023automatic}} & \multirow{5}{*}{'23} & \multirow{3}{*}{Reg} & \multirow{3}{*}{SRC} & \multirow{1}{*}{0.8800} & \multirow{1}{*}{JIGSAWS} \\
        & & & & 0.8126 & AQA-7 \\
        & & & & 0.8600 & PaSk \\	
        & & \multirow{2}{*}{PR} &\multirow{2}{*}{Acc} & 0.8340 & EPIC-Skill \\
        & & & & 0.8190 & BEST \\

        \arrayrulecolor{black} \bottomrule [1.5pt]
    \end{tabular}
    \label{table6}
\end{table}

\noindent\textbf{(5) Conclusion of Fine-grained}

A thorough review of the methodologies described and an analysis of the tables (\autoref{table3}, \autoref{table4}, \autoref{table5},  \autoref{table6}) above clearly demonstrates that fine-grained research has significantly improved model performance compared to \cite{pirsiavash2014assessing, venkataraman2015dynamical, parmar2016measuring, parmar2017learning, bertasius2017baller, doughty2018s, doughty2019pros}. In particular, this finding demonstrates the superiority of fine-grained research methods in capturing minutiae, suppressing extraneous variables, resolving temporal dependencies, and handling label ambiguity. The four sub-trends encompassed within the fine-grained research reflect the researchers' in-depth consideration of multiple dimensions, including temporal, spatial, outcome, and agent relationships. Given that AQA involves traversing the entire action sequence, research emphasizing segment-aware feature extraction has become the primary focus of fine-grained research.

\subsubsection{Multitask \& Multimodal AQA}\label{subsubsec4.2.2}
\begin{figure}[h]
    \centering
    \includegraphics[width=\columnwidth]{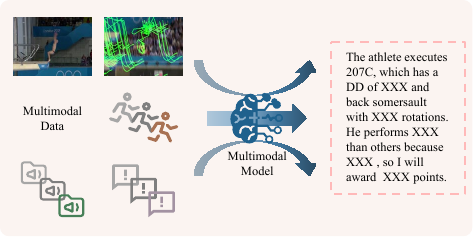}
    \caption{\textbf{Brief illustration of multimodal AQA.}}
    \label{fig_trend_mm}
\end{figure}

Early AQA research demonstrated promising performance using RGB videos\cite{pirsiavash2014assessing, parmar2017learning, doughty2018s, doughty2019pros} and skeletons\cite{vakanski2018data} as input modalities. However, with the advancement of time and technology, the limitations of single-modal data have become evident. Researchers now can handle more complex data modalities, leading to the rise of multimodal methodologies as a key research trend. Parmar et al.\cite{parmar2021piano} introduced multimodal AQA. Multimodal methodologies offer significant advantages, compensating for single-modality limitations in complex scenarios by providing complementary information. This enhances model robustness and enables a more detailed AQA. The brief illustration and performance of representative research is summarized in \autoref{fig_trend_mm} and \autoref{table7}.

\begin{table*}[!htbp]
\small
    \caption{Performance comparison of representative multitask and multimodal methodologies.}
    \centering
    \begin{tabular}{@{}lccccl@{}}
        \toprule [2pt]
        \textbf{Model} & \textbf{Year} & \textbf{Format} & \textbf{Metric} & \textbf{Performance} & \textbf{Datasets}\\ \midrule [1pt]
        
        C3D-AVG-MTL\cite{parmar2019and} & 2019 & Regression & SRC & 0.9044 & MTL-AQA \\ \arrayrulecolor{Grey0!30} \midrule [0.5pt]
        
        ST-GCN\cite{li2021skeleton} & 2021 & Regression & SRC & 0.6300 & MIT-Skate \\ \midrule [0.5pt]

        PISA\cite{parmar2021piano} & 2021 & Classification & Acc & 0.7460 & Piano-Skills \\ \midrule [0.5pt]

        \multirow{2}{*}{SGN\cite{du2023learning}} & \multirow{2}{*}{2023} & \multirow{2}{*}{Regression} & \multirow{2}{*}{SRC} & \multirow{1}{*}{0.9607} & \multirow{1}{*}{MTL-AQA} \\
        & & & & 0.7650 & Fis-V\\ \midrule [0.5pt]
        
        
        MS-GCN\cite{lei2023multi} & 2023 & Regression & SRC & 0.7200 & Rhy.Gym. \\ \midrule [0.5pt]

        Skating-Mixer\cite{xia2023skating} & 2023 & Regression & SRC & 0.7500 & Fis-V \\ \midrule [0.5pt]
        
        RI\cite{zheng2023skeleton} & 2023 & Classification & Acc & 0.9886 & UI-PRDM \\ \midrule [0.5pt]

        \multirow{2}{*}{EGCN++\cite{bruce2024egcn++}} & \multirow{2}{*}{2024} & \multirow{2}{*}{Classification} & \multirow{2}{*}{Acc} & \multirow{1}{*}{0.9500} & \multirow{1}{*}{UI-PRDM} \\
        & & & & 0.9282 & KIMORE \\ \midrule [0.5pt]

        \multirow{2}{*}{2M-AF\cite{ding20242m}} & \multirow{2}{*}{2024} & \multirow{2}{*}{Regression} & \multirow{2}{*}{SRC} & \multirow{1}{*}{0.8838} & \multirow{1}{*}{UNLV-Diving} \\
        & & & & 0.8901 & AQA-7 \\ \midrule [0.5pt]

        \multirow{4}{*}{VATP-Net\cite{gedamu2024visual}} & \multirow{4}{*}{2024} & \multirow{4}{*}{Regression} & \multirow{4}{*}{SRC} & \multirow{1}{*}{0.9588} & \multirow{1}{*}{MTL-AQA} \\
        & & & & 0.7960 & Fis-V \\
        & & & & 0.8000 & Rhy.Gym. \\
        & & & & 0.7725 & FineFS \\ \midrule [0.5pt]

        EK-GCN\cite{he2024expert} & 2024 & Regression & SRC & 0.8080 & KIMORE \\  \midrule [0.5pt]
        
        DNLA\cite{zahan2024learning} & 2024 & Regression & SRC & 0.6800 & AGF-Olympics \\ \midrule [0.5pt]

        \multirow{2}{*}{PAMFN\cite{zeng2024multimodal} } & \multirow{2}{*}{2024} & \multirow{2}{*}{Regression} & \multirow{2}{*}{SRC} & \multirow{1}{*}{0.8220} & \multirow{1}{*}{Fis-V} \\
        & & & & 0.8190 & Rhy.Gym. \\ \midrule [0.5pt]
        
        NAE\cite{zhang2024narrative} & 2024 & Regression & SRC & 0.9790 & MTL-AQA \\

        \arrayrulecolor{black} \bottomrule [2pt]
    \end{tabular}
    \label{table7}
\end{table*}

The initial multi-task model in AQA came from Parmar and Morris\cite{parmar2019and, parmar2019actionb}, who proposed an end-to-end multi-task learning framework (\textbf{C3D-AVG-MTL}), marking a pioneering introduction of text into the AQA. The C3D-AVG-MTL extracts spatiotemporal features through a shared C3D\cite{tran2015c3d} backbone and transmits them to independently task-specific branches, including AQA score regressor, factorized action classifier, and caption generator. The whole model, including the backbone and individual branches, is trained end-to-end to learn powerful AQA-oriented representations. An end-to-end joint optimization network enables the realization of fine-grained action descriptions and AQA scoring. 

Different from Parmar and Morris\cite{parmar2019and}, Du et al.\cite{du2023learning}, Xu et al.\cite{xu2024vision}, and Gedamu et al.\cite{gedamu2024visual} chose to amalgamate visual and semantic information rather than process them independently. \textbf{SGN}\cite{du2023learning} (semantics-guided network) employs teacher-student architecture to amalgamate visual and semantic information, with the teacher network providing knowledge containing semantic information and the student network adapting this knowledge by a set of learnable atomic queries and attention mechanism. Three loss functions were proposed to ensure the effective alignment of visual and semantic features. Similarly, \textbf{VATP-Net}\cite{gedamu2024visual} (visual-semantic alignment temporal parsing network) contains two modules. The self-supervised temporal parsing module is responsible for understanding the high-level representation along with internal temporal structures of action features, while the multimodal interaction module captures visual-semantic action features and the interaction between different modalities, enabling a better understanding of scene-invariant action execution. The entire network leverages self-supervised and multimodal learning, significantly improving action quality assessment performance without relying on external labels. Following Parmar and Morris\cite{parmar2019and}, Zhang et al.\cite{zhang2024narrative} proposed a new task called narrative action evaluation (\textbf{NAE}), transforming score regression into a video-text matching task. In this methodology, video features are extracted by a video encoder. At the same time, context-aware prompt learning employs a multi-head cross-attention mechanism to match learnable prompts with video features, allowing prompts to perceive the video's context. In the score-guided tokens learning, video features are combined with prompts containing scoring information. These combined features are fed into the multimodal-aware text generator, which generates natural language descriptions that include action details, scores, and quality assessments. A tri-token attention mask is employed throughout the process to balance the linguistic richness of the action description with the accuracy of the assessment, thereby generating detailed and professional narratives and evaluations. The interaction between different modal tasks, particularly the linkage between action assessment and text generation, facilitates enhanced mutual promotion.

In addition to the text modality, many researchers introduced skeleton into AQA\cite{ogata2019temporal, li2021skeleton, li2022skeleton, li2022tai, cui2023study, siow2023evaluating, yao2023contrastive, zhang2023action, lei2023multi, zheng2023skeleton, shen2024markerless, xu2024reveal, bruce2024egcn++, ding20242m, he2024expert, zahan2024learning, louis2024improving, zhang2023hand}. The majority of them conducted research based on GCN but with distinct methodologies. For example, \textbf{ST-GCN}\cite{li2021skeleton} (spatiotemporal GCN) extracts features from skeleton data and deep pose features in the spatiotemporal dimension, then fed into regression network to predict scores. \textbf{MS-GCN}\cite{lei2023multi} (multi-skeleton structures GCN) extracts and segments skeleton sequences in long videos into non-overlapping subsequences. Meanwhile, three types of skeleton graphs (self-connection, intra-part connection, and inter-part connection) are constructed to extract pose features. The temporal relationships between sub-sequences were learned by a temporal attention module and then fed into the regression network. \textbf{RI}\cite{zheng2023skeleton} calculates the relative directional relationships between body joints to construct a dot product matrix that effectively removes the influence of body orientation, thereby addressing the sensitivity of traditional models to orientation changes. \textbf{EGCN++}\cite{bruce2024egcn++} combines positional and orientation information from skeleton data at the data level for comprehensive feature integration. In the graph convolutional layers, the model captures spatial relationships between joints and extracts deep features through the graph structure. At the model level, an ensemble learning strategy is employed to combine predictions from multiple graph convolutional networks, improving accuracy and robustness. It is worth mentioning that EGCN++ relabels KIMORE\cite{capecci2019kimore} to change the score to a 2-level categorical label. \textbf{2M-AF}\cite{ding20242m} (multi-modality assessment framework) employs self-supervised mask encoded GCN (SME-GCN) and I3D\cite{joao2017i3d} to extract skeleton and RGB features. These are fused by a preference fusion module (PFM) and fed into a regression network to predict scores. \textbf{EK-GCN}\cite{he2024expert} captures spatial relationships between joints by GCN, with expert knowledge guiding the network to focus on key joints and motion features through weighted graph structures. The gated pooling module selectively aggregates features, while Transformer module models long-range temporal dependencies. \textbf{DNLA}\cite{zahan2024learning, zahan2024human} (discriminative nonlocal attention) extracts sparse features with cross-temporal correlations from the skeleton and video clips, which are fed into the regression network for predicting scores. DNLA eliminates superfluous data, enabling the model to concentrate on pivotal actions and frames.

Beyond text and skeleton, researchers\cite{parmar2021piano, xia2023skating, kim2024kinematic, zeng2024multimodal} combined visual information with audio information. \textbf{PISA}\cite{parmar2021piano} utilized 3DCNN and 2DCNN to extract video and audio features, respectively, concatenated the features, and finally output the piano performance skill level through the fully connected layer. \textbf{Skating-Mixer}\cite{xia2023skating} extracts video and audio features through CNN and MFCC, respectively. These features are then fused into a joint feature vector and passed through the fully connected layer. Meanwhile, Skating-Mixer enhances traditional MLP frameworks by incorporating a memory mechanism, making the model not only process the current frame but also recall and leverage information from previous frames. Notably, Zeng and Zheng et al.\cite{zeng2024multimodal} took optical flow into account, proposing a progressive, adaptive multimodal fusion network (\textbf{PAMFN}). The RGB, optical flow, and audio data were extracted through three independent modality-specific branches, and a mixed-modality branch progressively fused the information using a modality-specific feature decoder and an adaptive fusion module, selecting the optimal fusion strategy. The cross-modal feature decoder further transmits the cross-modal features after adaptive fusion to the mixed modality branch, leading to accurate action quality assessments.

\subsubsection{Generalization in AQA}\label{subsubsec4.2.3}
\begin{figure}[h]
    \centering
    \includegraphics[width=\columnwidth]{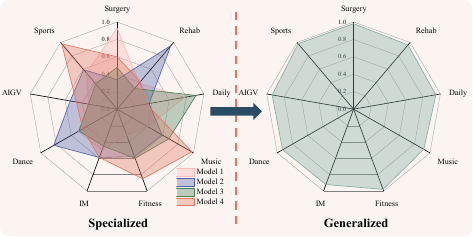}
    \caption{\centering \textbf{Brief illustration of generalization in AQA.}}
    \label{fig_trend_generalization}
\end{figure}

As discussed above, AQA research has typically focused on single-action scenarios. In the initial stage, due to the limitations of available datasets and technical capabilities, researchers prioritized single-action scenarios with clear assessment criteria. However, the generalization ability of models has become a major focus in AQA research. Single-action scene dataset trained models face significant limitations when applied to diverse, open environments and cross-task assessments. Researchers seek to enhance the generalization ability of models, enabling their deployment in disparate datasets, tasks, and application scenarios. The brief illustration and performance of representative research is summarized in \autoref{fig_trend_generalization} and \autoref{table8}.

\begin{table*}[!htbp]
\small
    \caption{Performance comparison of representative methodologies for generalization.}
    \centering
    \begin{tabular}{@{}lccccl@{}}
        \toprule [1.5pt]
        \textbf{Model} & \textbf{Year} & \textbf{Format} & \textbf{Metric} & \textbf{Performance} & \textbf{Datasets}\\ \midrule [1pt]
        
        C3D+LSTM\cite{parmar2019action} & 2019 & Regression & SRC & 0.6478 & AQA-7 \\ \arrayrulecolor{Grey0!30} \midrule [0.5pt]

        \multirow{4}{*}{Adaptive Net \cite{pan2022adaptive}} & \multirow{4}{*}{2022} & \multirow{2}{*}{Regression} & \multirow{2}{*}{SRC} & \multirow{1}{*}{0.7815} & \multirow{1}{*}{JIGSAWS} \\
        & & & & 0.8500 & AQA-7 \\
        & &\multirow{2}{*}{Pairwise Rank} &\multirow{2}{*}{Acc} & 0.8406 & EPIC-Skill \\
        & & & & 0.8327 & BEST \\ \midrule [0.5pt]

        \multirow{3}{*}{AdaST\cite{zhang2023adaptive}} & \multirow{3}{*}{2023} & \multirow{1}{*}{Regression} & \multirow{1}{*}{SRC} & \multirow{1}{*}{0.8443} & \multirow{1}{*}{AQA-7} \\
        & & \multirow{2}{*}{Pairwise Rank} & \multirow{2}{*}{Acc} &  \multirow{1}{*}{0.8832} & \multirow{1}{*}{EPIC-Skill} \\  
        & & & & 0.8534 & BEST \\ \midrule [0.5pt]

        \multirow{2}{*}{CoFInAl\cite{zhou2024cofinal}} & \multirow{2}{*}{2024} & \multirow{2}{*}{Classification} & \multirow{2}{*}{SRC} & \multirow{1}{*}{0.7880} & \multirow{1}{*}{Fis-V} \\
        & & & & 0.8070 & Rhy.Gym. \\
        
        \arrayrulecolor{black} \bottomrule [1.5pt]
    \end{tabular}
    \label{table8}
\end{table*}

The initial generalization model in AQA also came from Parmar and Morris\cite{parmar2019action}, who trained a \textbf{C3D+LSTM} model using the multi-action scene dataset AQA-7. The experimental results demonstrated that the model trained on a multi-action scene dataset showed superior performance and generalization ability. Parmar and Morris\cite{parmar2019action} highlighted that common features can be learned across different actions, and training on multi-action scene dataset allows the model to benefit from knowledge transfer and sharing, leading to faster convergence and improved evaluation results even with a limited number of samples. This study laid the foundation for subsequent research on the generalization of multi-action scenarios and provided a widely cited dataset.

Unlike Parmar and Morris\cite{parmar2019action}, who focused on training models by multi-action scenarios, some researchers\cite{pan2022adaptive, zhang2023adaptive, zhou2024cofinal} aspire to realize high generalization through advancements in model architecture. Pan et al.\cite{pan2022adaptive} addressed the problem of single architectures struggling to achieve high performance across all actions and the necessity for manually designing architectures for different types of actions. They proposed the initial adaptive network (\textbf{Adaptive Net}) for AQA, which is constructed by acyclic graph and inputs features extracted on JR-GCN. The network contains multiple operation selectors, each containing seven candidate operations. The optimal network architecture is searched using the differentiate architecture search (DARTS) mechanism. Adaptive Net demonstrated feasibility and superior performance in regression and ranking tasks across various datasets, including competitive sports, surgery, and daily actions. Zhang et al.\cite{zhang2023adaptive} propose an adaptive stage-aware assessment skill transfer framework (\textbf{AdaST}), different from the existing assessment skill transfer that pre-train the model with all source actions or jointly training with source actions and target actions, AdaST models the relationship between the source and target action. Firstly, AdaST employs a genetic-based adaptive source action search scheme to identify relevant source actions. Next, the target action samples are directly fed into the encoders to extract features, after which the stage-aware assessment module transfers different assessment skills from the source actions to the corresponding stages of the target action. Similarly, Zhou et al.\cite{zhou2024cofinal} proposed coarse-to-fine instruction alignment (\textbf{CoFInAl}) to solve issues of domain shifting and overfitting under limited data. CoFInAl mimics the referee's judging process by first providing coarse grade assessments and then fine-grain scores within each grade, aligning the AQA task with the pre-trained model.

\subsubsection{Continual learning AQA}\label{subsubsec4.2.4}
To adapt to external changes, humans and other living organisms have developed strong adaptive abilities, enabling them to acquire, update, accumulate, and utilize knowledge continuously. Naturally, we hope that AI can function similarly, which is why the concept of continuous learning has been proposed. Unlike traditional models that learn from static data distributions, continual learning involves learning from dynamic data distributions. A significant challenge identified in continual learning is catastrophic forgetting, where the model's adaptation to a new distribution typically results in a substantial decline in performance on the old distribution. Some researchers introduced continual learning to AQA\cite{dadashzadeh2024pecop, li2024continual, zhou2024magr} and tried to address the catastrophic forgetting. The brief illustration and performance of representative research is summarized in \autoref{fig_trend_actor} and \autoref{table9}.

\begin{figure}[h]
    \centering
    \includegraphics[width=\columnwidth]{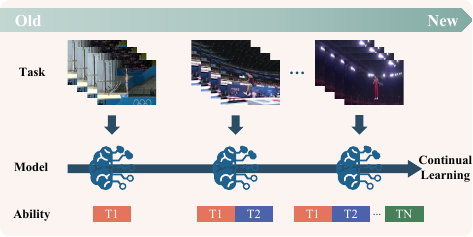}
    \caption{\textbf{Brief illustration of continual learning in AQA.}}
    \label{fig_trend_cl}
\end{figure}

\begin{table*}[!htbp]
\small
    \caption{Performance comparison of representative continual-learning methodologies.}
    \centering
    \begin{tabular}{@{}lccccl@{}}
        \toprule [1.5pt]
        \textbf{Model} & \textbf{Year} & \textbf{Format} & \textbf{Metric} & \textbf{Performance} & \textbf{Datasets}\\ \midrule [1pt]
        
        \multirow{2}{*}{PECoP\cite{dadashzadeh2024pecop}} & \multirow{2}{*}{2024} & \multirow{2}{*}{Regression} & \multirow{2}{*}{SRC} & \multirow{1}{*}{0.8900} & \multirow{1}{*}{JIGSAWS} \\
        & & & & 0.9520 & MTL-AQA \\ \arrayrulecolor{Grey0!30}  \midrule [0.5pt]

        \multirow{3}{*}{Continual-AQA\cite{li2024continual}} & \multirow{3}{*}{2024} & \multirow{2}{*}{Regression} & \multirow{2}{*}{SRC} & \multirow{1}{*}{0.6492} & \multirow{1}{*}{AQA-7} \\
        & & & & 0.7895 & MTL-AQA \\	
        & &\multirow{1}{*}{Pairwise Rank} &\multirow{1}{*}{Acc} & 0.7883 & BEST \\ \midrule [0.5pt]
        
        \multirow{3}{*}{MAGR\cite{zhou2024magr}} & \multirow{3}{*}{2024} & \multirow{3}{*}{Regression} & \multirow{3}{*}{SRC} & \multirow{1}{*}{0.7666} & \multirow{1}{*}{UNLV-Dive} \\
        & & & & 0.8979 & MTL-AQA \\
        & & & & 0.8580 & FineDiving \\	

        \arrayrulecolor{black} \bottomrule [1.5pt]
    \end{tabular}
    \label{table9}
\end{table*}

Dadashzadeh et al.\cite{dadashzadeh2024pecop, dadashzadeh2024learning} proposed a parameter-efficient continual pre-training (\textbf{PECoP}) framework to address this problem. A lightweight convolutional bottleneck block, 3D-Adapter, is employed to conduct domain-specific self-supervised pre-training of I3D\cite{joao2017i3d}. Only the parameters of the 3D-Adapter are updated, while the I3D weights are maintained in a frozen state. However, on the target dataset, I3D is fine-tuned in conjunction with 3D-Adapter to align with the requirements of AQA. \textbf{Continual-AQA}\cite{li2024continual} contains two core components: feature-score correlation-aware rehearsal (FSCAR) and action general-specific graph (AGSG) modules. Upon receiving a new task, FSCAR first extracts representative samples from the finite storage of previous tasks and enhances the features and scores in order to maintain the continuity of the feature distribution. Then, AGSG combines the action features of the new task with the general and specific knowledge of previous tasks to extract discriminative scoring features that are consistent with the task. Finally, the model is trained on the new task in order to mitigate the issue of forgetting by integrating the features of the current and historical tasks. Similarly, Zhou et al.\cite{zhou2024magr} proposed manifold-aligned graph regularization (\textbf{MAGR}), which addresses the misalignment between static old features and the dynamically changing feature manifold causes severe catastrophic forgetting. MAGR aligns the old features with the current feature manifold, constructs graph regularization to maintain the consistency of the feature distribution and mass fraction distribution, and combines feature replay to retain the memory of old data while adapting to the new data distribution.

\subsubsection{Explainable AQA}\label{subsubsec4.2.5}
\begin{figure}[h]
    \centering
    \includegraphics[width=\columnwidth]{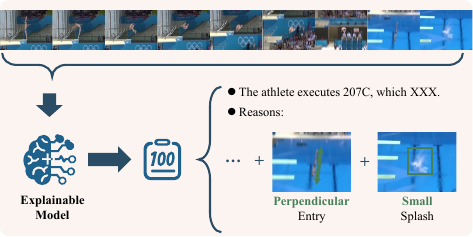}
    \caption{\textbf{Brief illustration of explainable AQA.}}
    \label{fig_trend_explainable}
\end{figure}

Interpretability is crucial in AQA to make systems trustworthy and adaptable to the wider population. For example, a community would not adopt an AI judge if it cannot explain how it arrived at a decision, especially if its decision was different than fellow human judges, or people would be hesitant to follow an AI physiotherapist if they cannot ``see" inner workings of this AI physiotherapists' brain. However, given that existing research is mainly based on end-to-end black box deep learning, interpretability is relatively challenging. Some researchers have attempted to use visualization or understandable neural symbols to increase the interpretability of models\cite{seo2021extracting, wang2021towards, hirosawa2022determinant, matsuyama2023iris, okamoto2024hierarchical, dong2024interpretable}. The brief illustration and performance of representative research is summarized in \autoref{fig_trend_explainable} and \autoref{table10}. 

\begin{table*}[!htbp]
\small
    \caption{Performance comparison of representative explainable methodologies.}
    \centering
    \begin{tabular}{@{}lccccl@{}}
        \toprule [1.5pt]
        \textbf{Model} & \textbf{Year} & \textbf{Format} & \textbf{Metric} & \textbf{Performance} & \textbf{Datasets}\\ \midrule [1pt]

        IRIS \cite{matsuyama2023iris} & 2023 & Regression & SRC & 0.8910 & MIT-Skate \\ [0.2cm]
        
        \arrayrulecolor{Grey0!30}  \midrule [0.5pt]

        NeuroSymbolic-AQA\cite{okamoto2024hierarchical} & 2024 & Regression & Expert Preference & 0.9610 & Finediving \\ [0.2cm]

        \arrayrulecolor{Grey0!30}  \midrule [0.5pt]
        
        \multirow{2}{*}{Interpretability-AQA\cite{dong2024interpretable}} & \multirow{2}{*}{2024} & \multirow{2}{*}{Regression} & \multirow{2}{*}{SRC} & \multirow{1}{*}{0.7880} & \multirow{1}{*}{Fis-V} \\
        & & & & 0.8420 & Rhy.Gym.\\ 

        \arrayrulecolor{black} \bottomrule [1.5pt]
    \end{tabular}
    \label{table10}
\end{table*}

Distinct from dominant black-box deep learning, Okamoto and Parmar\cite{okamoto2024hierarchical} introduced novel Hierarchical \textbf{NeuroSymbolic} paradigm (\textbf{NS-AQA})\cite{okamoto2024hierarchical}, which combines neural networks with \textbf{interpretable rule-based models}. Neural networks are used to deconstruct the visual scene of the performance by extracting interpretable symbols from video data, and rule-based models are constructed based on current rules of sport for action recognition and temporal segmentation. \textbf{Microprograms} are employed to assess fine-grained action quality at each stage, where a ranking-based percentage or fixed scoring formula is utilized to quantify the score of each action element, generating an overall score and detailed AQA reports, complete with visual evidence. Such an interpretable and flexible system also allows for personalized AQA, where users can modify the system on the fly to suit their needs. This formula provides a traceable and transparent scoring system that makes the assessment interpretable. Moreover, their interpretable system achieves SOTA performance on the tasks of finegrained \textbf{action recognition} and \textbf{temporal segmentation}. 

Dong et al.\cite{dong2024interpretable} proposed \textbf{Interpretability-AQA} which combines a novel attention loss function with a query-based Transformer decoder network to make sure the model focuses on keyframes and segments while avoiding temporal skipping. A weight-score regression module is used to simulate human scoring patterns, allowing each video segment's score to be explained in terms of different weights.

Matsuyama et al. \cite{matsuyama2023iris} proposed \textbf{IRIS} to address AQA in Figure Skating domain by leveraging rubrics to: 1) segment the action sequence; 2) compute technical element score differences of each segment relative to base scores; 3) compute multiple program component scores; and 4) compute the final score. They found that their approach offered interpretability while improving the performance on AQA task.

\subsubsection{Comprehensive AQA}\label{subsubsec4.2.6}
\begin{figure}[h]
    \centering
    \includegraphics[width=\columnwidth]{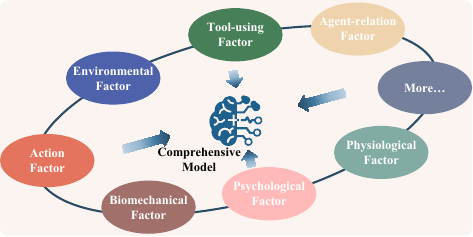}
    \caption{\textbf{Brief illustration of comprehensive AQA.}}
    \label{fig_trend_comprehensive}
\end{figure}

Action execution quality, in general, is composed of factors. Ideally, we want the models to take into consideration all these factors when quantifying the quality of an action. However, it was revealed that models do not take all the factors into consideration. This may be due to the fact that AQA datasets are trained on scores given out by human judges. Human judges are required to give out scores very quickly without any replays in most cases and domains. As such, the last part of the performance influences judges most, while the other factors are "forgotten". As a result, the human judges give out biased or non-comprehensive scores. Hence, machine learning models trained using such biased ground truth are also biased and non-comprehensive. Therefore, some researchers attempt to take more factors into account in addition to the action subject\cite{liu2021towards, wang2022will}. Although their research is still far from the ideal, there has been significant progress compared to existing research. The brief illustration and performance of representative research is summarized in \autoref{fig_trend_comprehensive} and \autoref{table11}. 

\begin{table*}[!htbp]
\small
    \caption{Performance comparison of representative comprehensive methodologies.}
    \centering
    \begin{tabular}{@{}lccccl@{}}
        \toprule [1.5pt]
        \textbf{Model} & \textbf{Year} & \textbf{Format} & \textbf{Metric} & \textbf{Performance} & \textbf{Datasets}\\ \midrule [1pt]

        VTPE\cite{liu2021towards} & 2021 & Regression & SRC & 0.8000 & JIGSAWS  \\ \arrayrulecolor{Grey0!30}  \midrule [0.5pt]

        \multirow{3}{*}{PSGCN-RTCN\cite{wang2022will}} & \multirow{3}{*}{2022} & \multirow{3}{*}{Regression} & \multirow{3}{*}{SRC} & \multirow{1}{*}{0.7200} & \multirow{1}{*}{MIT-Skate} \\
        & & & & 0.8800 & UNLV-Dive \\
        & & & & 0.7600 & UNLV-Vault \\ \midrule [0.5pt]

        NS-AQA \cite{okamoto2024hierarchical} & 2024 & Regression & Expert Preference & 0.9610 & FineDiving \\ 
        
        \arrayrulecolor{black} \bottomrule [1.5pt]
    \end{tabular}
    \label{table11}
\end{table*}

Liu et al.\cite{liu2021towards} proposed a unified multi-path framework for automatic surgical skill assessment (\textbf{VTPE}), which takes multiple aspects of surgical performance into consideration, including surgical tool usage, surgical field clearness, and surgical event pattern. Each path extracts features from different aspects and transfers them into skill score sequences. Moreover, a path dependency module models the interdependencies among these different aspects to provide weights of temporal importance for the score sequences. Lastly, the weighted score sequences are pooled over time and fused across paths as the final assessment prediction. Similarly, Wang et al.\cite{wang2022will} proposed to take the skeletal, holistic appearance, facial and scenic factors into consideration. Meanwhile, a relational temporal convolutional network (\textbf{RTCN}) and pyramidal skeleton GCN (\textbf{PSGCN}) are utilized to exploit appearance dynamics with non-local temporal relations and hierarchically refine human skeleton graphs, respectively. Finally, the adaptively learned attention weights for different streams according to the input video are used to fuse features and predict the score.  

\subsubsection{Self-supervised Representation Learning for AQA}\label{subsubsec4.2.7}
Annotating a dataset requires human efforts and financial resources. This is more so in the case of AQA because unlike in the case of object or action recognition, where untrained humans can easily annotate datasets, for AQA, trained experts are needed to assign a quality score to action samples. Trained experts/judges generally undergo elaborate training and get their licenses from governing bodies of the respective areas. This makes crowdsourcing of annotations infeasible. Datasets thus far have been created from Olympic footage. Unfortunately, a large amount of Olympic footage is unavailable for every action. Labels for action recognition samples can be obtained at a faster rate by mining video sites using keywords. However, compared to action recognition, more effort is required on the annotators' part to annotate the AQA dataset as the annotators actually need to go through the whole footage and stop and record scores given out by judges. To overcome the need for large labeled datasets, some research \cite{jain2019unsupervised, du2021assessing, roditakis2021towards, parmar2022domain, zhang2022semi, lee2022self, zhong2023contrastive, gedamu2024self, yun2024semi, zhong2024dancemvp} have explored self-supervised representation learning (SSL) for AQA.

\begin{figure}[h]
    \centering
    \includegraphics[width=\columnwidth]{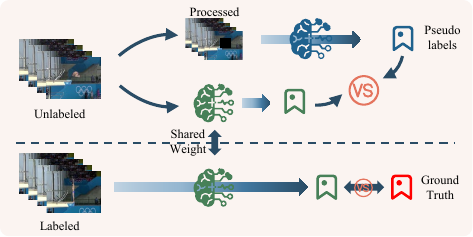}
    \caption{\textbf{Brief illustration of self-supervised representation learning for AQA.}}
    \label{fig_trend_ssl}
\end{figure}

The brief illustration and performance of representative research is summarized in \autoref{fig_trend_ssl} and \autoref{table12}. 

\begin{table*}[!htbp]
\small
    \caption{Performance comparison of representative self-supervised methodologies.}
    \centering
    \begin{tabular}{@{}lccccl@{}}
        \toprule [1.5pt]
        \textbf{Model} & \textbf{Year} & \textbf{Format} & \textbf{Metric} & \textbf{Performance} & \textbf{Datasets}\\ \midrule [1pt]

        Roditakis et al. \cite{roditakis2021towards} & 2021 & Regression & SRC & 0.7700 & MTL-AQA \\  \arrayrulecolor{Grey0!30} \midrule [0.5pt]
        
        Motion Disentangling \cite{parmar2022domain} & 2022 & Regression & SRC & 0.7763 & MTL-AQA \\  \arrayrulecolor{Grey0!30} \midrule [0.5pt]

        Pose Contrastive \cite{parmar2022domain} & 2022 & Classification & F1 Score & 0.7763 & MTL-AQA \\  \arrayrulecolor{Grey0!30} \midrule [0.5pt]

        \multirow{3}{*}{S$^4$AQA\cite{zhang2022semi}} & \multirow{3}{*}{2022} & \multirow{3}{*}{Regression} & \multirow{3}{*}{SRC} & \multirow{1}{*}{0.6550} & \multirow{1}{*}{JIGSAWS} \\
        & & & & 0.7460 & MTL-AQA \\
        & & & & 0.3420 & Rhy.Gym. \\ \midrule [0.5pt]

        SSL-CATE \cite{parmar2024learning} & 2024 & Regression & SRC & 0.7936 & MTL-AQA \\  \arrayrulecolor{Grey0!30} \midrule [0.5pt]

        \multirow{4}{*}{SAP-Net\cite{gedamu2024self}} & \multirow{4}{*}{2024} & \multirow{4}{*}{Regression} & \multirow{4}{*}{SRC} & \multirow{1}{*}{0.8010} & \multirow{1}{*}{MTL-AQA} \\
        & & & & 0.3930 & Rhy.Gym.\\
        & & & & 0.8450 & FineDiving \\ 
        & & & & 0.7090 & FineFS \\ \midrule [0.5pt]

        \multirow{3}{*}{TRS\cite{yun2024semi}} & \multirow{3}{*}{2024} & \multirow{3}{*}{Regression} & \multirow{3}{*}{SRC} & \multirow{1}{*}{0.7530} & \multirow{1}{*}{JIGSAWS} \\
        & & & & 0.9010 & MTL-AQA \\ 
        & & & & 0.5290 & Rhy.Gym. \\

        \arrayrulecolor{black} \bottomrule [1.5pt]
    \end{tabular}
    \label{table12}
\end{table*}

 Roditakis et al.\cite{roditakis2021towards} propose an SSL methodology based on temporal alignment pretext task. Parmar et al.\cite{parmar2022domain} propose a \textbf{Motion Disentangling (MD)} SSL methodology, where erroneous motion is isolated from global motion. Global motion is equivalent to the ideal way of executing actions; while erroneous motion refers to harmful deviations induced as a result of lower-quality execution. They also use domain knowledge in formulating pretext tasks. They show applicability to fitness actions and diving actions. Their methodology is applicable to other domains like diving as well. Parmar et al.\cite{parmar2022domain} also propose \textbf{Self-Supervised Pose-Contrastive} approach (CVCSPC) and \textbf{Self-Supervised Pose Disentangling} to learn robust pose sensitive representations from noisy in-the-wild videos. Parmar et al.\cite{parmar2024learning} propose a novel SSL methodology based on connecting actions and their effects (\textbf{SSL CATE}) to learn fine-grained, pose-sensitive representations from unlabeled videos.

Zhang et al.\cite{zhang2022semi} extended generalization research and took the lead in proposing self-supervised semi-supervised action quality assessment (\textbf{S$^4$AQA}), thereby alleviating the dependence of existing methods on a large amount of labeled data. S$^4$AQA employed multi-task learning by combining a self-supervised masked clip feature recovery task with a supervised score regression task. A representation distribution alignment module with a gradient inversion layer was used to align the representation distributions of labeled and unlabeled videos through adversarial training, enhancing the model's ability to generalize by efficiently utilizing unlabeled data. 

Gedamu et al.\cite{gedamu2024self} proposed a self-supervised sub-action parsing network (\textbf{SAP-Net}) that utilizes a teacher-student network structure to learn consistent semantic representations between labeled and unlabeled samples. Specifically, the teacher network generates high-quality pseudo-labels through actor-centric region detection, while the self-supervised sub-action parsing methodology disaggregates complex actions into fine-grained sub-action sequences. Additionally, SAP-Net employs a contrastive learning mechanism with pseudo-labels to ensure consistency in motion-oriented action features between the teacher and student branches.

Similarly, Yun et al.\cite{yun2024semi} proposed semi-supervised teacher-reference-student (\textbf{TRS}), where the teacher network generates pseudo-labels, the reference network provides supplementary supervised information, thereby facilitating the learning of unlabeled data by the student network. It is evident that introducing semi-supervised learning into AQA can mitigate the reliance of existing methodologies on extensive labeled data. By integrating limited labeled data with vast unlabeled data, models can more effectively discern the underlying structure and distribution of the data, thereby enhancing their robustness and generalization capabilities.

\subsection{Conclusion of Methodology}\label{subsec4.3}
Over the past decade, AQA has evolved from a niche research field to a thriving one, with a growing number of researchers coming into the field and proposing many novel research methodologies (see \autoref{fig_Num}). Consequently, AQA has progressed significantly, gradually evolving from fundamental coarse-grained research to more fine-grained, multimodal, generalized, interpretable, and comprehensive research, along with the incorporation of new research paradigms such as continuous learning and self-supervised learning. Research has gradually transitioned from a focus on single-task optimization to the integration of multi-dimensional methodologies and the expansion of application scenarios. However, there are still some limitations in existing research, while some challenges within AQA hinder its development (discussed in the following).

\section{Challenges \& Opportunities}\label{sec5}

As evidenced by the datasets (\autoref{sec3}) and methodologies (\autoref{sec4}), there have been dramatic advances in AQA over the past decade. Numerous datasets and publications have been proposed, particularly in 2023, with an increase of nearly 100\% compared to the previous (see \autoref{fig_Num}). Nevertheless, several challenges persist, hindering the advancement of AQA. Thus, to guide the community and stimulate advances in the field of AQA, in this section, we discuss challenges and corresponding future directions on three fronts: actions, datasets, and methodologies.

\subsection{Action}\label{subsec5.1}

\noindent\textbf{The Challenge of Inherent complexity.} Actions can be conscious or unconscious, natural or trained, but regardless of their type, they are imbued inherently with an individual intention. This means the same action command can be performed differently by various individuals. For instance, "run" can be executed fast, slow, or even with variations like jumping. In other words, a single action category may encompass numerous movement styles. Additionally, the same action can appear differently when captured from various views. This indicates significant inter- and intra-class differences in actions. Furthermore, the requirements for models vary across different action scenarios. To illustrate, competitive sports demand objective and highly interpretable models, whereas rehabilitation and medical care demand precision and long-term processing. 

Consequently, the inherent complexity of actions is considered the main challenge hindering the generalization of AQA to date. Despite the numerous research methodologies proposed, relatively few researchers have attempted to take on this challenge. The inherent complexity of the action is likened to an insurmountable barrier that stands before researchers.

\noindent\textbf{Gold Standard.} Owing to the action's inherent complexity, this has become the main challenge hindering model generalization. Therefore, some have proposed a framework for reference comparison\cite{yu2021group, li2022pairwise, an2024multi, ke2024two}, hoping by setting up simple comparison labels and abundant training data, the model can learn features autonomously from the latent space. Although this comparison methodology has achieved quite impressive results. However, regarding the selection of reference samples, researchers tend to choose samples with better performance within the dataset. This brings up the question: what samples can become gold standard actions?

The concept of the gold standard is widely used in medical diagnosis, referring to the most reliable diagnosis or assessment currently available and a benchmark for measuring other new methodologies. In the field of action quality assessment, what is the gold standard action? Is it the sample with the best performance in the dataset, the human with the best performance, or the sample created using generative models to be perfect? Much confusion exists in this matter, but one thing is worth mentioning: there is currently no sample that can be used as a gold standard. Therefore, a promising research direction in the future is to construct gold-standard action samples, which will become the benchmark for future research on actions, and all related research will align with this ground truth. Although this is not within the scope of computer science research, this will have a significant impact on AQA.

\subsection{Dataset}\label{subsec5.2}

\noindent\textbf{Challenges.} As illustrated in \autoref{table1}, there are presently 26 publicly available datasets in AQA encompassing 9 distinct domains. We can easily identify significant progress in existing datasets compared to pioneering datasets\cite{pirsiavash2014assessing,gao2014jhu}. However, challenges remain in terms of data scale, action diversity, data modality, and annotations compared to datasets in neighboring research fields\cite{joao2017i3d, carreira2018short}.

\begin{enumerate}[wide, labelwidth=!, labelindent=0pt,topsep=0pt]
    \item \textbf{Scant Scale.} Up to now, the biggest AQA dataset has more than 20000 samples\cite{parmar2022domain}. Different from the dataset for object detection, which was annotated by searching for suitable photos in a gallery and boxing them. Real-world AQA datasets involve collecting/selecting videos in a specific action domain and labeling them by domain experts. The collection/selection and annotation consume a huge amount of time and financial costs. As a result, the construction of a big-scale, real-world AQA dataset is still a challenge that remains to be addressed. While scant data astrict the possible methodologies researchers are capable of using and the performance.
    \item \textbf{Narrow Action.} As previously stated, existing datasets are limited in data sources, while construction is time-consuming and labor-intensive, which results in a narrow action category. To address this challenge, many researchers divide large limb actions into different sub-actions, hoping to enhance the inter-category differences\cite{xu2022finediving}. However, this approach does not effectively expand the action categories. Additionally, due to limited data sources, the variation in action categories included in different datasets is minimal.
    \item \textbf{Coarse Annotation.} The earliest datasets were limited by the data sources, and to save time on data annotation, the final scores of the athletes were usually used as labels directly. Obviously, a single score cannot allow the model to learn and utilize features sufficiently. Thus, researchers proposed to use multiple judges' scores, event commentary, etc., to improve the annotations, but a more refined annotation could not be achieved. The limitation of fine-grained labels has indeed restricted the advancement of AQA models in interpretability.
\end{enumerate}

\noindent\textbf{Future directions.} Obviously, the challenges on the dataset front remain in terms of data scale, action diversity, data modality, and annotations. Therefore, future directions will mainly revolve around these. However, the implementation details differ from other surveys' \cite{liu2024vision, zhou2024comprehensive}, which tend to employ AQA to evaluate the quality of generated actions and utilize generative models for data argumentation.

\begin{enumerate}[wide, labelwidth=!, labelindent=0pt,topsep=0pt]
    \item \textbf{Large-scale multi-action real-world dataset.} Despite the fact that we know the construction of datasets is a time-consuming and laborious process, with huge challenges. However, a large-scale multi-action real-world dataset is of great significance to AQA, which can not only promote the model to extract the commonalities between different actions to achieve higher generalization but also promote the model's ability to comprehensively assess within a specific action domain comrehensively.
   
    \item \textbf{Large-scale multi-action high-quality AIGC dataset.} Derived from the development of the recent T2V model, there is now a new paradigm for obtaining videos beyond collection. This has also inspired researchers to consider whether the T2V model can be used to generate videos in specific action domains instead of the currently used data set construction methodology to reduce consumption. GAIA\cite{chen2024gaia}, as pioneering research, introduced the first non-real-world dataset. However, we can also see (\autoref{fig_dataset}(h)) that there are many deficiencies, such as incomplete human body generation, incoherent actions, and the absence of physical laws. However, the introduction of physical laws in Genesis\cite{Genesis} may indeed achieve more realistic results. Meanwhile, video prompts can also be used as a fine-grained description of the video, which further reduces the difficulty and time-consuming of annotation. Therefore, we believe that one of the future directions of the dataset is to construct a large-scale, multi-action, high-quality AIGC dataset.
\end{enumerate}

\subsection{Methodology}\label{subsec5.3}

\noindent\textbf{Challenges}. Combine the methodologies described in \autoref{sec4} with performance comparisons. A significant methodological improvement in AQA over the past decade can be found, especially in the extraction of detailed features such as spatio-temporal modeling and skeletal relationships, where various novel network structures have been applied. However, this also brings many problems, especially in feedback, interpretability, and multimodal robustness, which remain unsolved challenges.

\begin{enumerate}[wide, labelwidth=!, labelindent=0pt,topsep=0pt]
\item\textbf{Computational Complexity \& Latency.} With the continuous advancement of network structures, models can extract features with more detail and achieve higher performance. However, this also requires powerful computing power to support model training, and even during the inference stage, it requires considerable computing resources. At the same time, existing datasets and methodologies are aimed at the quality assessment of the complete action process. Therefore, due to various factors, fewer researchers have proposed real-time and lightweight computing models, for example, \cite{parmar2021hallucinet, okamoto2024hierarchical}. 

\item\textbf{Blackbox models.} Interpretability of deep learning has always been a hot topic, especially in the field of AQA. Given that the purpose of AQA since its creation has been to eliminate personal bias and achieve objective action evaluation, as well as the fact that the main application scenario of competitive sports requires fairness, interpretability is one of the critical concerns of AQA models. However, due to the scant data scale and coarse annotation, interpretability has always been a difficult challenge to solve.

\item\textbf{Fragile Robustness} 
Many methodologies in AQA combine multimodal data to realize more accurate quality assessment. Researchers tend to focus on how to efficiently combine data from different modalities while ignoring the performance research of some modalities that are absent due to actual environmental conditions in practical applications. Therefore, ensuring the accuracy of multimodal models when some modalities are absent is also a challenge.
\end{enumerate}

\noindent\textbf{Future directions.} Despite the considerable challenges associated with existing models in terms of real-time feedback, interpretability, and robustness. However, adversity brings opportunities, which also indicate future directions of methodologies.

\begin{enumerate}[wide, labelwidth=!, labelindent=0pt,topsep=0pt]
\item\textbf{Lightweight, Efficient, Real-time.} Only a handful of AQA research has focused on developing lightweight, efficient, and real-time solutions \cite{parmar2021hallucinet, okamoto2024hierarchical}. One of the biggest areas in which the AQA can make a difference is improving the lives of socio-economically disadvantaged communities through providing automated AQA/physiotherapy/training solutions. Such communities cannot be expected to have high-end computational resources to run heavy AQA apps. Thus, we believe one of the future directions should focus on developing real-time AQA solutions, which are also efficient in terms of computing power and communication bandwidth requirements.
\item\textbf{Interpretability.} Interpretability is currently the main challenge at the methodological level. Researchers have tried various methods to improve the interpretability of models, most of which attempt to use visualization techniques to make the decision-making process public rather than optimizing the network structure. Given the current rise of neural symbolic computing\cite{wang2024nssurvey}, researchers have attempted to improve interpretability using symbols that humans can understand, with reasonable achievements\cite{okamoto2024hierarchical}. In addition, it is believed that the study of causal\cite{guo2020casualitysurvey} in medical imaging\cite{castro2020causalitymed} can be used to construct a causal relationship between features and quality, which can make the decision-making process transparent and credible. Thus, one of the future directions is to use neural symbolic computing or causal inference to improve interpretability and make the decision-making process transparent and credible.

\item\textbf{Robustness.} Most multimodal methods ignore the performance under modal missing. Some researchers\cite{parmar2019and, parsa2021multi, li2024finerehab} choose to use multi-task learning to mitigate the impact of modal missing, which will lead to partial output results. The future research direction of multimodality should not only consider the integration of cross-modal data but also the cross-modal data augmentation and prediction. Existing modalities can fill in the absent modalities to reduce the impact on the model. At the same time, the data mapping relationship between different modalities can be established.
\end{enumerate}

\section{Conclusion}\label{sec6}
Action Quality Assessment, an emerging and critical field in video understanding, has garnered increasing attention in the last decade, and AQA has made significant progress. To consolidate the research efforts in the area thus far, we present the largest and most comprehensive survey to date, to the best of our knowledge, on AQA. Towards this end, we systematically reviewed over 200 relevant papers that met the criteria of the PRISMA framework, providing a detailed description of the definition, classification of AQA problems, model evaluation metrics, and datasets. We distilled research trends from these papers and organized them into 7 principal trends, with detailed discussions on the research methodologies and performance comparisons within each trend. Furthermore, we analyzed the challenges in the field of AQA and identified areas and research opportunities that future work could tackle.

\section*{Acknowledgement}
This work is supported by Youth Innovation Promotion Association of the Chinese Academy of Sciences, Grant/Award Number: E1290301.

\section*{Data Availability}
This article is a survey of existing publications related to action quality assessment. No new datasets or codes were generated during the current study. All data presented or discussed are derived from publicly available sources, which are cited throughout the manuscript.


\begin{thebibliography}{100}

\bibitem{kong2022human}
Yu~Kong and Yun Fu.
\newblock Human action recognition and prediction: A survey.
\newblock {\em International Journal of Computer Vision}, 130(5):1366--1401,
  2022.

\bibitem{pirsiavash2014assessing}
Hamed Pirsiavash, Carl Vondrick, and Antonio Torralba.
\newblock Assessing the quality of actions.
\newblock In {\em European Conference on Computer Vision}, pages 556--571.
  Springer, 2014.

\bibitem{parmar2017learning}
Paritosh Parmar and Brendan Tran~Morris.
\newblock Learning to score olympic events.
\newblock In {\em Proceedings of the IEEE Conference on Computer Vision and
  Pattern Recognition Workshops}, pages 20--28, 2017.

\bibitem{gao2014jhu}
Yixin Gao, S~Swaroop Vedula, Carol~E Reiley, Narges Ahmidi, Balakrishnan
  Varadarajan, Henry~C Lin, Lingling Tao, Luca Zappella, Benjamın Béjar, and
  David~D Yuh.
\newblock Jhu-isi gesture and skill assessment working set (jigsaws): A
  surgical activity dataset for human motion modeling.
\newblock In {\em Medical Image Computing and Computer Assisted Intervention
  Workshop}, volume~3, page~3, 2014.

\bibitem{vakanski2018data}
Aleksandar Vakanski, Hyung-pil Jun, David Paul, and Russell Baker.
\newblock A data set of human body movements for physical rehabilitation
  exercises.
\newblock {\em Data}, 3(1):2, 2018.

\bibitem{capecci2019kimore}
Marianna Capecci, Maria~Gabriella Ceravolo, Francesco Ferracuti, Sabrina
  Iarlori, Andrea Monteriu, Luca Romeo, and Federica Verdini.
\newblock The kimore dataset: Kinematic assessment of movement and clinical
  scores for remote monitoring of physical rehabilitation.
\newblock {\em IEEE Transactions on Neural Systems and Rehabilitation
  Engineering}, 27(7):1436--1448, 2019.

\bibitem{li2019automated}
Ruimin Li, Hong Fu, Yang Zheng, Wai-Lun Lo, Jane J. Yu, Cindy H. P. Sit, Zheru Chi, Zongxi Song, and Desheng Wen.
\newblock Automated fine motor evaluation for developmental coordination disorder.
\newblock {\em IEEE Transactions on Neural Systems and Rehabilitation Engineering}, 27(5):963--973,2019.

\bibitem{parmar2022domain}
Paritosh Parmar, Amol Gharat, and Helge Rhodin.
\newblock Domain knowledge-informed self-supervised representations for workout
  form assessment.
\newblock In {\em European Conference on Computer Vision}, pages 105--123.
  Springer, 2022.

\bibitem{li2024egoexo}
Yuan-Ming Li, Wei-Jin Huang, An-Lan Wang, Ling-An Zeng, Jing-Ke Meng, and
  Wei-Shi Zheng.
\newblock Egoexo-fitness: Towards egocentric and exocentric full-body action
  understanding.
\newblock In {\em European Conference on Computer Vision}, 2024.

\bibitem{gu2024exechecker}
Yiwen Gu, Mahir Patel, and Margrit Betke.
\newblock ExeChecker: Where Did I Go Wrong?
\newblock {\em arXiv preprint arXiv:2412.10573}, 2024.

\bibitem{sener2022assembly101}
Fadime Sener, Dibyadip Chatterjee, Daniel Shelepov, Kun He, Dipika Singhania,
  Robert Wang, and Angela Yao.
\newblock Assembly101: A large-scale multi-view video dataset for understanding
  procedural activities.
\newblock In {\em Proceedings of the IEEE/CVF Conference on Computer Vision and
  Pattern Recognition}, pages 21096--21106, 2022.

\bibitem{seminara2024differentiable}
Luigi Seminara, Giovanni Maria, and Farinella Antonino Furnari.
\newblock Differentiable Task Graph Learning: Procedural Activity Representation and Online Mistake Detection from Egocentric Videos.
\newblock In {\em arXiv preprint arXiv:2406.01486}, 2024.

\bibitem{chen2024gaia}
Zijian Chen, Wei Sun, Yuan Tian, Jun Jia, Zicheng Zhang, Jiarui Wang, Ru~Huang,
  Xiongkuo Min, Guangtao Zhai, and Wenjun Zhang.
\newblock Gaia: Rethinking action quality assessment for ai-generated videos.
\newblock In {\em Advances in Neural Information Processing Systems}, 2024.

\bibitem{parmar2016measuring}
Paritosh Parmar and Brendan~Tran Morris.
\newblock Measuring the quality of exercises.
\newblock In {\em International Conference of the IEEE Engineering in Medicine
  and Biology Society}, pages 2241--2244. IEEE, 2016.

\bibitem{nespolo2024assessing}
Rogerio Nespolo, George R Nahass, Mahtab Faraji, Darvin Yi, Yannek Isaac Leiderman.
\newblock Assessing vitreoretinal surgical training experience by leveraging instrument maneuvers and visual attention with deep learning neural networks.
\newblock In {\em Investigative Ophthalmology \& Visual Science}, 65(7):900, 2024.

\bibitem{doughty2018s}
Hazel Doughty, Dima Damen, and Walterio Mayol-Cuevas.
\newblock Who's better? who's best? pairwise deep ranking for skill
  determination.
\newblock In {\em Proceedings of the IEEE Conference on Computer Vision and
  Pattern Recognition}, pages 6057--6066, 2018.

\bibitem{doughty2019pros}
Hazel Doughty, Walterio Mayol-Cuevas, and Dima Damen.
\newblock The pros and cons: Rank-aware temporal attention for skill
  determination in long videos.
\newblock In {\em Proceedings of the IEEE/CVF Conference on Computer Vision and
  Pattern Recognition}, pages 7862--7871, 2019.
  
\bibitem{lei2019survey}
Qing Lei, Ji-Xiang Du, Hong-Bo Zhang, Shuang Ye, and Duan-Sheng Chen.
\newblock A survey of vision-based human action evaluation methods.
\newblock {\em Sensors}, 19(19):4129, 2019.

\bibitem{wang2021survey}
Shunli Wang, Dingkang Yang, Peng Zhai, Qing Yu, Tao Suo, Zhan Sun, Ka~Li, and
  Lihua Zhang.
\newblock A survey of video-based action quality assessment.
\newblock In {\em International Conference on Networking Systems of AI}, pages
  1--9. IEEE, 2021.

\bibitem{page2021prisma}
Matthew~J Page, Joanne~E McKenzie, Patrick~M Bossuyt, Isabelle Boutron, Tammy~C
  Hoffmann, Cynthia~D Mulrow, Larissa Shamseer, Jennifer~M Tetzlaff, Elie~A
  Akl, Sue~E Brennan, et~al.
\newblock The prisma 2020 statement: an updated guideline for reporting
  systematic reviews.
\newblock {\em BMJ}, 372, 2021.

\bibitem{parmar2019and}
Paritosh Parmar and Brendan~Tran Morris.
\newblock What and how well you performed? a multitask learning approach to
  action quality assessment.
\newblock In {\em Proceedings of the IEEE/CVF Conference on Computer Vision and
  Pattern Recognition}, pages 304--313, 2019.

\bibitem{wang2024cpr}
Shunli Wang, Shuaibing Wang, Dingkang Yang, Mingcheng Li, Haopeng Kuang, Xiao
  Zhao, Liuzhen Su, Peng Zhai, and Lihua Zhang.
\newblock Cpr-coach: Recognizing composite error actions based on single-class
  training.
\newblock In {\em Proceedings of the IEEE/CVF Conference on Computer Vision and
  Pattern Recognition}, pages 18782--18792, 2024.

\bibitem{zeng2024multimodal}
Ling-An Zeng and Wei-Shi Zheng.
\newblock Multimodal action quality assessment.
\newblock {\em IEEE Transactions on Image Processing}, 2024.

\bibitem{ogata2019temporal}
Ryoji Ogata, Edgar Simo-Serra, Satoshi Iizuka, and Hiroshi Ishikawa.
\newblock Temporal distance matrices for squat classification.
\newblock In {\em Proceedings of the IEEE/CVF Conference on Computer Vision and
  Pattern Recognition Workshops}, pages 0--0, 2019.

\bibitem{li2021skeleton}
Hui-Ying Li, Qing Lei, Hong-Bo Zhang, and Ji-Xiang Du.
\newblock Skeleton based action quality assessment of figure skating videos.
\newblock In {\em International Conference on Information Technology in
  Medicine and Education}, pages 196--200. IEEE, 2021.

\bibitem{okamoto2024hierarchical}
Lauren Okamoto and Paritosh Parmar.
\newblock Hierarchical neurosymbolic approach for comprehensive and explainable
  action quality assessment.
\newblock In {\em Proceedings of the IEEE/CVF Conference on Computer Vision and
  Pattern Recognition}, pages 3204--3213, 2024.

\bibitem{parmar2021piano}
Paritosh Parmar, Jaiden Reddy, and Brendan Morris.
\newblock Piano skills assessment.
\newblock In {\em IEEE International Workshop on Multimedia Signal Processing},
  pages 1--5. IEEE, 2021.

\bibitem{hipiny2023danced}
Irwandi Hipiny, Hamimah Ujir, Aidil~Azli Alias, Musdi Shanat, and
  Mohamad~Khairi Ishak.
\newblock Who danced better? ranked tiktok dance video dataset and pairwise
  action quality assessment method.
\newblock {\em International Journal of Advances in Intelligent Informatics},
  9(1):96--107, 2023.

\bibitem{bruce2024egcn++}
XB~Bruce, Yan Liu, Keith~CC Chan, and Chang~Wen Chen.
\newblock Egcn++: A new fusion strategy for ensemble learning in skeleton-based
  rehabilitation exercise assessment.
\newblock {\em IEEE Transactions on Pattern Analysis and Machine Intelligence},
  2024.

\bibitem{bertasius2017baller}
Gedas Bertasius, Hyun Soo~Park, Stella~X Yu, and Jianbo Shi.
\newblock Am i a baller? basketball performance assessment from first-person
  videos.
\newblock In {\em Proceedings of the IEEE International Conference on Computer
  Vision}, pages 2177--2185, 2017.

\bibitem{yu2021group}
Xumin Yu, Yongming Rao, Wenliang Zhao, Jiwen Lu, and Jie Zhou.
\newblock Group-aware contrastive regression for action quality assessment.
\newblock In {\em Proceedings of the IEEE/CVF International Conference on
  Computer Vision}, pages 7919--7928, 2021.

\bibitem{parmar2019action}
Paritosh Parmar and Brendan Morris.
\newblock Action quality assessment across multiple actions.
\newblock In {\em Proceedings of the IEEE/CVF Winter Conference on Applications
  of Computer Vision}, pages 1468--1476. IEEE, 2019.

\bibitem{xu2019learning}
Chengming Xu, Yanwei Fu, Bing Zhang, Zitian Chen, Yu-Gang Jiang, and Xiangyang
  Xue.
\newblock Learning to score figure skating sport videos.
\newblock {\em IEEE Transactions on Circuits and Systems for Video Technology},
  30(12):4578--4590, 2019.

\bibitem{gao2020asymmetric}
Jibin Gao, Wei-Shi Zheng, Jia-Hui Pan, Chengying Gao, Yaowei Wang, Wei Zeng,
  and Jianhuang Lai.
\newblock An asymmetric modeling for action assessment.
\newblock In {\em European Conference on Computer Vision}, pages 222--238.
  Springer, 2020.

\bibitem{zeng2020hybrid}
Ling-An Zeng, Fa-Ting Hong, Wei-Shi Zheng, Qi-Zhi Yu, Wei Zeng, Yao-Wei Wang,
  and Jian-Huang Lai.
\newblock Hybrid dynamic-static context-aware attention network for action
  assessment in long videos.
\newblock In {\em Proceedings of the ACM International Conference on
  Multimedia}, pages 2526--2534, 2020.

\bibitem{wang2021tsa}
Shunli Wang, Dingkang Yang, Peng Zhai, Chixiao Chen, and Lihua Zhang.
\newblock Tsa-net: Tube self-attention network for action quality assessment.
\newblock In {\em Proceedings of the ACM International Conference on
  Multimedia}, pages 4902--4910, 2021.

\bibitem{chen2021sportscap}
Xin Chen, Anqi Pang, Wei Yang, Yuexin Ma, Lan Xu, and Jingyi Yu.
\newblock Sportscap: Monocular 3d human motion capture and fine-grained
  understanding in challenging sports videos.
\newblock {\em International Journal of Computer Vision}, 129:2846--2864, 2021.

\bibitem{xu2022finediving}
Jinglin Xu, Yongming Rao, Xumin Yu, Guangyi Chen, Jie Zhou, and Jiwen Lu.
\newblock Finediving: A fine-grained dataset for procedure-aware action quality
  assessment.
\newblock In {\em Proceedings of the IEEE/CVF Conference on Computer Vision and
  Pattern Recognition}, pages 2949--2958, 2022.

\bibitem{zhang2023logo}
Shiyi Zhang, Wenxun Dai, Sujia Wang, Xiangwei Shen, Jiwen Lu, Jie Zhou, and
  Yansong Tang.
\newblock Logo: A long-form video dataset for group action quality assessment.
\newblock In {\em Proceedings of the IEEE/CVF Conference on Computer Vision and
  Pattern Recognition}, pages 2405--2414, 2023.

\bibitem{ji2023localization}
Yanli Ji, Lingfeng Ye, Huili Huang, Lijing Mao, Yang Zhou, and Lingling Gao.
\newblock Localization-assisted uncertainty score disentanglement network for
  action quality assessment.
\newblock In {\em Proceedings of the ACM International Conference on
  Multimedia}, pages 8590--8597, 2023.

\bibitem{gao2023automatic}
Jibin Gao, Jia-Hui Pan, Shao-Jie Zhang, and Wei-Shi Zheng.
\newblock Automatic modelling for interactive action assessment.
\newblock {\em International Journal of Computer Vision}, 131(3):659--679,
  2023.

\bibitem{frade2008mmac}
Fernando~De la~Torre~Frade, Jessica~K. Hodgins, Adam~W. Bargteil, Xavier~Martin
  Artal, Justin~C. Macey, Alexandre Collado~I Castells, and Josep Beltran.
\newblock Guide to the carnegie mellon university multimodal activity
  (cmu-mmac) database.
\newblock Technical Report CMU-RI-TR-08-22, Pittsburgh, PA, April 2008.

\bibitem{xu2024procedure}
Jinglin Xu, Yongming Rao, Jie Zhou, and Jiwen Lu.
\newblock Procedure-aware action quality assessment: Datasets and performance
  evaluation.
\newblock {\em International Journal of Computer Vision}, pages 1--22, 2024.

\bibitem{ionescu2013human3}
Catalin Ionescu, Dragos Papava, Vlad Olaru, and Cristian Sminchisescu.
\newblock Human3. 6m: Large scale datasets and predictive methods for 3d human
  sensing in natural environments.
\newblock {\em IEEE Transactions on Pattern Analysis and Machine Intelligence},
  36(7):1325--1339, 2013.

\bibitem{kay2017kinetics}
Will Kay, Joao Carreira, Karen Simonyan, Brian Zhang, Chloe Hillier, Sudheendra
  Vijayanarasimhan, Fabio Viola, Tim Green, Trevor Back, Paul Natsev, et~al.
\newblock The kinetics human action video dataset.
\newblock {\em arXiv preprint arXiv:1705.06950}, 2017.

\bibitem{liu2020fsd}
Shenglan Liu, Xiang Liu, Gao Huang, Hong Qiao, Lianyu Hu, Dong Jiang, Aibin
  Zhang, Yang Liu, and Ge~Guo.
\newblock Fsd-10: A fine-grained classification dataset for figure skating.
\newblock {\em Neurocomputing}, 413:360--367, 2020.

\bibitem{sardari2020vi}
Faegheh Sardari, Adeline Paiement, Sion Hannuna, and Majid Mirmehdi.
\newblock Vi-net—view-invariant quality of human movement assessment.
\newblock {\em Sensors}, 20(18):5258, 2020.

\bibitem{tang2020comprehensive}
Yansong Tang, Jiwen Lu, and Jie Zhou.
\newblock Comprehensive instructional video analysis: The coin dataset and
  performance evaluation.
\newblock {\em IEEE Transactions on Pattern Analysis and Machine Intelligence},
  43(9):3138--3153, 2020.

\bibitem{moodley2022casa}
Tevin Moodley and Dustin van~der Haar.
\newblock Casa: Cricket action similarity assessment in video footage using
  deep metric learning.
\newblock In {\em Southern African Conference for Artificial Intelligence
  Research}, pages 139--153. Springer, 2022.

\bibitem{parmar2022win}
Paritosh Parmar and Brendan Morris.
\newblock Win-fail action recognition.
\newblock In {\em Proceedings of the IEEE/CVF Winter Conference on Applications
  of Computer Vision}, pages 161--171, 2022.

\bibitem{xing2022functional}
Qing-Jun Xing, Yuan-Yuan Shen, Run Cao, Shou-Xin Zong, Shu-Xiang Zhao, and
  Yan-Fei Shen.
\newblock Functional movement screen dataset collected with two azure kinect
  depth sensors.
\newblock {\em Scientific Data}, 9(1):104, 2022.

\bibitem{liu2023fine}
Sheng-Lan Liu, Yu-Ning Ding, Si-Fan Zhang, Wen-Yue Chen, Ning Zhou, Hao Liu,
  and Gui-Hong Lao.
\newblock Fine-grained action analysis: A multi-modality and multi-task dataset
  of figure skating.
\newblock {\em arXiv preprint arXiv:2307.02730}, 2023.

\bibitem{tang2023flag3d}
Yansong Tang, Jinpeng Liu, Aoyang Liu, Bin Yang, Wenxun Dai, Yongming Rao,
  Jiwen Lu, Jie Zhou, and Xiu Li.
\newblock Flag3d: A 3d fitness activity dataset with language instruction.
\newblock In {\em Proceedings of the IEEE/CVF Conference on Computer Vision and
  Pattern Recognition}, pages 22106--22117, 2023.

\bibitem{fang2024better}
Ming Fang, Xinning Du, Qi~Liu, Yunpeng Zhou, Qiwen Liang, and Shuhua Liu.
\newblock Which is the better teacher action? a new ranking model and dataset.
\newblock In {\em IEEE International Conference on Acoustics, Speech and Signal
  Processing}, pages 7695--7699. IEEE, 2024.

\bibitem{gan2024skatingverse}
Ziliang Gan, Lei Jin, Yi~Cheng, Yu~Cheng, Yinglei Teng, Zun Li, Yawen Li,
  Wenhan Yang, Zheng Zhu, and Junliang Xing.
\newblock Skatingverse: A large‐scale benchmark for comprehensive evaluation
  on human action understanding.
\newblock {\em IET Computer Vision}, 2024.

\bibitem{li2024finerehab}
Jianwei Li, Jun Xue, Rui Cao, Xiaoxia Du, Siyu Mo, Kehao Ran, and Zeyan Zhang.
\newblock Finerehab: A multi-modality and multi-task dataset for rehabilitation
  analysis.
\newblock In {\em Proceedings of the IEEE/CVF Conference on Computer Vision and
  Pattern Recognition}, pages 3184--3193, 2024.

\bibitem{nagai2024mmw}
Takasuke Nagai, Shoichiro Takeda, Satoshi Suzuki, and Hitoshi Seshimo.
\newblock Mmw-aqa: Multimodal in-the-wild dataset for action quality
  assessment.
\newblock {\em IEEE Access}, 2024.

\bibitem{venkataraman2015dynamical}
Vinay Venkataraman, Ioannis Vlachos, and Pavan~K Turaga.
\newblock Dynamical regularity for action analysis.
\newblock In {\em BMVC}, volume~67, pages 1--12, 2015.

\bibitem{gordon1995automated}
Andrew~S Gordon.
\newblock Automated video assessment of human performance.
\newblock In {\em Proceedings of AI-ED}, volume~2, pages 541--546, 1995.

\bibitem{jug2003trajectory}
Marko Jug, Janez Per{\v{s}}, Branko De{\v{z}}man, and Stanislav
  Kova{\v{c}}i{\v{c}}.
\newblock Trajectory based assessment of coordinated human activity.
\newblock In {\em International Conference on Computer Vision Systems}, pages
  534--543. Springer, 2003.

\bibitem{wnuk2010analyzing}
Kamil Wnuk and Stefano Soatto.
\newblock Analyzing diving: A dataset for judging action quality.
\newblock In {\em Asian conference on computer vision}, pages 266--276.
  Springer, 2010.

\bibitem{chang2011kinect}
Yao-Jen Chang, Shu-Fang Chen, and Jun-Da Huang.
\newblock A kinect-based system for physical rehabilitation: A pilot study for
  young adults with motor disabilities.
\newblock {\em Research in Developmental Disabilities}, 32(6):2566--2570, 2011.

\bibitem{tao2012hmm}
Lingling Tao, Ehsan Elhamifar, Sanjeev Khudanpur, Gregory~D Hager, and Ren{\'e}
  Vidal.
\newblock Sparse hidden markov models for surgical gesture classification and
  skill evaluation.
\newblock In {\em Information Processing in Computer-Assisted Interventions},
  pages 167--177. Springer, 2012.

\bibitem{tran2015c3d}
Du~Tran, Lubomir Bourdev, Rob Fergus, Lorenzo Torresani, and Manohar Paluri.
\newblock Learning spatiotemporal features with 3d convolutional networks.
\newblock In {\em Proceedings of the IEEE International Conference on Computer
  Vision}, pages 4489--4497, 2015.

\bibitem{lstm}
Sepp Hochreiter and J\"{u}rgen Schmidhuber.
\newblock Long short-term memory.
\newblock {\em Neural Comput.}, 9(8):1735–1780, November 1997.

\bibitem{joao2017i3d}
Joao Carreira and Andrew Zisserman.
\newblock Quo vadis, action recognition? a new model and the kinetics dataset.
\newblock In {\em Proceedings of the IEEE Conference on Computer Vision and
  Pattern Recognition}, pages 6299--6308, 2017.

\bibitem{carreira2018short}
Joao Carreira, Eric Noland, Andras Banki-Horvath, Chloe Hillier, and Andrew
  Zisserman.
\newblock A short note about kinetics-600.
\newblock {\em arXiv preprint arXiv:1808.01340}, 2018.

\bibitem{he2016resnet}
Kaiming He, Xiangyu Zhang, Shaoqing Ren, and Jian Sun.
\newblock Deep residual learning for image recognition.
\newblock In {\em Proceedings of the IEEE Conference on Computer Vision and
  Pattern Recognition}, pages 770--778, 2016.

\bibitem{qiu2017p3d}
Zhaofan Qiu, Ting Yao, and Tao Mei.
\newblock Learning spatio-temporal representation with pseudo-3d residual
  networks.
\newblock In {\em Proceedings of the IEEE International Conference on Computer
  Vision}, pages 5533--5541, 2017.

\bibitem{liu2022vst}
Ze~Liu, Jia Ning, Yue Cao, Yixuan Wei, Zheng Zhang, Stephen Lin, and Han Hu.
\newblock Video swin transformer.
\newblock In {\em Proceedings of the IEEE/CVF Conference on Computer Vision and
  Pattern Recognition}, pages 3202--3211, 2022.

\bibitem{doughty2021skill}
Hazel~R Doughty.
\newblock {\em Skill determination from long videos}.
\newblock Thesis, 2021.

\bibitem{xiang2018s3d}
Xiang Xiang, Ye~Tian, Austin Reiter, Gregory~D Hager, and Trac~D Tran.
\newblock S3d: Stacking segmental p3d for action quality assessment.
\newblock In {\em IEEE International Conference on Image Processing}, pages
  928--932. IEEE, 2018.

\bibitem{dong2021learning}
Li-Jia Dong, Hong-Bo Zhang, Qinghongya Shi, Qing Lei, Ji-Xiang Du, and Shangce
  Gao.
\newblock Learning and fusing multiple hidden substages for action quality
  assessment.
\newblock {\em Knowledge-Based Systems}, 229:107388, 2021.

\bibitem{lei2021temporal}
Qing Lei, Hongbo Zhang, and Jixiang Du.
\newblock Temporal attention learning for action quality assessment in sports
  video.
\newblock {\em Signal, Image and Video Processing}, 15(7):1575--1583, 2021.

\bibitem{xu2022likert}
Angchi Xu, Ling-An Zeng, and Wei-Shi Zheng.
\newblock Likert scoring with grade decoupling for long-term action assessment.
\newblock In {\em Proceedings of the IEEE/CVF Conference on Computer Vision and
  Pattern Recognition}, pages 3232--3241, 2022.

\bibitem{iyer2022action}
Abhay Iyer, Mohammad Alali, Hemanth Bodala, and Sunit Vaidya.
\newblock Action quality assessment using transformers.
\newblock {\em arXiv preprint arXiv:2207.12318}, 2022.

\bibitem{bai2022action}
Yang Bai, Desen Zhou, Songyang Zhang, Jian Wang, Errui Ding, Yu~Guan, Yang
  Long, and Jingdong Wang.
\newblock Action quality assessment with temporal parsing transformer.
\newblock In {\em European Conference on Computer Vision}, pages 422--438.
  Springer, 2022.

\bibitem{zhang2022learning}
Yu~Zhang, Wei Xiong, and Siya Mi.
\newblock Learning time-aware features for action quality assessment.
\newblock {\em Pattern Recognition Letters}, 158:104--110, 2022.

\bibitem{gedamu2023fine}
Kumie Gedamu, Yanli Ji, Yang Yang, Jie Shao, and Heng~Tao Shen.
\newblock Fine-grained spatio-temporal parsing network for action quality
  assessment.
\newblock {\em IEEE Transactions on Image Processing}, 32:6386--6400, 2023.

\bibitem{lian2023improving}
Pu-Xiang Lian and Zhi-Gang Shao.
\newblock Improving action quality assessment with across-staged temporal
  reasoning on imbalanced data.
\newblock {\em Applied Intelligence}, 53(24):30443--30454, 2023.

\bibitem{liu2023figure}
Yanchao Liu, Xina Cheng, and Takeshi Ikenaga.
\newblock A figure skating jumping dataset for replay-guided action quality
  assessment.
\newblock In {\em Proceedings of the ACM International Conference on
  Multimedia}, pages 2437--2445, 2023.

\bibitem{zhang2023labe}
Hong-Bo Zhang, Li-Jia Dong, Qing Lei, Li-Jie Yang, and Ji-Xiang Du.
\newblock Label-reconstruction-based pseudo-subscore learning for action
  quality assessment in sporting events.
\newblock {\em Applied Intelligence}, 53(9):10053--10067, 2023.

\bibitem{zhou2023hierarchical}
Kanglei Zhou, Yue Ma, Hubert~PH Shum, and Xiaohui Liang.
\newblock Hierarchical graph convolutional networks for action quality
  assessment.
\newblock {\em IEEE Transactions on Circuits and Systems for Video Technology},
  33(12):7749--7763, 2023.

\bibitem{an2024multi}
Qi~An, Mengshi Qi, and Huadong Ma.
\newblock Multi-stage contrastive regression for action quality assessment.
\newblock In {\em IEEE International Conference on Acoustics, Speech and Signal
  Processing}, pages 4110--4114. IEEE, 2024.

\bibitem{huang2024assessing}
Feng Huang and Jianjun Li.
\newblock Assessing action quality with semantic-sequence performance
  regression and densely distributed sample weighting.
\newblock {\em Applied Intelligence}, 54(4):3245--3259, 2024.

\bibitem{ke2024two}
Xiao Ke, Huangbiao Xu, Xiaofeng Lin, and Wenzhong Guo.
\newblock Two-path target-aware contrastive regression for action quality
  assessment.
\newblock {\em Information Sciences}, 664:120347, 2024.

\bibitem{xu2024fineparser}
Jinglin Xu, Sibo Yin, Guohao Zhao, Zishuo Wang, and Yuxin Peng.
\newblock Fineparser: A fine-grained spatio-temporal action parser for
  human-centric action quality assessment.
\newblock In {\em Proceedings of the IEEE/CVF Conference on Computer Vision and
  Pattern Recognition}, pages 14628--14637, 2024.

\bibitem{luo2024rhythmer}
Zhuang Luo, Yang Xiao, Feng Yang, Joey~Tianyi Zhou, and Zhiwen Fang.
\newblock Rhythmer: Ranking-based skill assessment with rhythm-aware
  transformer.
\newblock {\em IEEE Transactions on Circuits and Systems for Video Technology},
  2024.

\bibitem{lea2017edtcn}
Colin Lea, Michael~D Flynn, Rene Vidal, Austin Reiter, and Gregory~D Hager.
\newblock Temporal convolutional networks for action segmentation and
  detection.
\newblock In {\em Proceedings of the IEEE Conference on Computer Vision and
  Pattern Recognition}, pages 156--165, 2017.

\bibitem{vaswani2017attention}
A~Vaswani.
\newblock Attention is all you need.
\newblock {\em Advances in Neural Information Processing Systems}, 2017.

\bibitem{fang2023end}
Hang Fang, Wengang Zhou, and Houqiang Li.
\newblock End-to-end action quality assessment with action parsing transformer.
\newblock In {\em IEEE International Conference on Visual Communications and
  Image Processing}, pages 1--5. IEEE, 2023.

\bibitem{bai2023towards}
Yang Bai.
\newblock {\em Towards Interaction-level Video Action Understanding}.
\newblock Thesis, 2023.

\bibitem{han2023mla}
Chaoyu Han, Fangyao Shen, Lina Chen, Xiaoyi Lian, Hongjie Gou, and Hong Gao.
\newblock Mla-lstm: A local and global location attention lstm learning model
  for scoring figure skating.
\newblock {\em Systems}, 11(1):21, 2023.

\bibitem{lin2024automatic}
Xiuchun Lin, Yichao Liu, Chen Feng, Zhide Chen, Xu~Yang, and Hui Cui.
\newblock Automatic evaluation method for functional movement screening based
  on multi-scale lightweight 3d convolution and an encoder–decoder.
\newblock {\em Electronics}, 13(10):1813, 2024.

\bibitem{hao2023establishment}
Ning Hao, Sihan Ruan, Yiheng Song, Jiashun Chen, and Longgang Tian.
\newblock The establishment of a precise intelligent evaluation system for
  sports events: Diving.
\newblock {\em Heliyon}, 9(11), 2023.

\bibitem{mourchid2023d}
Youssef Mourchid and Rim Slama.
\newblock D-stgcnt: A dense spatio-temporal graph conv-gru network based on
  transformer for assessment of patient physical rehabilitation.
\newblock {\em Computers in Biology and Medicine}, 165:107420, 2023.

\bibitem{chen2024long}
Lina Chen, Junbo Zhang, Weijie Wu, Chaoyu Han, and Hong Gao.
\newblock Long video scoring method fusing high-precision pose and
  spatio-temporal attention modules.
\newblock In {\em Asia-Pacific Web and Web-Age Information Management Joint
  International Conference on Web and Big Data}, pages 466--475. Springer,
  2024.

\bibitem{wang2024action}
Wei Wang, Hongyu Wang, Yingguang Hao, and Qiong Wang.
\newblock Action quality assessment with multi-scale temporal attention
  mechanism.
\newblock In {\em International Conference on Advanced Algorithms and Control
  Engineering}, pages 247--251. IEEE, 2024.

\bibitem{millan2020fine}
Mégane Millan and Catherine Achard.
\newblock Fine-tuning siamese networks to assess sport gestures quality.
\newblock In {\em International Joint Conference on Computer Vision, Imaging
  and Computer Graphics Theory and Applications}, pages 57--65, 2020.

\bibitem{li2021improving}
Jicheng Li, Anjana Bhat, and Roghayeh Barmaki.
\newblock Improving the movement synchrony estimation with action quality
  assessment in children play therapy.
\newblock In {\em Proceedings of the International Conference on Multimodal
  Interaction}, pages 397--406, 2021.

\bibitem{li2018end}
Yongjun Li, Xiujuan Chai, and Xilin Chen.
\newblock End-to-end learning for action quality assessment.
\newblock In {\em Pacific Rim Conference on Multimedia}, pages 125--134.
  Springer, 2018.

\bibitem{li2018scoringnet}
Yongjun Li, Xiujuan Chai, and Xilin Chen.
\newblock Scoringnet: Learning key fragment for action quality assessment with
  ranking loss in skilled sports.
\newblock In {\em Asian Conference on Computer Vision}, pages 149--164.
  Springer, 2018.

\bibitem{liu2019low}
Hou-Chin Liu and Chung-Ta King.
\newblock A low-cost virtual coach for diagnosis and guidance in
  baseball/softball batting training.
\newblock In {\em International Conference on Systems}, 2019.

\bibitem{wang2020assessing}
Jiahao Wang, Zhengyin Du, Annan Li, and Yunhong Wang.
\newblock Assessing action quality via attentive spatio-temporal convolutional
  networks.
\newblock In {\em Chinese Conference on Pattern Recognition and Computer
  Vision}, pages 3--16. Springer, 2020.

\bibitem{wang2020towards}
Tianyu Wang, Yijie Wang, and Mian Li.
\newblock Towards accurate and interpretable surgical skill assessment: A
  video-based method incorporating recognized surgical gestures and skill
  levels.
\newblock In {\em Medical Image Computing and Computer Assisted Intervention},
  pages 668--678. Springer, 2020.

\bibitem{epstein2021learning}
Dave Epstein and Carl Vondrick.
\newblock Learning goals from failure.
\newblock In {\em Proceedings of the IEEE/CVF Conference on Computer Vision and
  Pattern Recognition}, pages 11194--11204, 2021.

\bibitem{parmar2021hallucinet}
Paritosh Parmar and Brendan Morris.
\newblock Hallucinet-ing spatiotemporal representations using a 2d-cnn.
\newblock {\em Signals}, 2(3):604--618, 2021.

\bibitem{parsa2021multi}
Behnoosh Parsa and Ashis~G Banerjee.
\newblock A multi-task learning approach for human activity segmentation and
  ergonomics risk assessment.
\newblock In {\em Proceedings of the IEEE/CVF Winter Conference on Applications
  of Computer Vision}, pages 2352--2362, 2021.

\bibitem{parsa2020deep}
Behnoosh Parsa.
\newblock {\em Deep Learning Methods for Video-Based Human Activity Recognition
  in Industrial Settings}.
\newblock Thesis, 2020.

\bibitem{abedi2021improving}
Ali Abedi and Shehroz~S Khan.
\newblock Improving state-of-the-art in detecting student engagement with
  resnet and tcn hybrid network.
\newblock In {\em Conference on Robots and Vision}, pages 151--157. IEEE, 2021.

\bibitem{li2022precise}
Qiaoge Li, Zhenghang Cui, Itaru Kitahara, and Ryusuke Sagawa.
\newblock Precise gymnastic scoring from tv playback.
\newblock In {\em IEEE Global Conference on Consumer Electronics}, pages
  412--415. IEEE, 2022.

\bibitem{wang2022skeleton}
Xinyu Wang, Jianwei Li, and Haiqing Hu.
\newblock Skeleton-based action quality assessment via partially connected lstm
  with triplet losses.
\newblock In {\em Chinese Conference on Pattern Recognition and Computer
  Vision}, pages 220--232. Springer, 2022.

\bibitem{liu2023multi}
Langxi Liu, Pengjun Zhai, Dulei Zheng, and Yu~Fang.
\newblock Multi-stage action quality assessment method.
\newblock In {\em Proceedings of the International Conference on Control,
  Robotics and Intelligent System}, pages 116--122, 2023.

\bibitem{gao2024resfnn}
Honghao Gao, Si~Yu, Muddesar Iqbal, and Mohsen Guizani.
\newblock Resfnn: Residual structure-based feedforward neural network for
  action quality assessment in sports consumer electronics.
\newblock {\em IEEE Transactions on Consumer Electronics}, 2024.

\bibitem{farabi2022improving}
Shafkat Farabi, Hasibul Himel, Fakhruddin Gazzali, Md~Bakhtiar Hasan,
  Md~Hasanul Kabir, and Moshiur Farazi.
\newblock Improving action quality assessment using weighted aggregation.
\newblock In {\em Iberian Conference on Pattern Recognition and Image
  Analysis}, pages 576--587. Springer, 2022.

\bibitem{li2022hand}
Chenglong Li, Qiwen Zhu, Tubiao Liu, Jin Tang, and Yu~Su.
\newblock Hand hygiene assessment via joint step segmentation and key action
  scorer.
\newblock {\em arXiv preprint arXiv:2209.12221}, 2022.

\bibitem{ding2023sedskill}
Xinpeng Ding, Xiaowei Xu, and Xiaomeng Li.
\newblock Sedskill: Surgical events driven method for skill assessment from
  thoracoscopic surgical videos.
\newblock In {\em International Conference on Medical Image Computing and
  Computer-Assisted Intervention}, pages 35--45. Springer, 2023.

\bibitem{murthy2023divenet}
Pramod Murthy, Bertram Taetz, Arpit Lekhra, and Didier Stricker.
\newblock Divenet: Dive action localization and physical pose parameter
  extraction for high performance training.
\newblock {\em IEEE Access}, 11:37749--37767, 2023.

\bibitem{li2024segmentation}
Zeying Li, Hongtao Chen, Jing Cai, and Yanbing Xue.
\newblock Segmentation and quality assessment of continuous fitness movements
  based on vision.
\newblock In {\em International Conference on Intelligent Computing}, pages
  96--107. Springer, 2024.

\bibitem{farha2019mstcn}
Yazan~Abu Farha and Jurgen Gall.
\newblock Ms-tcn: Multi-stage temporal convolutional network for action
  segmentation.
\newblock In {\em Proceedings of the IEEE/CVF Conference on Computer Vision and
  Pattern Recognition}, pages 3575--3584, 2019.

\bibitem{pan2019action}
Jia-Hui Pan, Jibin Gao, and Wei-Shi Zheng.
\newblock Action assessment by joint relation graphs.
\newblock In {\em Proceedings of the IEEE/CVF International Conference on
  Computer Vision}, pages 6331--6340, 2019.

\bibitem{nagai2021action}
Takasuke Nagai, Shoichiro Takeda, Masaaki Matsumura, Shinya Shimizu, and Susumu
  Yamamoto.
\newblock Action quality assessment with ignoring scene context.
\newblock In {\em IEEE International Conference on Image Processing}, pages
  1189--1193. IEEE, 2021.

\bibitem{nekoui2021eagle}
Mahdiar Nekoui, Fidel Omar~Tito Cruz, and Li~Cheng.
\newblock Eagle-eye: Extreme-pose action grader using detail bird's-eye view.
\newblock In {\em Proceedings of the IEEE/CVF Winter Conference on Applications
  of Computer Vision}, pages 394--402, 2021.

\bibitem{li2022pairwise}
Mingzhe Li, Hong-Bo Zhang, Qing Lei, Zongwen Fan, Jinghua Liu, and Ji-Xiang Du.
\newblock Pairwise contrastive learning network for action quality assessment.
\newblock In {\em European Conference on Computer Vision}, pages 457--473.
  Springer, 2022.

\bibitem{huang2025dual}
Keyi Huang, Yi~Tian, Chen Yu, and Yaping Huang.
\newblock Dual-referenced assistive network for action quality assessment.
\newblock {\em Neurocomputing}, 614:128786, 2025.

\bibitem{sardari2019view}
Faegheh Sardari, Adeline Paiement, and Majid Mirmehdi.
\newblock View-invariant pose analysis for human movement assessment from rgb
  data.
\newblock In {\em Image Analysis and Processing}, pages 237--248. Springer,
  2019.

\bibitem{qiu2022pose}
Yuhang Qiu, Jiping Wang, Zhe Jin, Honghui Chen, Mingliang Zhang, and Liquan
  Guo.
\newblock Pose-guided matching based on deep learning for assessing quality of
  action on rehabilitation training.
\newblock {\em Biomedical Signal Processing and Control}, 72:103323, 2022.

\bibitem{zhu2022robust}
Yun Zhu, Shuang Liang, Peng Li, and Xiaojun Wu.
\newblock Robust human pose quality assessment using optimal sub-pattern
  assignment.
\newblock In {\em International Conference on Control, Automation and
  Information Sciences}, pages 419--423. IEEE, 2022.

\bibitem{hirosawa2023action}
Seiji Hirosawa, Takaaki Kato, Takayoshi Yamashita, and Yoshimitsu Aoki.
\newblock Action quality assessment model using specialists’ gaze location
  and kinematics data—focusing on evaluating figure skating jumps.
\newblock {\em Sensors}, 23(22):9282, 2023.

\bibitem{hirosawa2023expert}
Seiji Hirosawa, Takayoshi Yamashita, and Yoshimitsu Aoki.
\newblock Expert’s gaze-based prediction model for assessing the quality of
  figure skating jumps.
\newblock In {\em International Symposium on Computer Science in Sport}, pages
  42--52. Springer, 2023.

\bibitem{hirosawa2024computer}
Seiji Hirosawa.
\newblock {\em Computer Vision-based Action Quality Assessment Incorporating
  Human Expert’s Knowledge: Focusing on the Figure Skating Jump}.
\newblock Thesis, 2024.

\bibitem{lei2020learning}
Qing Lei, Hong-Bo Zhang, Ji-Xiang Du, Tsung-Chih Hsiao, and Chih-Cheng Chen.
\newblock Learning effective skeletal representations on rgb video for
  fine-grained human action quality assessment.
\newblock {\em Electronics}, 9(4):568, 2020.

\bibitem{nekoui2020falcons}
Mahdiar Nekoui, Fidel Omar~Tito Cruz, and Li~Cheng.
\newblock Falcons: Fast learner-grader for contorted poses in sports.
\newblock In {\em Proceedings of the IEEE/CVF Conference on Computer Vision and
  Pattern Recognition Workshops}, pages 900--901, 2020.

\bibitem{fan2022hightlight}
Shun Fan, Yuantai Wei, Jingfei Xia, and Feng Zheng.
\newblock Hightlight video detection in figure skating.
\newblock In {\em Chinese Conference on Pattern Recognition and Computer
  Vision}, pages 651--664. Springer, 2022.

\bibitem{zhang2023toward}
Zhitao Zhang, Zhengyou Wang, Shanna Zhuang, and Jiahui Wang.
\newblock Toward action recognition and assessment using sfagcn and combinative
  regression model of spatiotemporal features.
\newblock {\em Applied Intelligence}, 53(1):757--768, 2023.

\bibitem{zhou2023video}
Kanglei Zhou, Ruizhi Cai, Yue Ma, Qingqing Tan, Xinning Wang, Jianguo Li,
  Hubert~PH Shum, Frederick~WB Li, Song Jin, and Xiaohui Liang.
\newblock A video-based augmented reality system for human-in-the-loop muscle
  strength assessment of juvenile dermatomyositis.
\newblock {\em IEEE Transactions on Visualization and Computer Graphics},
  29(5):2456--2466, 2023.

\bibitem{chen2024unlabeled}
Renguang Chen, Guolong Zheng, Xu~Yang, Zhide Chen, Jiwu Shu, Wencheng Yang,
  Kexin Zhu, and Chen Feng.
\newblock Unlabeled action quality assessment based on multi-dimensional
  adaptive constrained dynamic time warping.
\newblock {\em arXiv preprint arXiv:2410.14161}, 2024.

\bibitem{gallardo2024gymetricpose}
Ulises Gallardo, Fernando Caro, Eluney Hernández, Ricardo Espinosa, and
  Gilberto Ochoa-Ruiz.
\newblock Gymetricpose: A light-weight angle-based graph adaptation for action
  quality assessment.
\newblock In {\em IEEE International Symposium on Computer-Based Medical
  Systems}, pages 43--50. IEEE, 2024.

\bibitem{wen2024learning}
Hongli Wen and Yang Xu.
\newblock Learning to score sign language with two-stage method.
\newblock {\em arXiv preprint arXiv:2404.10383}, 2024.

\bibitem{wu2024research}
Gang Wu, Tan Li, Yuqi Zhou, Jin Guo, Jingyu Zhu, Nanjiang Chen, Weining Song,
  Yalan Xing, Xianghui Meng, and Yanwen Lin.
\newblock Research on worker action recognition and evaluation in intelligent
  manufacturing training based on industrial metaverse.
\newblock In {\em Asia Simulation Conference}, pages 365--380. Springer, 2024.

\bibitem{zhou2024attention}
Chengju Zhou, Jiayu Zeng, Lina Qiu, Shuxi Wang, Pingzhi Liu, and Jiahui Pan.
\newblock An attention-based adaptive spatial–temporal graph convolutional
  network for long-video ergonomic risk assessment.
\newblock {\em Engineering Applications of Artificial Intelligence},
  131:107780, 2024.

\bibitem{lemos2021human}
Renato~Manuel Lemos~Baptista.
\newblock {\em Human motion analysis using 3D skeleton representation in the
  context of real-world applications: from home-based rehabilitation to sensing
  in the wild}.
\newblock Thesis, 2021.

\bibitem{nekoui2022intelligent}
Mahdiar Nekoui.
\newblock {\em Intelligent Video-based Quality Assessment of Human Activities}.
\newblock Thesis, 2022.

\bibitem{ganesh2019novel}
Yaparla Ganesh, Allaparthi Sri~Teja, Sai~Krishna Munnangi, and Garimella
  Rama~Murthy.
\newblock A novel framework for fine grained action recognition in soccer.
\newblock In {\em Advances in Computational Intelligence}, pages 137--150.
  Springer, 2019.

\bibitem{li2019manipulation}
Zhenqiang Li, Yifei Huang, Minjie Cai, and Yoichi Sato.
\newblock Manipulation-skill assessment from videos with spatial attention
  network.
\newblock In {\em Proceedings of the IEEE/CVF International Conference on
  Computer Vision Workshops}, pages 0--0, 2019.

\bibitem{li2021and}
Jianwei Li, Qingrui Hu, Tianxiao Guo, Siqi Wang, and Yanfei Shen.
\newblock What and how well you exercised? an efficient analysis framework for
  fitness actions.
\newblock {\em Journal of Visual Communication and Image Representation},
  80:103304, 2021.

\bibitem{freire2022towards}
David Freire-Obregón, Javier Lorenzo-Navarro, Oliverio~J Santana, Daniel
  Hernández-Sosa, and Modesto Castrillón-Santana.
\newblock Towards cumulative race time regression in sports: I3d convnet
  transfer learning in ultra-distance running events.
\newblock In {\em International Conference on Pattern Recognition}, pages
  805--811. IEEE, 2022.

\bibitem{wang2023three}
Yan Wang, JinWei Wang, and Xue Bai.
\newblock Three-stream fusion networks for student engagement recognition based
  on timesformer.
\newblock In {\em International Conference on Artificial Intelligence and
  Computer Engineering}, volume 12610, pages 758--764. SPIE, 2023.

\bibitem{zhao2023knowledge}
Wenting Zhao, Shigang Wang, Yan Zhao, Jian Wei, and Tianshu Li.
\newblock Knowledge and data co-driven intelligent assessment of chinese zither
  fingerings.
\newblock {\em Displays}, 78:102442, 2023.

\bibitem{anastasiou2023keep}
Dimitrios Anastasiou, Yueming Jin, Danail Stoyanov, and Evangelos Mazomenos.
\newblock Keep your eye on the best: contrastive regression transformer for
  skill assessment in robotic surgery.
\newblock {\em IEEE Robotics and Automation Letters}, 8(3):1755--1762, 2023.

\bibitem{freire2023x3d}
David Freire-Obregón, Javier Lorenzo-Navarro, Oliverio~J Santana, Daniel
  Hernández-Sosa, and Modesto Castrillón-Santana.
\newblock An x3d neural network analysis for runner’s performance assessment
  in a wild sporting environment.
\newblock In {\em International Conference on Machine Vision and Applications},
  pages 1--5. IEEE, 2023.

\bibitem{zhou2023prior}
Haoyang Zhou, Teng Hou, and Jitao Li.
\newblock Prior knowledge-guided hierarchical action quality assessment with 3d
  convolution and attention mechanism.
\newblock In {\em Journal of Physics: Conference Series}, volume 2632, page
  012027. IOP Publishing, 2023.

\bibitem{kondo2024zeal}
Satoshi Kondo.
\newblock Zeal: Surgical skill assessment with zero-shot tool inference using
  unified foundation model.
\newblock {\em arXiv preprint arXiv:2407.02738}, 2024.

\bibitem{huang2024full}
Yuxin Huang, Yiwei Yuan, Xiangyu Zeng, Ling Xie, Yiyu Fu, Guanghui Yue, and
  Baoquan Zhao.
\newblock Full-reference motion quality assessment based on efficient monocular
  parametric 3d human body reconstruction.
\newblock In {\em IEEE International Conference on Multimedia and Expo}, pages
  1--6. IEEE, 2024.

\bibitem{emerson2009bais1}
John~W Emerson, Miki Seltzer, and David Lin.
\newblock Assessing judging bias: An example from the 2000 olympic games.
\newblock {\em The American Statistician}, 63(2):124--131, 2009.

\bibitem{ansorge1988bais2}
Charles~J Ansorge and John~K Scheer.
\newblock International bias detected in judging gymnastic competition at the
  1984 olympic games.
\newblock {\em Research Quarterly for Exercise and Sport}, 59(2):103--107,
  1988.

\bibitem{morgan2014harder}
Hillary~N Morgan and Kurt~W Rotthoff.
\newblock The harder the task, the higher the score: Findings of a difficulty
  bias.
\newblock {\em Economic Inquiry}, 52(3):1014--1026, 2014.

\bibitem{tang2020uncertainty}
Yansong Tang, Zanlin Ni, Jiahuan Zhou, Danyang Zhang, Jiwen Lu, Ying Wu, and
  Jie Zhou.
\newblock Uncertainty-aware score distribution learning for action quality
  assessment.
\newblock In {\em Proceedings of the IEEE/CVF Conference on Computer Vision and
  Pattern Recognition}, pages 9839--9848, 2020.

\bibitem{zhou2022uncertainty}
Caixia Zhou, Yaping Huang, and Haibin Ling.
\newblock Uncertainty-driven action quality assessment.
\newblock {\em arXiv preprint arXiv:2207.14513}, 2022.

\bibitem{karunaratne2021objectively}
Anandi Karunaratne, Chamin Jayasooriya, Sampath Deegalla, and Rajitha
  Navarathna.
\newblock Objectively measure player performance on olympic weightlifting.
\newblock In {\em International Conference on Information and Automation for
  Sustainability}, pages 410--415. IEEE, 2021.

\bibitem{li2023gaussian}
Ming-Zhe Li, Hong-Bo Zhang, Li-Jia Dong, Qing Lei, and Ji-Xiang Du.
\newblock Gaussian guided frame sequence encoder network for action quality
  assessment.
\newblock {\em Complex \& Intelligent Systems}, 9(2):1963--1974, 2023.

\bibitem{majeedi2024rica}
Abrar Majeedi, Viswanatha~Reddy Gajjala, Satya Sai Srinath~Namburi GNVV, and
  Yin Li.
\newblock Rica2: Rubric-informed, calibrated assessment of actions.
\newblock {\em arXiv preprint arXiv:2408.02138}, 2024.

\bibitem{zhang2024auto}
Boyu Zhang, Jiayuan Chen, Yinfei Xu, Hui Zhang, Xu~Yang, and Xin Geng.
\newblock Auto-encoding score distribution regression for action quality
  assessment.
\newblock {\em Neural Computing and Applications}, 36(2):929--942, 2024.

\bibitem{jain2021action}
Hiteshi Jain, Gaurav Harit, and Avinash Sharma.
\newblock Action quality assessment using siamese network-based deep metric
  learning.
\newblock {\em IEEE Transactions on Circuits and Systems for Video Technology},
  31(6):2260--2273, 2021.

\bibitem{sun2023novel}
WenHao Sun, YanXiang Hu, Bo~Zhang, XinRan Chen, CaiXia Hao, and YaRu Gao.
\newblock A novel blind action quality assessment based on multi-headed gru
  network and attention mechanism.
\newblock In {\em International Conference on Artificial Intelligence,
  Automation, and High-Performance Computing}, volume 12717, pages 835--843.
  SPIE, 2023.

\bibitem{joung2023contrastive}
Chung-In Joung, Seunghwan Byun, and Seungjun Baek.
\newblock Contrastive learning for action assessment using graph convolutional
  networks with augmented virtual joints.
\newblock {\em IEEE Access}, 2023.

\bibitem{liu2024hierarchical}
Yanchao LIU, Xina CHENG, and Takeshi IKENAGA.
\newblock A hierarchical joint training based replay-guided contrastive
  transformer for action quality assessment of figure skating.
\newblock {\em IEICE Transactions on Fundamentals of Electronics,
  Communications and Computer Sciences}, 2024.

\bibitem{jain2020assessment}
Hiteshi Jain.
\newblock {\em Assessment of Human Actions in Videos}.
\newblock Thesis, 2020.

\bibitem{du2023learning}
Zexing Du, Di~He, Xue Wang, and Qing Wang.
\newblock Learning semantics-guided representations for scoring figure skating.
\newblock {\em IEEE Transactions on Multimedia}, 2023.

\bibitem{lei2023multi}
Qing Lei, Huiying Li, Hongbo Zhang, Jixiang Du, and Shangce Gao.
\newblock Multi-skeleton structures graph convolutional network for action
  quality assessment in long videos.
\newblock {\em Applied Intelligence}, 53(19):21692--21705, 2023.

\bibitem{xia2023skating}
Jingfei Xia, Mingchen Zhuge, Tiantian Geng, Shun Fan, Yuantai Wei, Zhenyu He,
  and Feng Zheng.
\newblock Skating-mixer: Long-term sport audio-visual modeling with mlps.
\newblock In {\em Proceedings of the AAAI Conference on Artificial
  Intelligence}, volume~37, pages 2901--2909, 2023.

\bibitem{zheng2023skeleton}
Kaili Zheng, Ji~Wu, Jialin Zhang, and Chenyi Guo.
\newblock A skeleton-based rehabilitation exercise assessment system with
  rotation invariance.
\newblock {\em IEEE Transactions on Neural Systems and Rehabilitation
  Engineering}, 31:2612--2621, 2023.

\bibitem{ding20242m}
Yuning Ding, Sifan Zhang, Liu Shenglan, Jinrong Zhang, Wenyue Chen, Duan
  Haifei, Bingcheng Dong, and Tao Sun.
\newblock 2m-af: A strong multi-modality framework for human action quality
  assessment with self-supervised representation learning.
\newblock In {\em Proceedings of the ACM International Conference on
  Multimedia}, pages 1564--1572, 2024.

\bibitem{gedamu2024visual}
Kumie Gedamu, Yanli Ji, Yang Yang, Jie Shao, and Heng~Tao Shen.
\newblock Visual-semantic alignment temporal parsing for action quality
  assessment.
\newblock {\em IEEE Transactions on Circuits and Systems for Video Technology},
  2024.

\bibitem{he2024expert}
Tian He, Yang Chen, Ling Wang, and Hong Cheng.
\newblock An expert-knowledge-based graph convolutional network for
  skeleton-based physical rehabilitation exercises assessment.
\newblock {\em IEEE Transactions on Neural Systems and Rehabilitation
  Engineering}, 2024.

\bibitem{zahan2024learning}
Sania Zahan, Ghulam~Mubashar Hassan, and Ajmal Mian.
\newblock Learning sparse temporal video mapping for action quality assessment
  in floor gymnastics.
\newblock {\em IEEE Transactions on Instrumentation and Measurement}, 2024.

\bibitem{zhang2024narrative}
Shiyi Zhang, Sule Bai, Guangyi Chen, Lei Chen, Jiwen Lu, Junle Wang, and
  Yansong Tang.
\newblock Narrative action evaluation with prompt-guided multimodal
  interaction.
\newblock In {\em Proceedings of the IEEE/CVF Conference on Computer Vision and
  Pattern Recognition}, pages 18430--18439, 2024.

\bibitem{parmar2019actionb}
Paritosh Parmar.
\newblock {\em On Action Quality Assessment}.
\newblock Thesis, 2019.

\bibitem{xu2024vision}
Huangbiao Xu, Xiao Ke, Yuezhou Li, Rui Xu, Huanqi Wu, Xiaofeng Lin, and
  Wenzhong Guo.
\newblock Vision-language action knowledge learning for semantic-aware action
  quality assessment.
\newblock In {\em European Conference on Computer Vision}, pages 423--440.
  Springer, 2024.

\bibitem{li2022skeleton}
Huiying Li, Qing Lei, Hongbo Zhang, Jixiang Du, and Shangce Gao.
\newblock Skeleton-based deep pose feature learning for action quality
  assessment on figure skating videos.
\newblock {\em Journal of Visual Communication and Image Representation},
  89:103625, 2022.

\bibitem{li2022tai}
Jianwei Li, Haiqing Hu, Qingjun Xing, Xinyu Wang, Jinyang Li, and Yanfei Shen.
\newblock Tai chi action quality assessment and visual analysis with a consumer
  rgb-d camera.
\newblock In {\em IEEE International Workshop on Multimedia Signal Processing},
  pages 1--6. IEEE, 2022.

\bibitem{cui2023study}
Wen-Bo Cui, Wen-Ai Song, Zi-Tong Pei, Yi~Lei, Qing Wang, Yan-Jie Chen, and
  Ji-Jiang Yang.
\newblock Study on assessment methods of developmental coordination disorder in
  children.
\newblock In {\em IEEE Annual Computers, Software, and Applications
  Conference}, pages 1507--1512. IEEE, 2023.

\bibitem{siow2023evaluating}
Chyan~Zheng Siow, Wei~Hong Chin, and Naoyuki Kubota.
\newblock Evaluating simple exercises with a fuzzy system based on human
  skeleton poses.
\newblock In {\em IEEE International Conference on Fuzzy Systems}, pages 1--6.
  IEEE, 2023.

\bibitem{yao2023contrastive}
Long Yao, Qing Lei, Hongbo Zhang, Jixiang Du, and Shangce Gao.
\newblock A contrastive learning network for performance metric and assessment
  of physical rehabilitation exercises.
\newblock {\em IEEE Transactions on Neural Systems and Rehabilitation
  Engineering}, 2023.

\bibitem{zhang2023action}
Dinghuang Zhang, Dalin Zhou, and Honghai Liu.
\newblock Action quality assessment for asd behaviour evaluation.
\newblock In {\em International Conference on Machine Learning and
  Cybernetics}, pages 483--488. IEEE, 2023.

\bibitem{shen2024markerless}
Yuan-Yuan Shen, Qing-Jun Xing, and Yan-Fei Shen.
\newblock Markerless vision-based functional movement screening movements
  evaluation with deep neural networks.
\newblock {\em Iscience}, 27(1), 2024.

\bibitem{xu2024reveal}
Siyuan Xu, Peilin Chen, Yue Liu, Meng Wang, Shiqi Wang, and Sam Kwong.
\newblock Reveal fluidity behind frames: A multi-modality framework for action
  quality assessment.
\newblock In {\em IEEE International Workshop on Multimedia Signal Processing},
  pages 1--6. IEEE, 2024.

\bibitem{louis2024improving}
Nathan Louis.
\newblock {\em Improving Articulated Pose Tracking and Contact Force Estimation
  for Qualitative Assessment of Human Actions}.
\newblock Thesis, 2024.

\bibitem{zhang2023hand}
Dinghuang Zhang.
\newblock {\em Hand-Eye Behaviour Analytics for Children with Autism Spectrum
  Disorder}.
\newblock Thesis, 2023.

\bibitem{zahan2024human}
Sania Zahan.
\newblock {\em Human Motion Analysis}.
\newblock Thesis, 2024.

\bibitem{kim2024kinematic}
Doyoung Kim, Taewan Kim, Inwoong Lee, and Sanghoon Lee.
\newblock Kinematic diversity and rhythmic alignment in choreographic quality
  transformers for dance quality assessment.
\newblock {\em IEEE Transactions on Circuits and Systems for Video Technology},
  2024.

\bibitem{pan2022adaptive}
Jia-Hui Pan, Jibin Gao, and Wei-Shi Zheng.
\newblock Adaptive action assessment.
\newblock {\em IEEE Transactions on Pattern Analysis and Machine Intelligence},
  44(12):8779--8795, 2022.

\bibitem{zhang2023adaptive}
Shao-Jie Zhang, Jia-Hui Pan, Jibin Gao, and Wei-Shi Zheng.
\newblock Adaptive stage-aware assessment skill transfer for skill
  determination.
\newblock {\em IEEE Transactions on Multimedia}, 2023.

\bibitem{zhou2024cofinal}
Kanglei Zhou, Junlin Li, Ruizhi Cai, Liyuan Wang, Xingxing Zhang, and Xiaohui
  Liang.
\newblock Cofinal: Enhancing action quality assessment with coarse-to-fine
  instruction alignment.
\newblock In {\em International Joint Conference on Artificial Intelligence},
  2024.

\bibitem{dadashzadeh2024pecop}
Amirhossein Dadashzadeh, Shuchao Duan, Alan Whone, and Majid Mirmehdi.
\newblock Pecop: Parameter efficient continual pretraining for action quality
  assessment.
\newblock In {\em Proceedings of the IEEE/CVF Winter Conference on Applications
  of Computer Vision}, pages 42--52, 2024.

\bibitem{li2024continual}
Yuan-Ming Li, Ling-An Zeng, Jing-Ke Meng, and Wei-Shi Zheng.
\newblock Continual action assessment via task-consistent score-discriminative
  feature distribution modeling.
\newblock {\em IEEE Transactions on Circuits and Systems for Video Technology},
  2024.

\bibitem{zhou2024magr}
Kanglei Zhou, Liyuan Wang, Xingxing Zhang, Hubert~PH Shum, Frederick~WB Li,
  Jianguo Li, and Xiaohui Liang.
\newblock Magr: Manifold-aligned graph regularization for continual action
  quality assessment.
\newblock In {\em European Conference on Computer Vision}, pages 375--392.
  Springer, Cham, 2024.

\bibitem{dadashzadeh2024learning}
Amirhossein Dadashzadeh.
\newblock {\em Learning Strategies for Parkinson’s Disease Severity
  Assessment}.
\newblock Thesis, 2024.

\bibitem{seo2021extracting}
Chanjin Seo, Masato Sabanai, Yuta Goto, Koji Tagami, Hiroyuki Ogata, Kazuyuki
  Kanosue, and Jun Ohya.
\newblock Extracting and interpreting unknown factors with classifier for foot
  strike types in running.
\newblock In {\em International Conference on Pattern Recognition}, pages
  3217--3224. IEEE, 2021.

\bibitem{wang2021towards}
Tianyu Wang, Minhao Jin, and Mian Li.
\newblock Towards accurate and interpretable surgical skill assessment: a
  video-based method for skill score prediction and guiding feedback
  generation.
\newblock {\em International Journal of Computer Assisted Radiology and
  Surgery}, 16(9):1595--1605, 2021.

\bibitem{hirosawa2022determinant}
Seiji Hirosawa, Michiko Watanabe, and Yoshimitsu Aoki.
\newblock Determinant analysis and developing evaluation indicators of grade of
  execution score of double axel jump in figure skating.
\newblock {\em Journal of Sports Sciences}, 40(4):470--481, 2022.

\bibitem{matsuyama2023iris}
Hitoshi Matsuyama, Nobuo Kawaguchi, and Brian~Y Lim.
\newblock Iris: Interpretable rubric-informed segmentation for action quality
  assessment.
\newblock In {\em Proceedings of the International Conference on Intelligent
  User Interfaces}, pages 368--378, 2023.

\bibitem{dong2024interpretable}
Xu~Dong, Xinran Liu, Wanqing Li, Anthony Adeyemi-Ejeye, and Andrew Gilbert.
\newblock Interpretable long-term action quality assessment.
\newblock {\em arXiv preprint arXiv:2408.11687}, 2024.

\bibitem{liu2021towards}
Daochang Liu, Qiyue Li, Tingting Jiang, Yizhou Wang, Rulin Miao, Fei Shan, and
  Ziyu Li.
\newblock Towards unified surgical skill assessment.
\newblock In {\em Proceedings of the IEEE/CVF Conference on Computer Vision and
  Pattern Recognition}, pages 9522--9531, 2021.

\bibitem{wang2022will}
Jiahao Wang, Yunhong Wang, Nina Weng, Tianrui Chai, Annan Li, Faxi Zhang, and
  Sansi Yu.
\newblock Will you ever become popular? learning to predict virality of dance
  clips.
\newblock {\em ACM Transactions on Multimedia Computing, Communications, and
  Applications}, 18(2):1--24, 2022.

\bibitem{jain2019unsupervised}
Hiteshi Jain and Gaurav Harit.
\newblock An unsupervised sequence-to-sequence autoencoder based human action
  scoring model.
\newblock In {\em IEEE Global Conference on Signal and Information Processing},
  pages 1--5. IEEE, 2019.

\bibitem{du2021assessing}
Chen Du, Sarah Graham, Colin Depp, and Truong Nguyen.
\newblock Assessing physical rehabilitation exercises using graph convolutional
  network with self-supervised regularization.
\newblock In {\em Annual International Conference of the IEEE Engineering in
  Medicine \& Biology Society}, pages 281--285. IEEE, 2021.

\bibitem{roditakis2021towards}
Konstantinos Roditakis, Alexandros Makris, and Antonis Argyros.
\newblock Towards improved and interpretable action quality assessment with
  self-supervised alignment.
\newblock In {\em Proceedings of the PErvasive Technologies Related to
  Assistive Environments Conference}, pages 507--513, 2021.

\bibitem{zhang2022semi}
Shao-Jie Zhang, Jia-Hui Pan, Jibin Gao, and Wei-Shi Zheng.
\newblock Semi-supervised action quality assessment with self-supervised
  segment feature recovery.
\newblock {\em IEEE Transactions on Circuits and Systems for Video Technology},
  32(9):6017--6028, 2022.

\bibitem{lee2022self}
Kyoungoh Lee, Yeseung Park, Jungwoo Huh, Jiwoo Kang, and Sanghoon Lee.
\newblock Self-updatable database system based on human motion assessment
  framework.
\newblock {\em IEEE Transactions on Circuits and Systems for Video Technology},
  32(10):7160--7176, 2022.

\bibitem{zhong2023contrastive}
Yun Zhong, Fan Zhang, and Yiannis Demiris.
\newblock Contrastive self-supervised learning for automated multi-modal dance
  performance assessment.
\newblock In {\em IEEE International Conference on Acoustics, Speech and Signal
  Processing}, pages 1--5. IEEE, 2023.

\bibitem{gedamu2024self}
Kumie Gedamu, Yanli Ji, Yang Yang, Jie Shao, and Heng~Tao Shen.
\newblock Self-supervised subaction parsing network for semi-supervised action
  quality assessment.
\newblock {\em IEEE Transactions on Image Processing}, 2024.

\bibitem{yun2024semi}
Wulian Yun, Mengshi Qi, Fei Peng, and Huadong Ma.
\newblock Semi-supervised teacher-reference-student architecture for action
  quality assessment.
\newblock In {\em European Conference on Computer Vision}, pages 161--178.
  Springer, Cham, 2024.

\bibitem{zhong2024dancemvp}
Yun Zhong and Yiannis Demiris.
\newblock Dancemvp: Self-supervised learning for multi-task primitive-based
  dance performance assessment via transformer text prompting.
\newblock In {\em Proceedings of the AAAI Conference on Artificial
  Intelligence}, volume~38, pages 10270--10278, 2024.

\bibitem{dong2024lucidaction}
Linfeng Dong, Wei Wang, Yu~Qiao, and Xiao Sun.
\newblock Lucidaction: A hierarchical and multi-model dataset for comprehensive
  action quality assessment.
\newblock In {\em Advances in Neural Information Processing Systems}, 2024.

\bibitem{liu2024vision}
Jiang Liu, Huasheng Wang, Katarzyna Stawarz, Shiyin Li, Yao Fu, and Hantao Liu.
\newblock Vision-based human action quality assessment: A systematic review.
\newblock {\em Expert Systems with Applications}, page 125642, 2024.

\bibitem{zhou2024comprehensive}
Kanglei Zhou, Ruizhi Cai, Liyuan Wang, Hubert~PH Shum, and Xiaohui Liang.
\newblock A comprehensive survey of action quality assessment: Method and
  benchmark.
\newblock {\em arXiv preprint arXiv:2412.11149}, 2024.

\bibitem{Genesis}
Genesis Authors.
\newblock Genesis: A universal and generative physics engine for robotics and
  beyond, December 2024.

\bibitem{wang2024nssurvey}
Wenguan Wang, Yi~Yang, and Fei Wu.
\newblock Towards data-and knowledge-driven ai: a survey on neuro-symbolic
  computing.
\newblock {\em IEEE Transactions on Pattern Analysis and Machine Intelligence},
  2024.

\bibitem{guo2020casualitysurvey}
Ruocheng Guo, Lu~Cheng, Jundong Li, P~Richard Hahn, and Huan Liu.
\newblock A survey of learning causality with data: Problems and methods.
\newblock {\em ACM Computing Surveys}, 53(4):1--37, 2020.

\bibitem{castro2020causalitymed}
Daniel~C Castro, Ian Walker, and Ben Glocker.
\newblock Causality matters in medical imaging.
\newblock {\em Nature Communications}, 11(1):3673, 2020.

\bibitem{parmar2024learning}
Paritosh Parmar, Eric Peh, and Basura Fernando.
\newblock Learning to Visually Connect Actions and their Effects.
\newblock {\em arXiv preprint arXiv:2401.10805}, 2024.

\bibitem{grauman2024ego}
Kristen Grauman, et al. 
\newblock Ego-exo4d: Understanding skilled human activity from first-and third-person perspectives.
\newblock In {\em Proceedings of the IEEE/CVF Conference on Computer Vision and
  Pattern Recognition}, pages 19383--19400, 2024.

\end{thebibliography}

\end{document}